%% file: eacl2021.tex
\pdfoutput=1

\documentclass[11pt]{article}

\usepackage{naacl2021}

\usepackage{times}
\usepackage{latexsym}

\usepackage[T1]{fontenc}

\usepackage[utf8]{inputenc}

\usepackage{microtype}
\usepackage{epsfig}
\usepackage{graphicx}
\usepackage{amsmath}
\usepackage{amssymb}
\usepackage[font={small}]{caption}
\usepackage{pifont}
\usepackage{float}

\usepackage{url}            
\usepackage{booktabs}       
\usepackage{nicefrac}       
\usepackage{microtype}      
\usepackage[inline]{enumitem}
\usepackage{color}

\newcommand*{\Scale}[2][4]{\scalebox{#1}{$#2$}}%
\usepackage{footnote}
\usepackage{footmisc}
\newcommand{\cmark}{\ding{51}}%
\newcommand{\xmark}{\ding{55}}%

\newcommand{\beas}{\begin{eqnarray*}}
\newcommand{\eeas}{\end{eqnarray*}}
\newcommand{\bea}{\begin{eqnarray}}
\newcommand{\eea}{\end{eqnarray}}
\newcommand{\bes}{\begin{equation*}}
\newcommand{\ees}{\end{equation*}}
\newcommand{\be}{\begin{equation}}
\newcommand{\ee}{\end{equation}}

\makeatletter
\usepackage{xspace}
\def\@onedot{\ifx\@let@token.\else.\null\fi\xspace}
\DeclareRobustCommand\onedot{\futurelet\@let@token\@onedot}
\newcommand{\figref}[1]{Fig\onedot~\ref{#1}}

\newcommand{\tabref}[1]{Tab\onedot~\ref{#1}}

\def\eg{\emph{e.g}\onedot} 
\def\ie{\emph{i.e}\onedot} \def\Ie{\emph{I.e}\onedot}
 
 \def\vs{\emph{vs}\onedot}
 
\def\etal{\emph{et al}\onedot}

\usepackage{microtype}

\title{Ensemble of MRR and NDCG models for Visual Dialog}

\author{Idan Schwartz \\
  Technion \\
  NetApp \\
  \texttt{idansc@cs.technion.ac.il} }

\date{}

\begin{document}
\maketitle
\begin{abstract}
Assessing an AI agent that can converse in human language and understand visual content is challenging. Generation metrics, such as BLEU scores favor correct syntax over semantics. Hence a discriminative approach is often used, where an agent ranks a set of candidate options. The mean reciprocal rank (MRR)  metric evaluates the model performance by taking into account the rank of a single human-derived answer.  This approach, however, raises a new challenge: the ambiguity and synonymy of answers, for instance, semantic equivalence (\eg, `yeah' and `yes'). To address this, the normalized discounted cumulative gain (NDCG) metric has been used to capture the relevance of all the correct answers via dense annotations. However, the NDCG metric favors the usually applicable uncertain answers such as `I don't know.' Crafting a model that excels on both MRR and NDCG metrics is challenging~\citep{murahari2019large}. Ideally, an AI agent should answer a human-like reply and validate the correctness of any answer. To address this issue, we describe a two-step non-parametric ranking approach that can merge strong MRR and NDCG models. Using our approach, we manage to keep most MRR state-of-the-art performance (70.41\% \vs 71.24\%)  and the NDCG state-of-the-art performance (72.16\% \vs 75.35\%). Moreover, our approach won the recent Visual Dialog 2020 challenge. Source code is available at \url{https://github.com/idansc/mrr-ndcg}.

\end{abstract}

\section{Introduction}

\citet{visdial} introduced the task of Visual Dialog, which requires an agent to converse about visual input. Evaluating visually aware conversation should examine both linguistic properties and visual reasoning. Analysis of generative metrics for dialog often shows no correlation with human judgments \citep{liu2016not}. Hence, to evaluate the correctness of the candidate answers, a retrieval approach is preferred. Two metrics are standard, MRR and NDCG. The MRR metric focuses on a single human-derived ground-truth answer. Despite preferring the more human-like answer, the metric ignores many correct candidate answers. Differently, the NDCG considers the rank of all the correct answers.  The metric relies on dense annotation, where three annotators were asked to mark all the correct candidate answers. However, the candidate answers are generated plausible answers. The analysis shows that the NDCG metric favors uncertain, generally correct answers, such as ``not sure'' \citep{murahari2019large, qi2019two}.


\begin{figure}[t]
	\centering
	\includegraphics[width=1\linewidth]{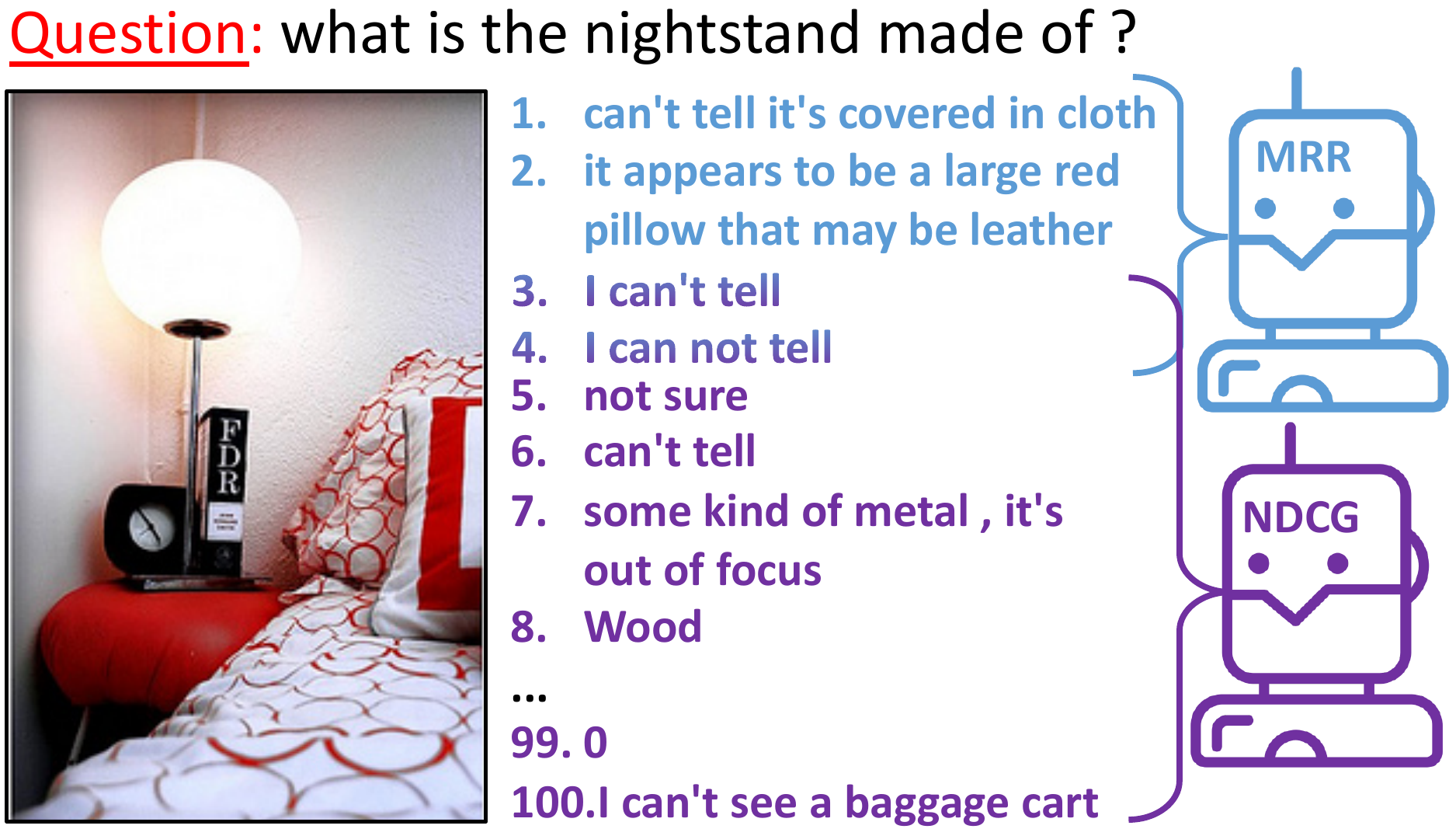}
	 \caption{A visual dialog interaction. The question asks, ``what is the nightstand made of ?''. We show our final ranking, created by the ensemble of an MRR/NDCG models' rankings.  The MRR/NDCG models are trained to optimize the MRR/NDCG metric.  The MRR metric measures the number of retrievals to retrieve the human-derived answer. Hence, the MRR model favors human-like and detailed answers. On the other hand, the NDCG metric measures the rank of all the correct candidates based on dense annotation, which are often general and uncertain. Our ensemble approach seeks a minimal candidate set that is likely to contain the human-derived answer. The remaining candidates are ranked according to the NDCG model.}
	\label{fig:intro}
\end{figure}
Prior work in visual dialog focused on a single metric. Ideally, an AI agent should answer human-like and detailed reply (the MRR metric) and be able to validate the correctness of any answer (the NDCG metric). However, crafting a model that excels in both metrics is challenging~\citep{murahari2019large}. 
To this end, we propose principals to ensemble the rankings of strong MRR and NDCG models. Our approach is to find a minimal set that is likely to hold the human-derived answer. This permits ranking the rest of the candidates according to the NDCG model. Our approach won the recent Visual Dialog 2020 challenge and achieved strong performance on both the MRR and the NDCG metrics simultaneously.

\section{Related Work}

\noindent\textbf{Visual conversation evaluation:}  Early attempts to marry conversation with vision used street scene images, and binary questions \citep{geman2015visual}.  While binary answers are easy to verify, such an approach is limiting for an AI agent.  On the other hand, analysis of generative metrics for dialog often show no correlation with human judgements \citep{liu2016not}. Intuitively, metrics like BLEU-scores rely on corresponding words with the ground-truth answer and often miss synonyms or the subjective nature. More importantly, generative metrics are geared toward textual assessment rather than visual reasoning, which results in models mainly relying on textual cues \citep{schwartz2019simple}.  \citet{malinowski2014multi}  suggest Wu-Palmer similarity metric that calculates similarity based on the depth of two words based on the WordNet taxonomy \citep{miller1995wordnet}. A different approach suggested in the VQA dataset focus only on brief, mostly 1-word answers \citep{VQA}. In this setup, the task turns into popular answers classification, alleviating many text-generation challenges. Notably, VQA requires 3 out of 10 annotators to agree on the answer, which is robust to inter-person variation.  Still, accuracy ignores the reasoning process. \citet{hudson2019gqa} propose GQA, which extends the accuracy metric and uses a scene graph for both question generation and evaluation. Following, \citet{visdial} propose the VisDial dataset for the visual dialog task, which formulates multiple image-language interactions via a dialog. Concurrently, \citet{de2016guesswhat} propose GuessWhat, a goal-driven dialog dataset for object identification. Different from VQA and goal-driven dialogs, the VisDial answers are detailed and more human-like.  For instance, in \figref{fig:intro}, the answer is ``Can't tell...cloth'', while a VQA answer would be ``cloth''. Therefore, metrics that require exact matching are no longer suitable. Instead, each question is accompanied with 100 candidate answers. Consequently, the metric has been shifted from accuracy to retrieval-based metrics, \eg, MRR and NDCG. Prior works focus on optimizing a single metric \citep{guo2019image, jiang2019dualvd, hu2017learning, gan2019multi}. Differently,  \citet{murahari2019large} attempt to optimize both metrics with a joint loss. Still, a dedicated single metric model is superior. Instead, we propose principals to ensemble two dedicated models, one for NDCG and one for MRR. Our approach allows most of the MRR and NDCG to be preserved simultaneously. 

\noindent\textbf{Visual dialog models:}
Various approaches were proposed to solve the Visual Dialog task. Most of them focus on dialog history reasoning per interaction. \citet{serban2017hierarchical} propose history hierarchical encoding. \citet{seo2017visual} introduce a memory network based on attention, which also addressed co-referential issues. \citet{kottur2018visual} focus on visual co-reference.  \citet{jain2018two} concatenate representations of all the cues (\eg, image, question, history, and caption) per candidate answer. \citet{zheng2019reasoning} employ a graph structure learning. \citet{schwartz2019factor} propose a model, namely Factor Graph Attention (FGA),  that lets all entities (\eg, question-words, image-regions, answer-candidate, and caption-words) interact to infer an attention map for each modality. An ensemble of five FGA models achieves the state-of-the-art MRR performance. However, FGA optimizes using the sparse annotations, \ie, the human-derived answer. \citet{murahari2019large} recently propose Large-Scale(LS) model, which pre-trains on related vision-language datasets, \eg,  Conceptual Captions and Visual Question Answering\citep{sharma2018conceptual, VQA}. Concurrently, \citet{wang2020vd} leverage the pretrained BERT language models, and \citet{nguyen2019efficient} propose a lightweight Transformer that handles the interplay between many modalities. The three methods mentioned above finetune using the dense annotation (\ie, human assessment of all the candidates), resulting in a substantial improvement on the NDCG metric. Importantly, Murahari~\etal find that finetuning a model for NDCG  hurts MRR performance. This work demonstrates that re-ranking MRR model (\eg, FGA) and NDCG model (\eg, LS) with simple principles keeps most MRR and NDCG performance. 





\section{Two-step Rank Ensemble}
The MRR metric depends on a single human-derived answer. Hence, given that this answer is ranked highly,  the remaining candidates can be ranked according to the NDCG model. In the following, we describe two steps: (i) the MRR step responsible for preserving the human-derived rank high, and (ii) the NDCG step responsible for ranking the remaining candidates based on the NDCG model. 

\subsection{Setup}
We are given a set of dialog questions $\{(q, \mathcal{C}_q)_i\}_{i=1}^d$, where $d$ is the dataset size, $q$ is a dialog question, and $\mathcal{C}_q=\{c_{q,j} \}_{j=1}^{100}$ are the corresponding candidates. The MRR metric, \ie, the inverse harmonic mean of rank, is defined as:
\bea \label{eq:mrr}
 \mathrm {MRR} = \frac{1}{d}\sum_{i=1}^{d}\frac{1}{r_{\text{i}}},
\eea 
where $r_{i}$ is the rank of the human response for the $i$-th dialog question. The DCG, \ie, discounted cumulative gain over the $K$ correct answers, is defined as:
\bea
{ \mathrm {DCG_{K}} =\sum _{i=1}^{K}{\frac {s_{i}}{\log _{2}(i+1)}}, }
\eea 
where $s_{i}$ is a binary score, representing the fraction of annotators that marked the candidate at position as correct. We normalize by the ideal $\textstyle \mathrm {DCG_{K}}$ score ($\textstyle \mathrm {IDCG_{K}}$), \ie, $\textstyle {\mathrm  {NDCG_{{K}}}}={\mathrm{\frac{DCG_{{K}}}{IDCG_{{K}}}}}$.
We denote the set of MRR models as $ \mathcal{M} = \{M_1, \ldots, M_{n_m} \}$ where $n_m$ is the number of MRR models. Each MRR model is built by altering the initial conditions. We denote the NDCG model as $N$. 
We define an operator $\operatorname{T}(M,n,q)$ that returns the model $M$'s top $n$ responses given a question $q$.
Next, we describe the MRR step that aims to keep the MRR score.   

\subsection{MRR step}
\label{sec:mrr}
The purpose of the MRR step is to find a minimal candidate set $\mathcal{C}_{\text{MRR}, q}$ that is likely to contain the human-derived answer given a question $q$. We build this set as a union of three sets, as follows:
\vspace{-2px}\bea
\mathcal{C}_{\text{MRR}, q} = {\mathcal{T}_q \cup \mathcal{N}_q \cup \mathcal{H}_q},
\eea
where $\mathcal{T}_q$ is a set of first ranked candidates according to MRR models, $\mathcal{N}_q$ is a set of high ranked candidates by both MRR and NDCG models, $\mathcal{H}_q$ is a set of high-certainty candidates agreed by all the MRR models. All sets are conditioned by the question $q$. 
In the following, we formally define those sets.

\subsubsection{High-certainty answers}
One of the most significant signals to be the human-derived answer is being a top MRR-model's answer. However, in many subjective questions, the MRR model is not certain. We found that in those cases, the top answers often varies between different MRR models. Thus, to verify the top candidate's certainty, we require an agreement of MRR-models. Let $q$ be a dialog question, we define the \textit{high-certainty} set as follows: 
\bea
\mathcal{H}_q = \{c | \left(\forall M \in \mathcal{M}; c\in \operatorname{T}(M,\rho_h,q)\right)\},
\eea
where $\rho_h\in\mathbb{R}$ is an hyperparameter. Intuitively, a low $\rho_h$  results in higher certainty. We 
Next, we add the MRR-models' answer at first retrieval. 

\subsubsection{Top answers}

The MRR metric prioritizes the first-ranked answer (see Eq.\eqref{eq:mrr}). This property suits the nature of dialog models that reply with a single response. Consequently, we keep the first responses of the MRR models. Let $q$ be a dialog question,  the \textit{top-answers} set is defined as:
\bea
\mathcal{T}_q = \{c |  \left(\exists M \in \mathcal{M}; c\in \operatorname{T}(M,\rho_t,q)\right) \},
\eea
where $\rho_t\in\mathbb{R}$ is an hyperparameter.  We note that $\rho_t$ should be low to maintain candidates' certainty. In the next step, we consider top NDCG candidates. 

\subsubsection{NDCG-agreement answers}
When the NDCG model and the MRR model agree that a candidate is likely to be correct, it implies that both the NDCG and MRR metrics gain by ranking this candidate high. Thus, we want to rank it high. We note that the MRR set is ranked first, so we include these candidates in the MRR set.  Let $q$ be a dialog question, the \textit{ndcg-agreement} set is defined as: 
\bea
\Scale[0.78]{\mathcal{N}_q=\{c | \exists M\in\mathcal{M}; c\in \operatorname{T}(N,\rho^{n}_{n},q) \cap \operatorname{T}(M,\rho^{n}_{m},q)\}},
\eea
where $\rho^n_{n},\rho^n_{m}\in\mathbb{R}$  are hyperparameters that indicate relevancy to NDCG and MRR, respectively. \Ie, as $\rho_{nn}$ increases, we may include more relevant candidates according to the NDCG model. 

Up until this stage we have built a minimal set $C_{\text{MRR}, q}$ that is likely to hold the human-derived answer. In the following we describe how we rank this set.




\subsubsection{MRR ranking}
 
 Let  $r_{M_i, c, q}$ denote the rank according to $M_i\in\mathcal{M}$ of candidate $c$ for a question $q$. We compute the MRR rank of candidate $c\in \mathcal{C}_{\text{MRR}, q}$ via geometric mean: $ r_{\text{MRR},c, q} = \prod_{i=1}^{n_m} r_{M_i, c,q}$.


\begin{table}[t]
	\setlength{\tabcolsep}{5pt}
	\centering
	\resizebox{\linewidth}{!}{%
		\begin{tabular}{lcccccc}
			\hline
			Model    & MRR$\uparrow$     & R@1$\uparrow$   & R@5$\uparrow$    & R@10$\uparrow$   & Mean$\downarrow$  & NDCG$\uparrow$   \\ \hline
			NMN~\citep{kottur2018visual}                   & 58.80 & 44.15 & 76.88 & 86.88 & 4.81 & 58.10 \\
			NN~\citep{zheng2019reasoning}                   &  61.37 & 47.33 & 77.98 & 87.83 & 4.57 & 52.82 \\
            CorefNMN~\citep{kottur2018visual}             & 61.50 & 47.55 & 78.10 & 88.80 & 4.40 & 54.70  \\
            RvA~\citep{niu2019recursive}                      & 63.03 & 49.03 & 80.40 & 89.83 & 4.18 & 55.59 \\
            HACAN~\citep{yang2019making}                 & 64.22 & 50.88 & 80.63 & 89.45 & 4.20 & 57.17 \\
            MReal - BDAI$\ddagger$ ~\citep{qi2019two}    & 52.62 & 40.03 & 68.85 & 79.15 & 6.76 & 74.02  \\
            ReDAN~\citep{gan2019multi}                   & 53.13 & 41.38 & 66.07 & 74.50 & 8.91 & 61.86 \\
            ReDAN+$^\dag$~\citep{gan2019multi}          & 53.74 & 42.45 & 64.68 & 75.68 & 6.64 & 64.47  \\
            DualVD~\citep{jiang2019dualvd}                & 63.23 & 49.25 & 80.23 & 89.70 & 4.11 & 56.32  \\
            DL-61~\citep{guo2019image}                    & 62.20 & 47.90 & 80.43 & 89.95 & 4.17 & 57.32 \\
            DL-61$^\dag$~\citep{guo2019image}             & 63.42 & 49.30 & 80.77 & 90.68 & 3.97 & 57.88 \\
            DAN~\citep{kang2019dual}                   & 63.20 & 49.63 & 79.75 & 89.35 & 4.30 & 57.59  \\
            DAN$^\dag$~\citep{kang2019dual}             & 64.92 & 51.28 & 81.60 & 90.88 & 3.92 & 59.36 \\
            LTMI~\citep{nguyen2019efficient}                   & 64.20 & 50.20 & 80.68 & 90.35 & 4.05 & 59.03  \\
            LTMI$^\dag\ddagger$~\citep{nguyen2019efficient}    & 52.14 & 38.93 & 66.60 & 80.65 & 6.53 & 74.88 \\            
			FGA~\citep{schwartz2019factor} & 63.75 & 49.58 &80.97 & 88.55 &   4.51 & 52.12	\\
			5$\times$FGA$^\dag$~\citep{schwartz2019factor} & 69.37 & 55.65 &  86.73 & 94.05 & 3.14 &  57.29 	\\ 
			LS(CE)*$\ddagger$ ~\citep{murahari2019large}   & 50.74 & 37.95 & 64.13 & 80.00 & 6.28 & 74.47 \\
			LS(CE+NSP)*$\ddagger$ ~\citep{murahari2019large}  &  63.92 & 50.78 & 79.53 & 89.60 & 4.28 & 68.08 \\
			LS*~\citep{murahari2019large}  & 67.50 & 53.85 & 84.68 & 93.25 & 3.32 & 63.87 \\
			VD-BERT*$\dag\ddagger$ ~\citep{wang2020vd} & 51.17 & 38.90 & 62.82 & 77.98 & 6.69 & \textbf{75.35} \\
			\hline
	        5xFGA + LS*$\dag$  &	\textbf{71.24} & \textbf{58.28}  &	\textbf{87.55}  &	\textbf{94.45}  &	\textbf{2.96}  & 64.04 \\
	        5xFGA + LS + LS(CE)*$\dag\ddagger$   &	68.78 & 55.72  & 85.02  &	93.55  &	3.26  & 69.22 \\
  		    Ours*$\dag\ddagger$ & 70.41 & 58.18 & 83.85 & 90.83 & 3.66 &  72.16	\\
  		    \hline
  		    \multicolumn{7}{c}{Visual Dialog Challenge 2020 Leaderboard} \\
  		    \hline
	        LS & 68.79 & 55.20 & \underline{86.15} & \underline{93.88} & \underline{3.12} & 63.34 \\
  		    VD-BERT	& 51,84 & 39.91 & 63.45 & 78.56 & 6.57 & \underline{75.92} \\
	        MReaL Lab~(3\textsuperscript{rd} Place)   & 64.12 & 50.81 & 80.03 & 90.92 & 3.83 & 75.70 \\
	        SES-100M~(2\textsuperscript{nd} Place)   & 63.84 & 55.62 & 72.20 & 83.70 & 5.84 & 75.86 \\
	        \hline
	        Ours~(1\textsuperscript{st} Place) & \underline{70.42} & \underline{58.59} & 82.85 & 88.84 & 3.96 &  73.35	\\
			\hline
	    \end{tabular}}
	\caption{Performance on VisDial v1.0 test-std. (*) denotes the use of external knowledge. ($\dag$) indicates ensemble model, and ($\ddagger$) signifies fine-tuning on the dense annotations. Shown are the MRR, NDCG, the mean rank of the human-derived answer, and the recall at a certain number of retrievals.
    }
	\label{baselines}
\end{table}
\begin{table}[t]
	\setlength{\tabcolsep}{10pt}
    \centering
	\resizebox{\linewidth}{!}{%
        \begin{tabular}{cccccccccc} 
    		\hline
       		$\mathcal{H}$            & $\mathcal{T}$            & $\mathcal{N}$       & MRR$\uparrow$     & R@1$\uparrow$   & R@5$\uparrow$    & R@10$\uparrow$   & Mean$\downarrow$  & NDCG$\uparrow$  &  $|\mathcal{C}_{\text{MRR}}|$ \\ 
       		\hline
       		\cmark    & \xmark    & \xmark         & 70.83   & 58.87 & 84.32 & 90.67 & 3.67 & 74.32 & 2.34 \\
    		\xmark    & \cmark    & \xmark       & 68.63   & 59.12 & 79.08 & 88.53 & 4.15 & 74.31 & 1.12 \\
    		\xmark    & \xmark    & \cmark         & 61.75   & 51.74 & 70.35 & 85.88 & 4.94 & \textbf{74.69} & 3.97 \\ 
       		\xmark    & \cmark    & \cmark         & 69.21   & 59.17 & 78.68 & 88.53 & 4.11 &  74.33 & 4.27 \\
    		\cmark    & \xmark    & \cmark        & 71.15   & 59.11 & 84.38 & 90.67 & 3.65 & 73.29 & 4.87 \\
    		\cmark    & \cmark    & \xmark         & 71.06   & 59.15 & 84.49 & 90.78 & 3.64 & 72.98 & 2.39 \\
    		\cmark    & \cmark    & \cmark        & \textbf{71.26} & \textbf{59.18} & \textbf{84.62} & \textbf{90.78} & \textbf{3.62} &  73.23 & 4.91  \\
    		\hline
            \multicolumn{3}{l}{LS(CE)} & 52.21 & 39.92 & 65.04 & 80.62 & 6.16 &  \underline{75.24} & - \\
            \multicolumn{3}{l}{LS} & 69.00 & 55.80 & 85.36 & 93.13 & 3.35 &  64.89 & - \\
       		\multicolumn{3}{l}{5xFGA} & 69.38 & 56.17 & 86.15 & 92.95 & 3.32 &  58.68  & - \\
            \multicolumn{3}{l}{5xFGA + LS} & \underline{72.25} & \underline{59.20} & \underline{88.55} & \underline{94.52} & \underline{2.84} &  65.34 & - \\
            \multicolumn{3}{l}{5xFGA + LS + LS(CE)} & 69.14 & 56.79 & 84.24 & 92.37 & 3.43 &  73.78 & - \\

    		\hline


    		\hline
        \end{tabular}
    }
    \caption{MRR candidates set ablation analysis. Performance reported on the VisDialv1.0 val set. }
    \label{tab:ablation}
\end{table}

\subsection{NDCG step}
In this step, we rank the remaining candidates $\mathcal{C}_{\text{NDCG}, q} = \mathcal{C}_{q} \setminus \mathcal{C}_{\text{MRR}, q}$. We assume the correct MRR answer is in $\mathcal{C}_{\text{MRR}}$. Thus, we rank the remaining candidates, according to the NDCG model via geometric mean: $r_{\text{NDCG},c,q} =  (r_{N, c, q})^p\cdot r_{M, c, q},$ where  $M\in\mathcal{M}$ is the most accurate MRR model, and $p\in \mathbb{R}$ is a calibration hyperparameter which controls the trade-off between MRR and NDCG.  

To conclude, let $q$ be a dialog question and $\mathcal{C}_q$ the corresponding candidates. We first find $\mathcal{C}_{\text{MRR}, q}$, and rank the set according to $ r_{\text{MRR},c, q}$. We then rank the remaining candidates, according to $r_{\text{NDCG},c,q}$.

\begin{figure}[t]
	\centering
	\includegraphics[width=0.7\linewidth]{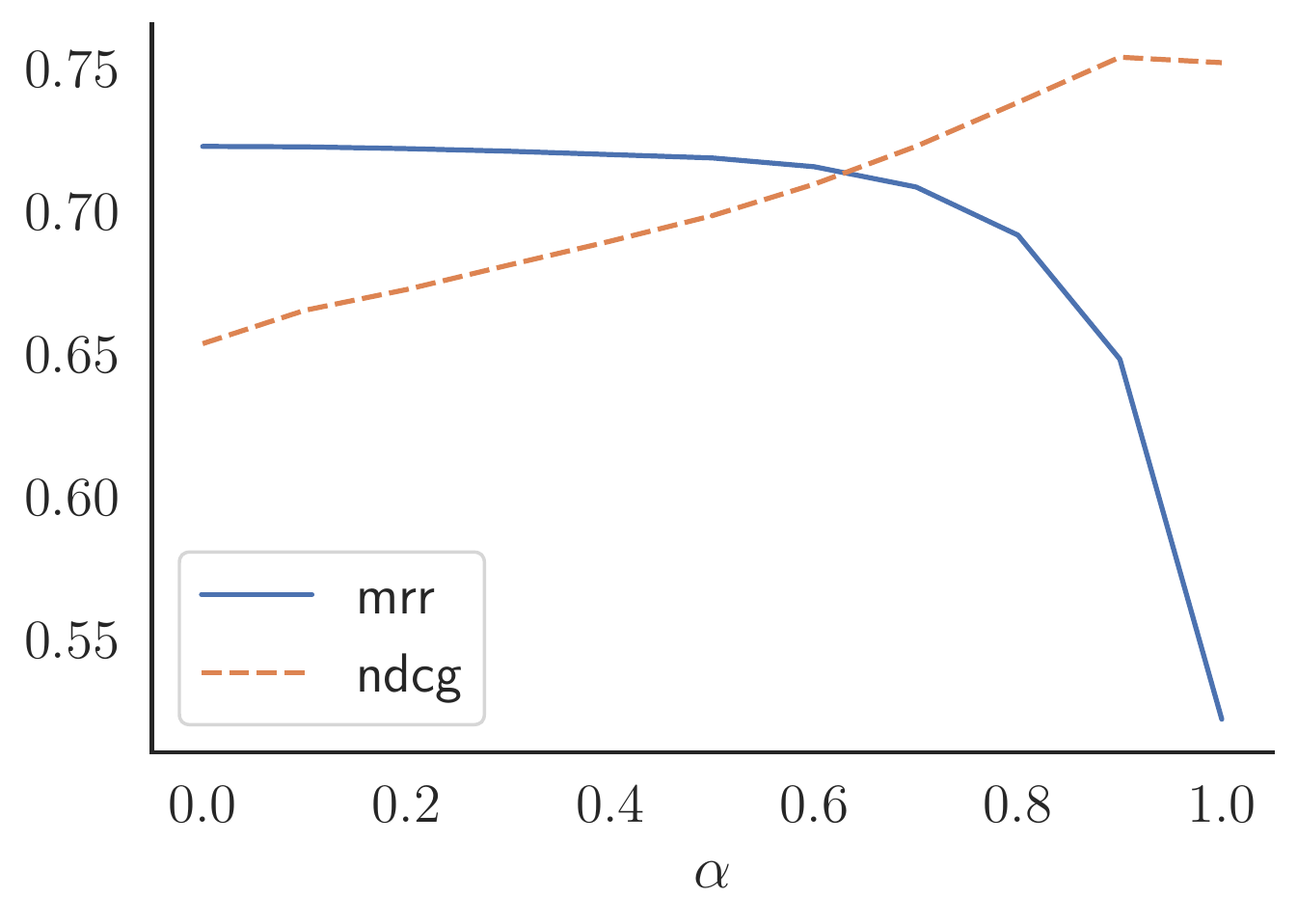}
	 \caption{Performance of a na\"ive score ensemble of the MRR model and the NDCG model  on the VisDialv1.0 val set. We calibrate the importance of each model with a scalar $\alpha$.}
	\label{fig:mrrndcg}
\end{figure}

\begin{figure*}
        \includegraphics[width=0.196\textwidth]{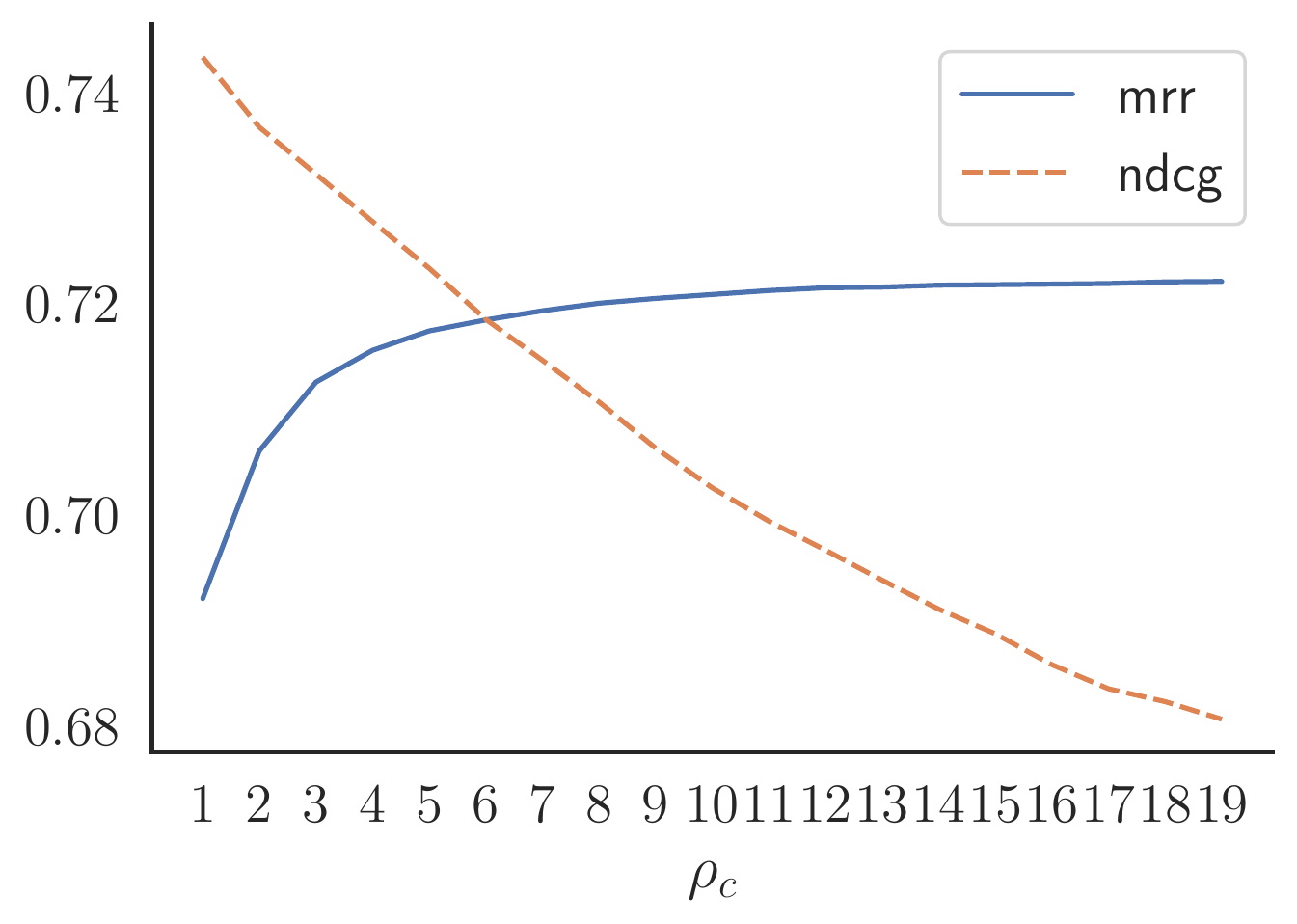}
        \includegraphics[width=0.196\textwidth]{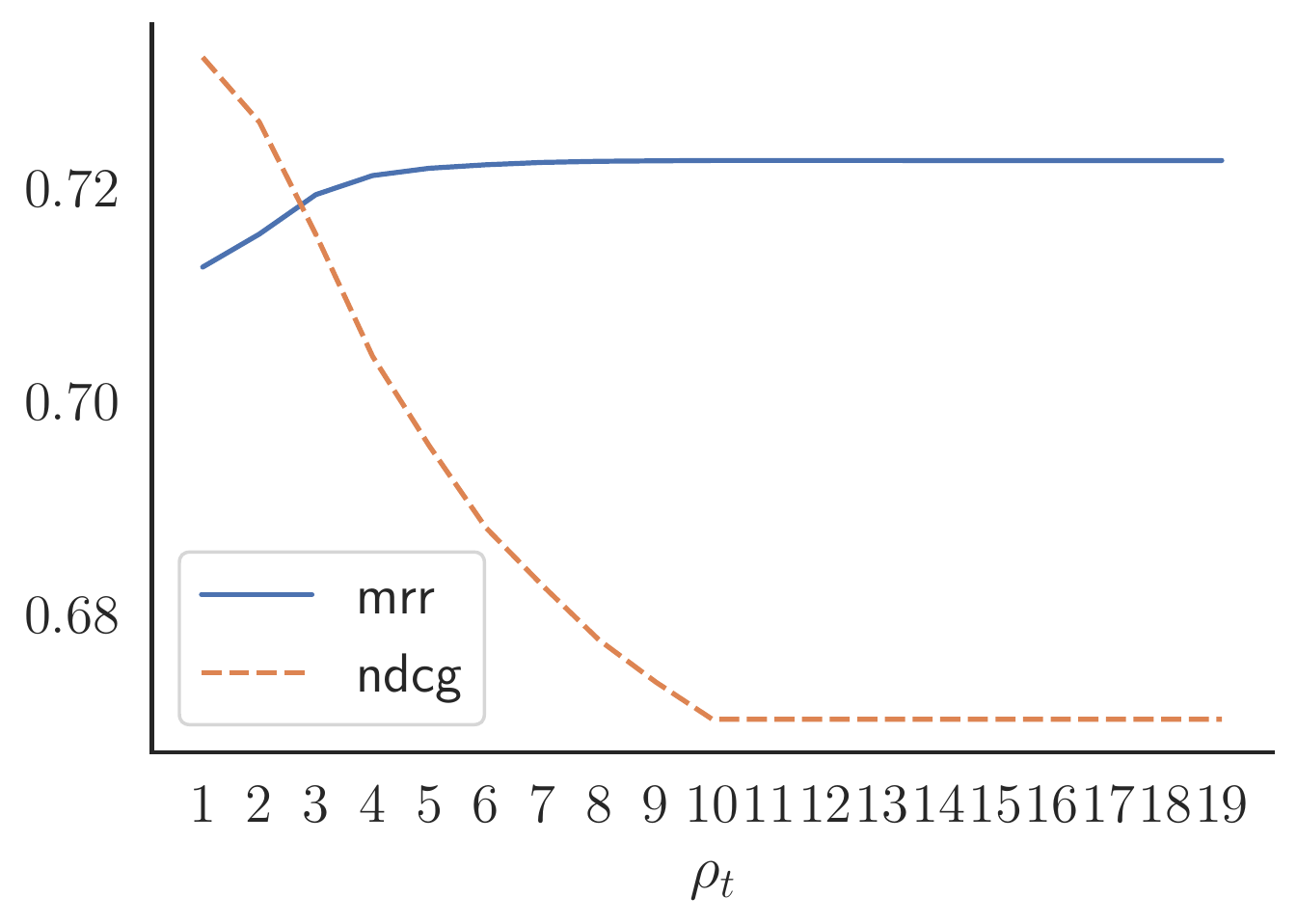}
        \includegraphics[width=0.196\textwidth]{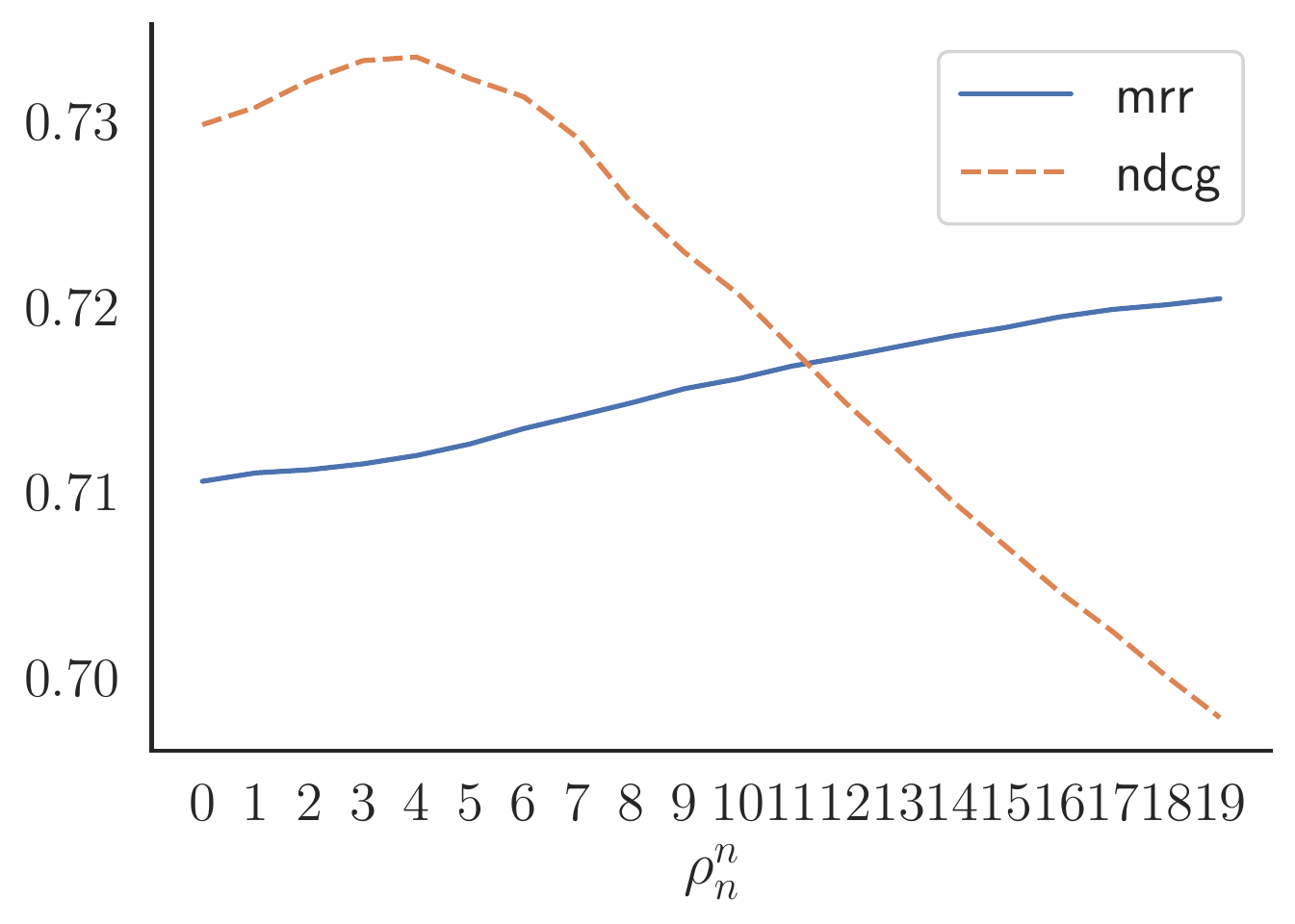}
        \includegraphics[width=0.196\textwidth]{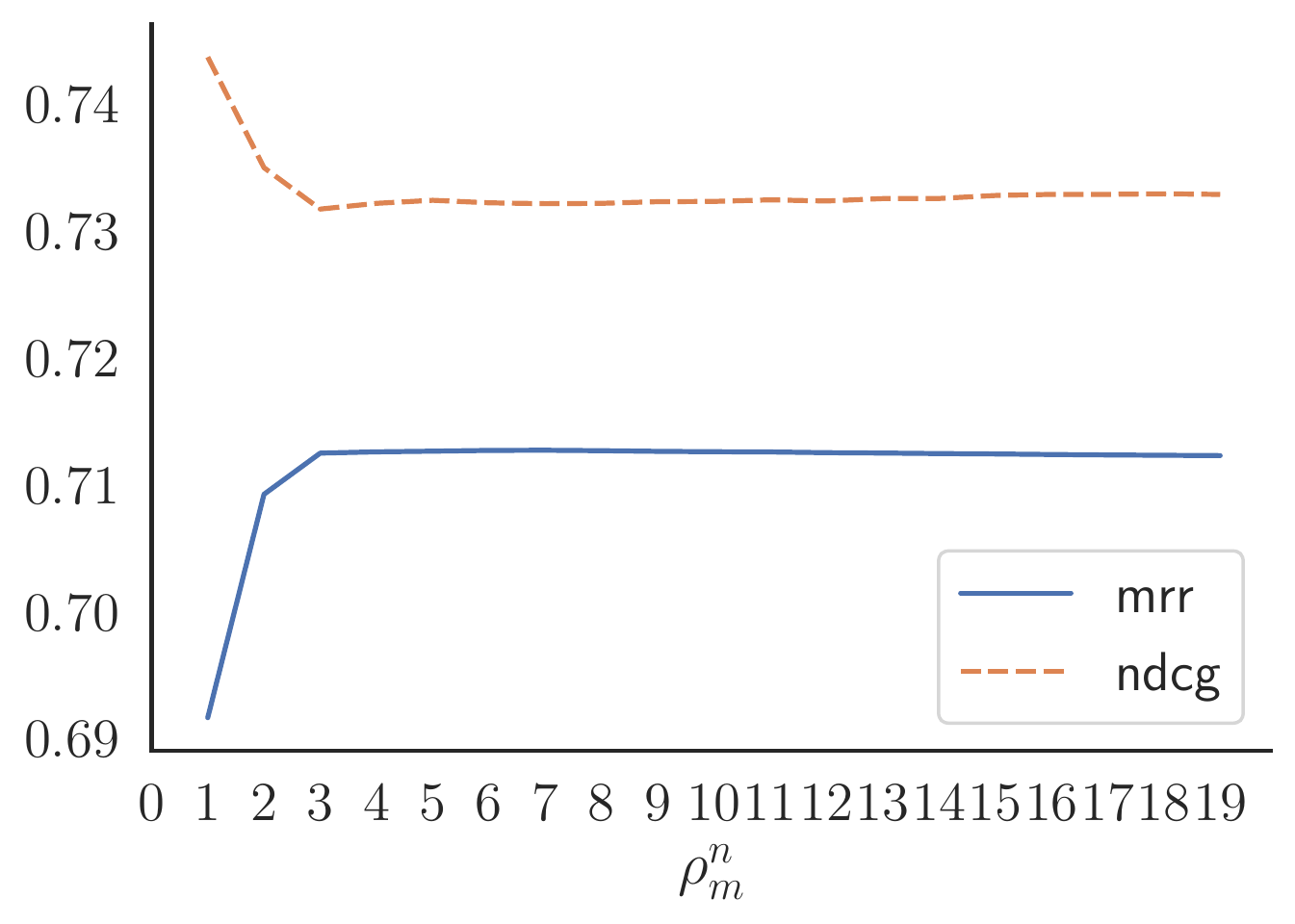}
        \includegraphics[width=0.196\textwidth]{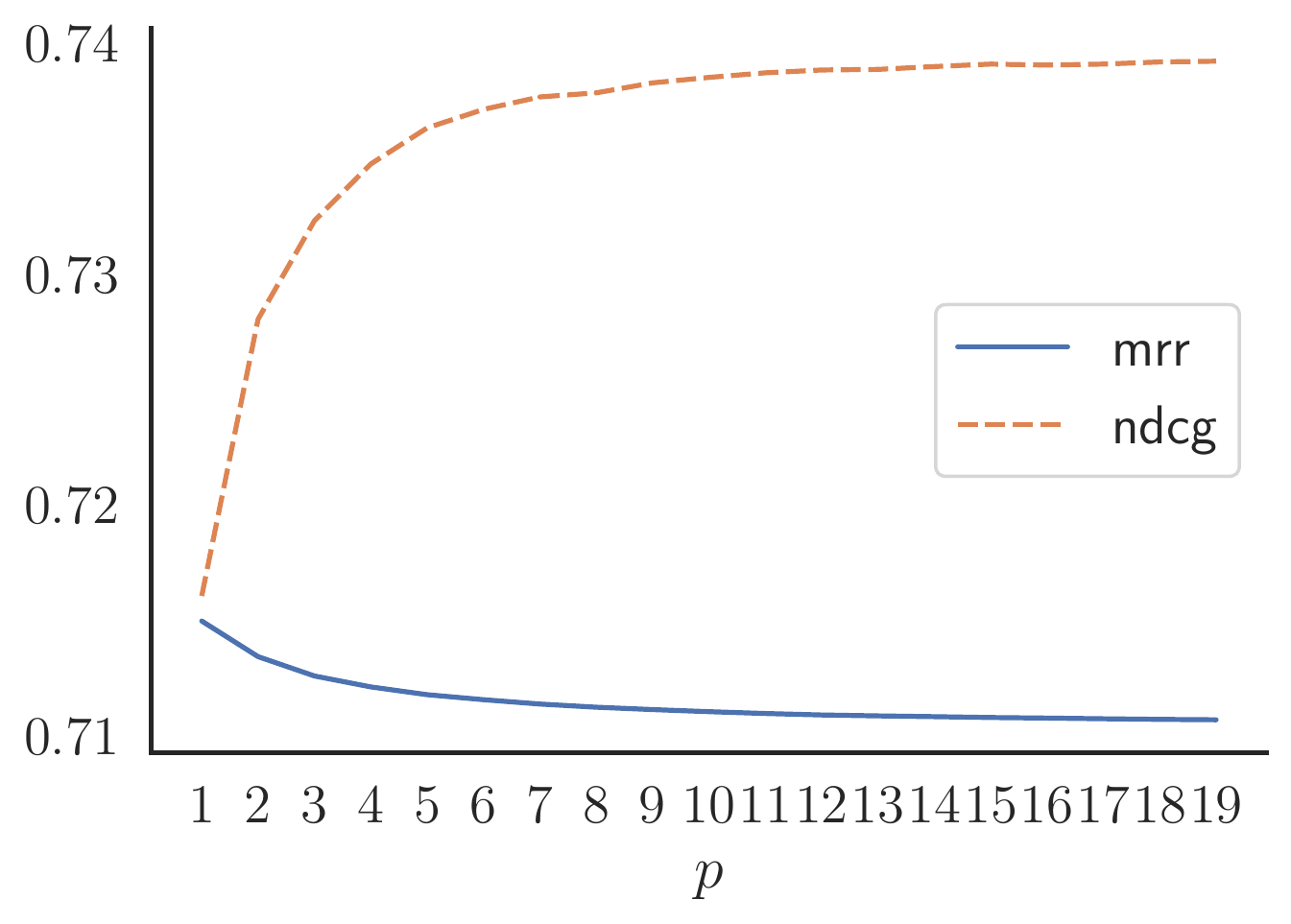}
        \caption{MRR and NDCG scores for different hyperparameter values.} 
        \label{fig:hyperparam}
    \end{figure*}
\section{Results}
We show our results on the VisDial v1.0 dataset, where 123,287 images are used for training, 2,000 images for validation, and 8,000 images for testing~\citep{visdial}. Each image is associated with ten questions, and each question has 100 corresponding answer candidates. We use two MRR models (\ie, $n_m=2$), FGA~\citep{schwartz2019factor} and an ensemble of LS~\citep{murahari2019large} with FGA. We use LS(CE) as the NDCG model. We set $\rho_h=3$, $\rho_t=1$, $\rho^{n}_{n}=5$, $\rho^{n}_{m}=10$, and $p=3$. We tune these parameters using the
validation set.

\noindent\textbf{Comparison to state-of-the-art}: In Tab.~\ref{baselines} we compare our method to na\"ive ensembles and previous baselines.  We first ensemble the LS's output with the FGA's output. By combining them, we achieve the new MRR state-of-the-art (71.24\% \vs 69.37\%). Next, we build a na\"ive ensemble of the MRR model and the NDCG model. We do so by adding the MRR ensemble scores (denoted by $\mathcal{S}_M$) and LS(CE) scores (denoted by $\mathcal{S}_N$), as follows:  $\alpha\cdot\mathcal{S}_M + (1-\alpha)\mathcal{S}_N$, where $\alpha\in\mathbb{R}$ calibrates the trade-off between MRR and NDCG performance. We show in \figref{fig:mrrndcg} an analysis of different $\alpha$ values on the validation set. In Tab.~\ref{baselines}, we report results for $\alpha=0.8$. Our two-step method outperforms the MRR (70.41\% \vs 68.78\%) and NDCG (72.16\% \vs 69.22\%) metrics, despite lacking the output scores and only requiring rankings.

We also compare our approach to previous baselines. Most methods use the sparse annotations, \ie, the human-derived answer, while MReal-BDAI, VD-BERT, and LS(CE) finetune using the dense annotations. Finetuning with the dense annotations tremendously boosts the NDCG performance but loses MRR performance. The MRR performance decline can be attributed to NDCG being biased toward uncertain answers. We also note that LS leverages large-scale image-text corpora. LS(CE+NSP) optimizes both the dense and sparse annotations but still suffers from a performance drop compared to metric-dedicated LS models, \ie, MRR (63.92\% \vs 67.50\%) and NDCG (68.08\% \vs 74.47\%). Unlike the method mentioned above, our method re-rank the candidates based on two distinct models, with two distinct steps, to keep the human-derived answer high. In doing so, we achieve a good MRR performance (70.41\% \vs 71.24\%), yet notably with limited NDCG drop (72.25\% \vs 75.35\%). This property comes in handy in the recent Visual Dialog challenge, where the winners were picked based on both the NDCG and MRR evaluation metrics. Our method performs well on both metrics simultaneously and won the challenge.  

\noindent\textbf{Ablation analysis}: The MRR candidate set consists of different subsets. In \tabref{tab:ablation} we show the influence of each of subset independently on the retrieval metrics. Further, omitting a subset harms the performance, \ie each component is essential to preserve both the MRR and NDCG metrics. We also report the average size of the MRR-candidate set, and the validation performance of the MRR model (\ie, 5xFGA) and the NDCG model (\ie, LS(CE)). In addition we provide the results of the MRR ensemble, and the na\"ive NDCG and MRR ensemble for $\alpha=0.8$.

In \figref{fig:hyperparam}, we examine how the NDCG and MRR metrics are affected by modifying one hyperparameter while maintaining the others. On the first figure from the left, we alter $\rho_c$. The higher $\rho_c$, we require higher agreement between the MRR models, resulting in higher certainty for elements in the MRR set.  Because the MRR models are responsible for the MRR set ranking, an MRR set that is too large hurts the NDCG metric. For the same reason, in the second image from the left, increasing $\rho_t$,  significantly harms the NDCG performance. In the third figure from the left, we show that considering more candidates that both NDCG and MRR models agree upon (\ie, increasing $\rho^n_n$) helps both metrics' performance. However, adding too many candidates harms the NDCG metric. In the fourth image from the left, we show that the performance remains stable when $\rho^n_m$ is larger than three. Last, on the fifth image from the left, we show the effect of changing $p$, which calibrates the trade-off between MRR and NDCG during the NDCG ranking step. 

\begin{figure}[t]
	\centering
	\includegraphics[width=1\linewidth]{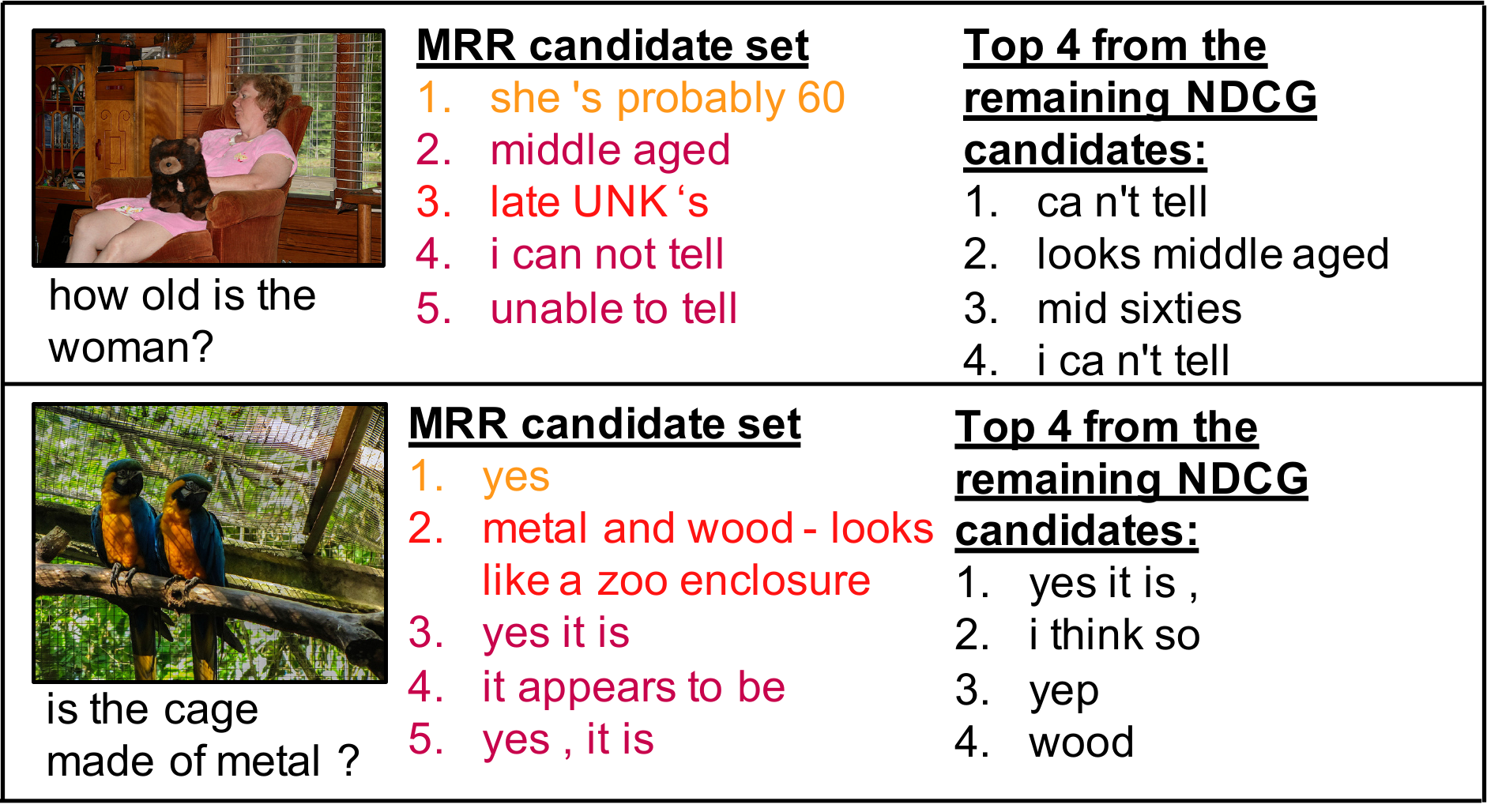}
	\caption{An illustration of two visual dialog samples. Each sample includes the MRR candidate set and four answers from the remaining NDCG candidates. We find that the MRR candidate set has more certain answers.  We colorize the \textit{high-certainty} candidates ($\mathcal{H}$) with \textcolor{orange}{orange}, the  \textit{NDCG-agreement} candidates ($\mathcal{N}$)  with \textcolor{purple}{purple}, and the \textit{top-answers} subset  ($\mathcal{T}$)  with \textcolor{red}{red}. Note, if a candidate belongs to more than one set, we sketch the colors in the following order: \textcolor{orange}{orange}$\to$\textcolor{red}{red}$\to$\textcolor{purple}{purple}. }
	\label{fig:qual}
\end{figure}

\noindent\textbf{Qualitative analysis}: In \figref{fig:qual}, we show two sample visual dialogs from test-std. For each sample, we provide the ranked MRR candidate set and the next 4 NDCG candidates.  The analysis reveals the answers' ambiguity and that the MRR candidate set mostly consists of certain responses.   In addition, we highlight the candidates within each MRR candidate subset with different colors.  Additional samples can be found in the appendix. 


\section{Conclusions}
We describe a non-parametric method to ensemble the candidate ranks of two strong MRR and NDCG models into a single ranking that excels on both NDCG and MRR. Intuitively,  we use the MRR-model for non-ambiguous questions with certain answers. The dense-annotations cue is more applicable in ambiguous questions than the sparse annotations. Thus, in the case of low certainty, our method relies almost entirely on the NDCG model. We hope the proposed principles can guide the community towards a parametric model that can employ answers' semantics to measure certainty.   
 
 \section*{Acknowledgements} We thank Yftah Ziser, Itai Gat, Alexander Schwing and Tamir Hazan for useful discussions. 
\bibliography{eacl2021}
\bibliographystyle{acl_natbib}
\appendix
\input{naacl2021_new.tex}

\end{document}

%% file: naacl2021_new.tex
\section{Qualitative Analysis}
In the following, we show 200 randomly picked visual dialog samples from test-std. For each sample, we provide the ranked MRR candidate set and the next 10 NDCG candidates.  The analysis reveals the answers' ambiguity and that the MRR candidate set mostly consists of the human-derived response.   To further detail our approach, we illustrate the candidates within each  MRR candidate subset by highlighting the candidates with different colors. We colorize the \textit{high-certainty} candidates ($\mathcal{H}$) with \textcolor{orange}{orange}, the  \textit{NDCG-agreement} candidates ($\mathcal{N}$)  with \textcolor{purple}{purple}, and the \textit{top-answers} subset  ($\mathcal{T}$)  with \textcolor{red}{red}. Note, if a candidate belongs to more than one set, we sketch the colors in the following order: \textcolor{orange}{orange}$\to$\textcolor{red}{red}$\to$\textcolor{purple}{purple}. 
 
 \includegraphics[width=1\linewidth]{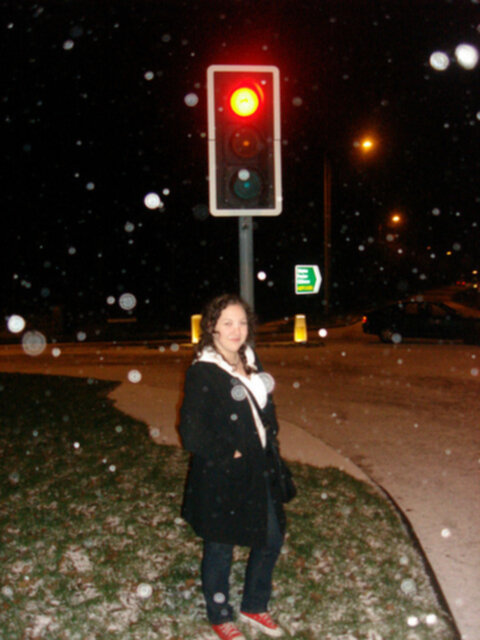}
 
 \noindent\textbf{Question:}what color is the light ?

 \noindent\textbf{MRR candidate set}
 \begin{enumerate}
 	\item ﻿\textcolor{orange}{red                   }\item ﻿\textcolor{purple}{green                   }\item ﻿\textcolor{purple}{the light is currently green               }\item ﻿\textcolor{purple}{the lights are n't visible               }\end{enumerate}
 \noindent\textbf{Top 10 from the remaining NDCG candidates}
 \begin{enumerate}
 	\item yellow                   \item not sure                  \item i ca n't tell                \item it 's kinda dark out so i am unsure           \item i can not tell                \item ca n't tell                 \item black and red                 \item UNK                   \item black                   \item it 's dark                 \end{enumerate}
 ﻿\includegraphics[width=1\linewidth]{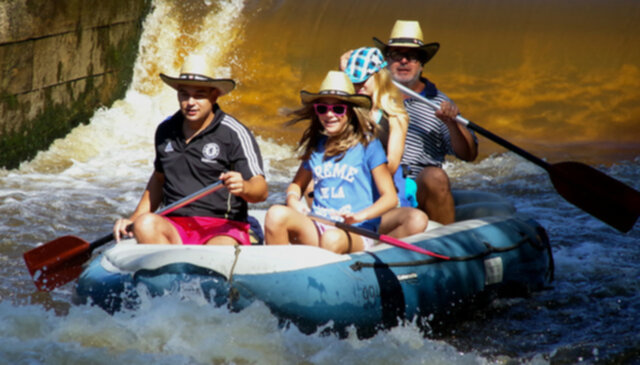}
 
 \noindent\textbf{Question:}what ethnicity do they appear to be ?

 \noindent\textbf{MRR candidate set}
 \begin{enumerate}
 	\item ﻿\textcolor{orange}{all white                  }\item ﻿\textcolor{orange}{white                   }\item ﻿\textcolor{purple}{i can not tell                }\item ﻿\textcolor{purple}{ca n't tell                 }\item ﻿\textcolor{purple}{not sure                  }\end{enumerate}
 \noindent\textbf{Top 10 from the remaining NDCG candidates}
 \begin{enumerate}
 	\item black                   \item i ca n't tell                \item both have brown hair                \item there are 2 teens , and 1 adult            \item i think like 40 's maybe              \item maybe younger seniors                 \item dark brown                  \item no , the tv is not that visible            \item 1 looks older the other is quite young            \item she appears to be caucasian               \end{enumerate}
 ﻿\includegraphics[width=1\linewidth]{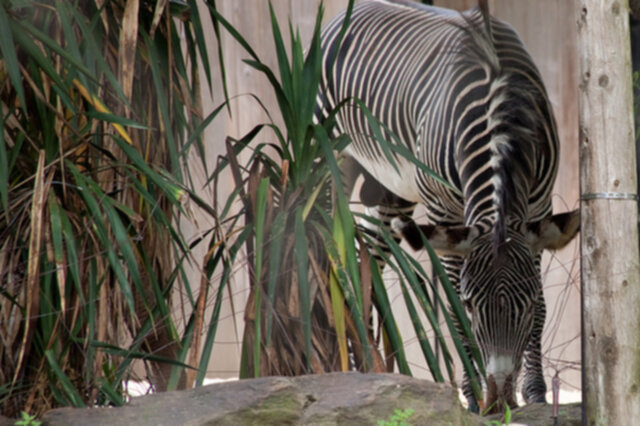}
 
 \noindent\textbf{Question:}are there any other animals ?

 \noindent\textbf{MRR candidate set}
 \begin{enumerate}
 	\item ﻿\textcolor{orange}{no                   }\item ﻿\textcolor{purple}{not that i see                }\item ﻿\textcolor{purple}{nope                   }\item ﻿\textcolor{purple}{not that i can see               }\end{enumerate}
 \noindent\textbf{Top 10 from the remaining NDCG candidates}
 \begin{enumerate}
 	\item no other animals                 \item 0 that i can see               \item 0                   \item 0 visible                  \item lol , no                 \item non                   \item i do n't think so               \item i can not see                \item no , only animals                \item i do not see any animals              \end{enumerate}
 ﻿\includegraphics[width=1\linewidth]{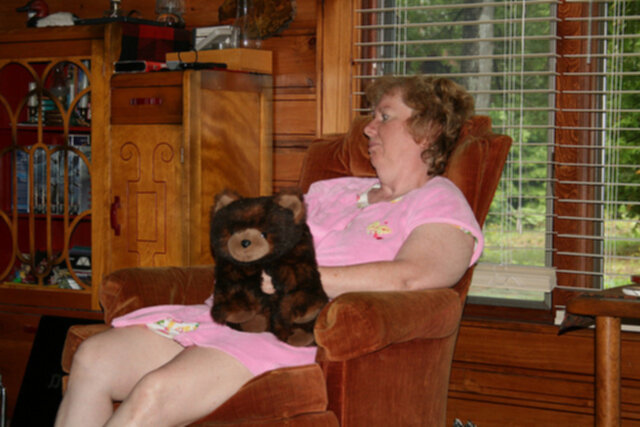}
 
 \noindent\textbf{Question:}how old is the woman ?

 \noindent\textbf{MRR candidate set}
 \begin{enumerate}
 	\item ﻿\textcolor{orange}{she 's probably 60                }\item ﻿\textcolor{purple}{middle aged                  }\item ﻿\textcolor{red}{late UNK 's                 }\item ﻿\textcolor{purple}{i can not tell                }\item ﻿\textcolor{purple}{unable to tell                 }\end{enumerate}
 \noindent\textbf{Top 10 from the remaining NDCG candidates}
 \begin{enumerate}
 	\item ca n't tell                 \item looks middle aged                 \item mid sixties                  \item i ca n't tell                \item late 40 's                 \item about middle aged maybe 40 something              \item 70s                   \item not sure                  \item about 30                  \item at least 30s                 \end{enumerate}
 ﻿\includegraphics[width=1\linewidth]{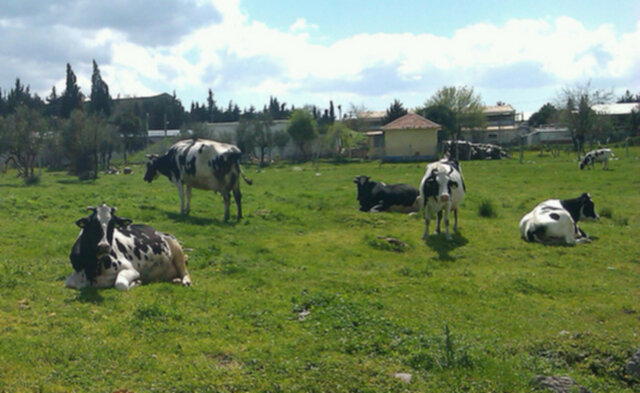}
 
 \noindent\textbf{Question:}do they all have any spots on them ?

 \noindent\textbf{MRR candidate set}
 \begin{enumerate}
 	\item ﻿\textcolor{orange}{yes , all but 1 has spots             }\item ﻿\textcolor{orange}{no                   }\item ﻿\textcolor{orange}{yes                   }\item ﻿\textcolor{purple}{not that i can see               }\item ﻿\textcolor{purple}{it 's too hard to tell              }\item ﻿\textcolor{purple}{nope                   }\end{enumerate}
 \noindent\textbf{Top 10 from the remaining NDCG candidates}
 \begin{enumerate}
 	\item i do n't think so               \item i can not tell                \item very few                  \item some do , yes                \item not really                  \item some do                  \item can not tell                 \item i ca n't tell                \item ca n't tell                 \item 0                   \end{enumerate}
 ﻿\includegraphics[width=1\linewidth]{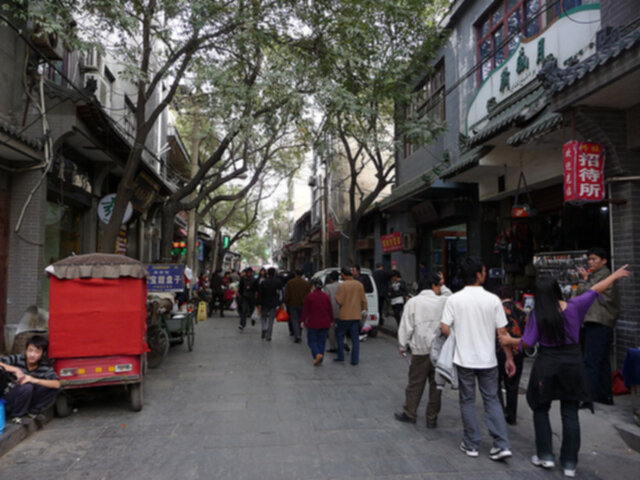}
 
 \noindent\textbf{Question:}see any store names ?

 \noindent\textbf{MRR candidate set}
 \begin{enumerate}
 	\item ﻿\textcolor{orange}{no                   }\item ﻿\textcolor{orange}{yes                   }\item ﻿\textcolor{orange}{i can not tell                }\item ﻿\textcolor{purple}{unable to tell                 }\item ﻿\textcolor{purple}{not that i can tell               }\item ﻿\textcolor{purple}{not that i can see               }\end{enumerate}
 \noindent\textbf{Top 10 from the remaining NDCG candidates}
 \begin{enumerate}
 	\item nope                   \item not really                  \item i ca n't see it               \item not sure                  \item i ca n't tell                \item ca n't tell                 \item probably just 1                 \item not at all                 \item yes 1                  \item i do n't think so               \end{enumerate}
 ﻿\includegraphics[width=1\linewidth]{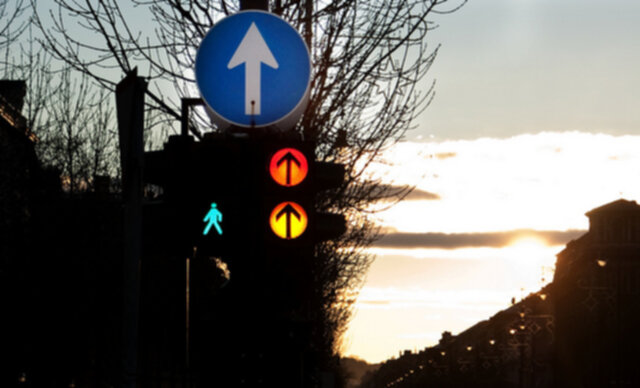}
 
 \noindent\textbf{Question:}what color is their hair ?

 \noindent\textbf{MRR candidate set}
 \begin{enumerate}
 	\item ﻿\textcolor{red}{i can not tell                }\item ﻿\textcolor{orange}{black                   }\item ﻿\textcolor{purple}{ca n't see them                }\item ﻿\textcolor{purple}{ca n't tell                 }\item ﻿\textcolor{red}{it doesn ’ t have hair it 's a light          }\item ﻿\textcolor{purple}{i ca n't tell                }\end{enumerate}
 \noindent\textbf{Top 10 from the remaining NDCG candidates}
 \begin{enumerate}
 	\item not sure                  \item dark is all i can tell              \item brown i believe                 \item hard to tell , but it looks very blackened           \item brown                   \item it looks brown                 \item no people                  \item dark brown                  \item it is dark brown                \item it 's brown                 \end{enumerate}
 ﻿\includegraphics[width=1\linewidth]{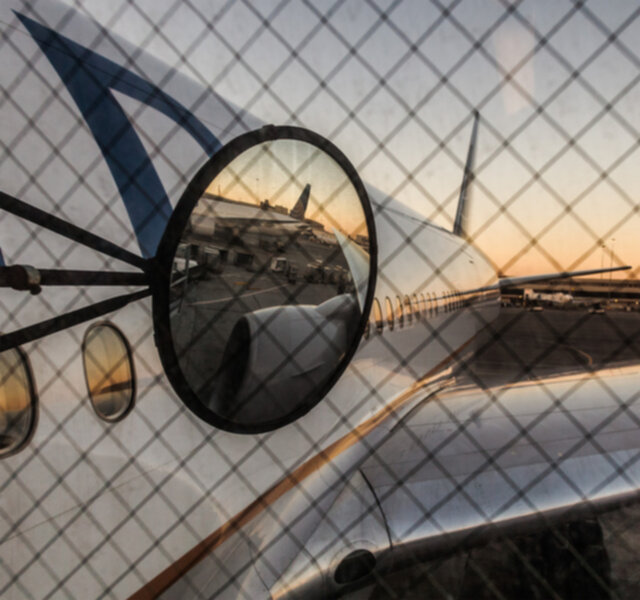}
 
 \noindent\textbf{Question:}is it daytime or night time ?

 \noindent\textbf{MRR candidate set}
 \begin{enumerate}
 	\item ﻿\textcolor{orange}{daytime                   }\item ﻿\textcolor{orange}{day                   }\item ﻿\textcolor{purple}{it appears to be daytime               }\item ﻿\textcolor{purple}{it is daytime                 }\item ﻿\textcolor{purple}{it looks like it 's daytime              }\item ﻿\textcolor{purple}{it 's daytime                 }\end{enumerate}
 \noindent\textbf{Top 10 from the remaining NDCG candidates}
 \begin{enumerate}
 	\item it is day time                \item day time                  \item it is day                 \item it is during the day               \item hard to tell probably day               \item looks like early morning                \item daytime maybe dusk                 \item hello , fellow turkey : yes , it is day time         \item in the even or morning               \item i ca n't tell                \end{enumerate}
 ﻿\includegraphics[width=1\linewidth]{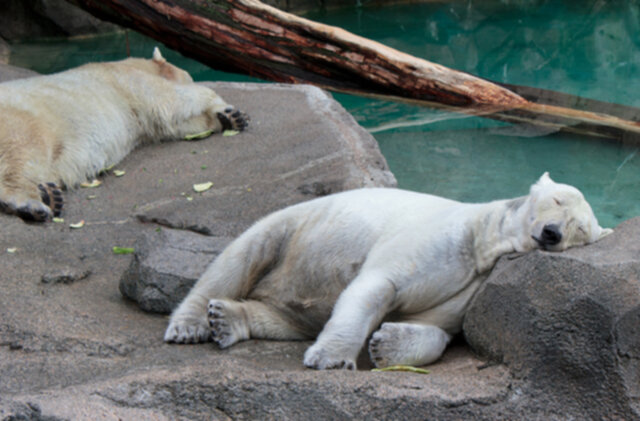}
 
 \noindent\textbf{Question:}is this photo in color ?

 \noindent\textbf{MRR candidate set}
 \begin{enumerate}
 	\item ﻿\textcolor{orange}{yes                   }\item ﻿\textcolor{orange}{yes it is                 }\item ﻿\textcolor{purple}{yes , it is                }\item ﻿\textcolor{purple}{yes it is in color               }\item ﻿\textcolor{purple}{yes , it 's in color              }\end{enumerate}
 \noindent\textbf{Top 10 from the remaining NDCG candidates}
 \begin{enumerate}
 	\item yes the picture is in color              \item yes in color                 \item it is                  \item it is in color                \item it 's in color                \item yes , it 's very colorful !             \item i think so                 \item looks like it                 \item it is , but there is not much color in the photo due to the objects    \item yes it is well lit               \end{enumerate}
 ﻿\includegraphics[width=1\linewidth]{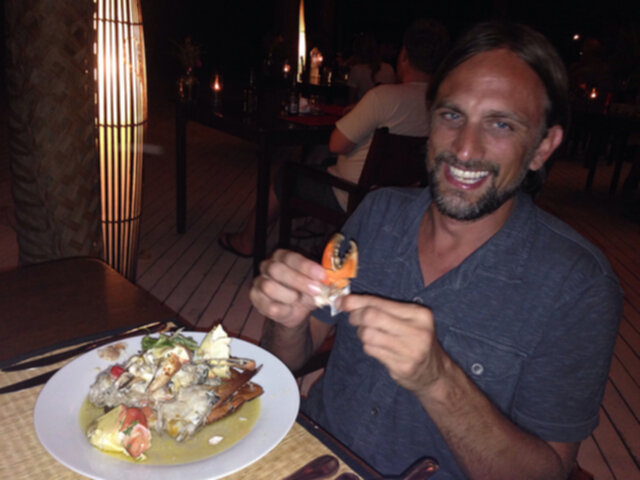}
 
 \noindent\textbf{Question:}what are the vegetables he has ?

 \noindent\textbf{MRR candidate set}
 \begin{enumerate}
 	\item ﻿\textcolor{orange}{salmon and greens , maybe spinach              }\item ﻿\textcolor{orange}{it looks like greens , hard to tell            }\item ﻿\textcolor{purple}{ca n't really tell                }\item ﻿\textcolor{purple}{i can not tell                }\item ﻿\textcolor{purple}{i ca n't tell                }\item ﻿\textcolor{purple}{ca n't tell                 }\end{enumerate}
 \noindent\textbf{Top 10 from the remaining NDCG candidates}
 \begin{enumerate}
 	\item not sure                  \item i 'm not a 100 \% sure , look like it         \item yes                   \item not that i can see               \item can not see grass                \item green                   \item plate                   \item looks old                  \item i do n't think so               \item nope                   \end{enumerate}
 	﻿\includegraphics[width=1\linewidth]{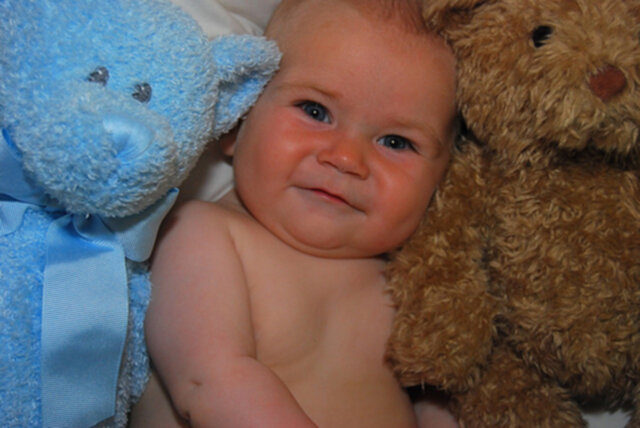}
 	
 	\noindent\textbf{Question:}are they all facing the camera ?

 	\noindent\textbf{MRR candidate set}
 	\begin{enumerate}
 		\item ﻿\textcolor{orange}{yes                   }\item ﻿\textcolor{orange}{no                   }\item ﻿\textcolor{orange}{slightly                   }\item ﻿\textcolor{purple}{yes they are                 }\item ﻿\textcolor{purple}{nope                   }\end{enumerate}
 	\noindent\textbf{Top 10 from the remaining NDCG candidates}
 	\begin{enumerate}
 		\item not really                  \item i don ’ t think so              \item yes they all are                \item yes i believe so                \item no side view                 \item i do n't think so               \item i think so                 \item all but 1                 \item not all of them                \item not that i can see               \end{enumerate}
 	﻿\includegraphics[width=1\linewidth]{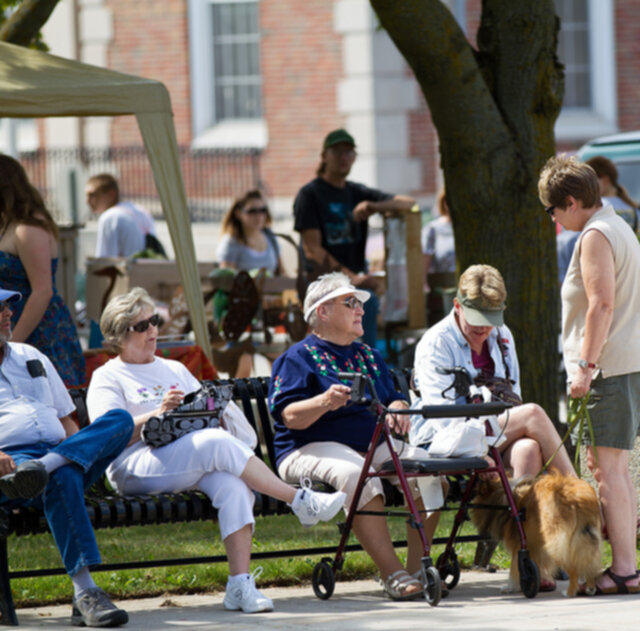}
 	
 	\noindent\textbf{Question:}is it a poodle ?

 	\noindent\textbf{MRR candidate set}
 	\begin{enumerate}
 		\item ﻿\textcolor{orange}{no                   }\item ﻿\textcolor{purple}{i do n't think so               }\item ﻿\textcolor{purple}{nope                   }\end{enumerate}
 	\noindent\textbf{Top 10 from the remaining NDCG candidates}
 	\begin{enumerate}
 		\item does not look like it               \item i ca n't tell                \item i can not tell                \item not sure                  \item i can not say                \item ca n't tell                 \item not that i can see               \item i do n't think so , but it 's hard to tell        \item i 'm not sure what that is             \item no i am unable to tell              \end{enumerate}
 
 	﻿\includegraphics[width=1\linewidth]{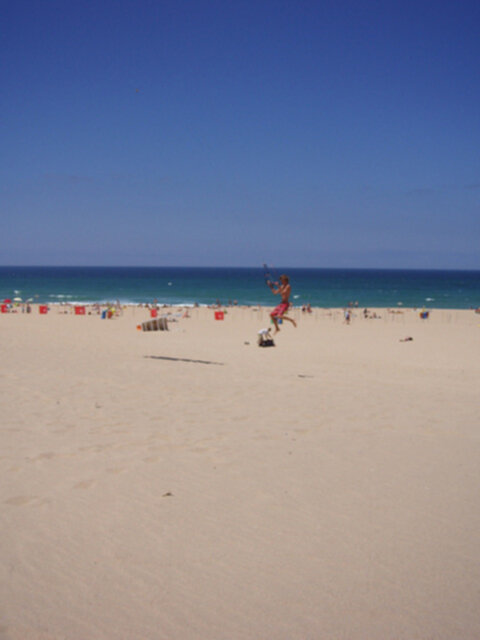}
 	
 	\noindent\textbf{Question:}how old is the man ?

 	\noindent\textbf{MRR candidate set}
 	\begin{enumerate}
 		\item ﻿\textcolor{red}{i can not tell                }\item ﻿\textcolor{purple}{can not tell                 }\item ﻿\textcolor{purple}{unable to tell                 }\item ﻿\textcolor{red}{ca n't see his face               }\item ﻿\textcolor{purple}{i ca n't tell                }\item ﻿\textcolor{purple}{ca n't tell                 }\end{enumerate}
 	\noindent\textbf{Top 10 from the remaining NDCG candidates}
 	\begin{enumerate}
 		\item i ca n't tell i ca n't see their face          \item twenties maybe , hard to tell              \item hard to tell but i would say mid 20 's          \item hard to say but i guess around thirty            \item not sure                  \item i can not see their face              \item maybe 25ish                  \item he looks to be about 30ish              \item it 's hard to tell , it 's from a distance         \item he looks to be maybe 20 or so            \end{enumerate}
 	﻿\includegraphics[width=1\linewidth]{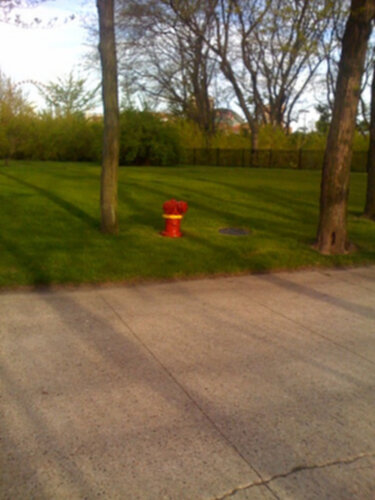}
 	
 	\noindent\textbf{Question:}is there anything near it ?

 	\noindent\textbf{MRR candidate set}
 	\begin{enumerate}
 		\item ﻿\textcolor{orange}{no                   }\item ﻿\textcolor{red}{grass and tree 's                }\item ﻿\textcolor{purple}{nothing                   }\item ﻿\textcolor{purple}{not that i can see               }\item ﻿\textcolor{purple}{nope                   }\item ﻿\textcolor{purple}{ca n't see anything                }\end{enumerate}
 	\noindent\textbf{Top 10 from the remaining NDCG candidates}
 	\begin{enumerate}
 		\item yes                   \item i do n't think so               \item i don ’ t see any              \item not really                  \item you can see a pitcher and maybe a wall           \item a few things                 \item no , it 's just dirt              \item in the distance                 \item no but it has a lot of stuff on it          \item not sure there may be , but it does n't show enough to tell      \end{enumerate}
 	﻿\includegraphics[width=1\linewidth]{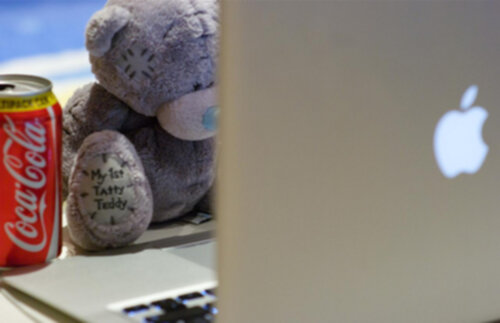}
 	
 	\noindent\textbf{Question:}is there anything on the wallpaper of the laptop ?

 	\noindent\textbf{MRR candidate set}
 	\begin{enumerate}
 		\item ﻿\textcolor{orange}{no                   }\item ﻿\textcolor{purple}{nope                   }\item ﻿\textcolor{purple}{not that i can see               }\end{enumerate}
 	\noindent\textbf{Top 10 from the remaining NDCG candidates}
 	\begin{enumerate}
 		\item no there is not                \item i can not tell                \item i do n't see any               \item ca n't tell                 \item i ca n't tell                \item 0                   \item i do n't think so               \item just a screen over it               \item not that i can UNK for sure             \item not sure                  \end{enumerate}
 	﻿\includegraphics[width=1\linewidth]{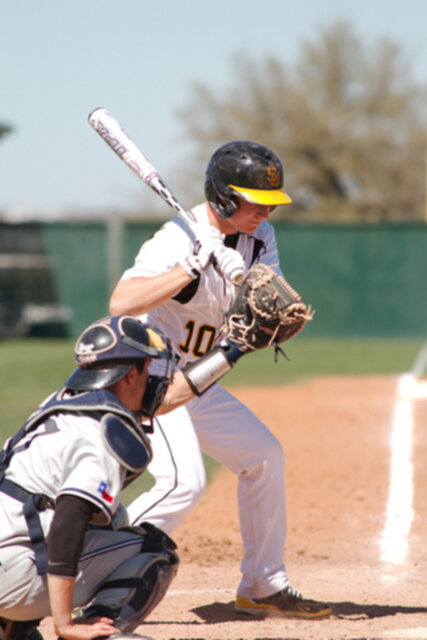}
 	
 	\noindent\textbf{Question:}is it date stamped ?

 	\noindent\textbf{MRR candidate set}
 	\begin{enumerate}
 		\item ﻿\textcolor{orange}{no                   }\item ﻿\textcolor{purple}{no it 's not                }\item ﻿\textcolor{purple}{no it is n't                }\item ﻿\textcolor{purple}{nope                   }\item ﻿\textcolor{purple}{no , it 's not               }\end{enumerate}
 	\noindent\textbf{Top 10 from the remaining NDCG candidates}
 	\begin{enumerate}
 		\item not that i can see               \item i do n't think so               \item i do n't know                \item do n't know                 \item i ca n't tell                \item ca n't tell                 \item not sure                  \item i can not tell                \item not readable                  \item nah                   \end{enumerate}
 	﻿\includegraphics[width=1\linewidth]{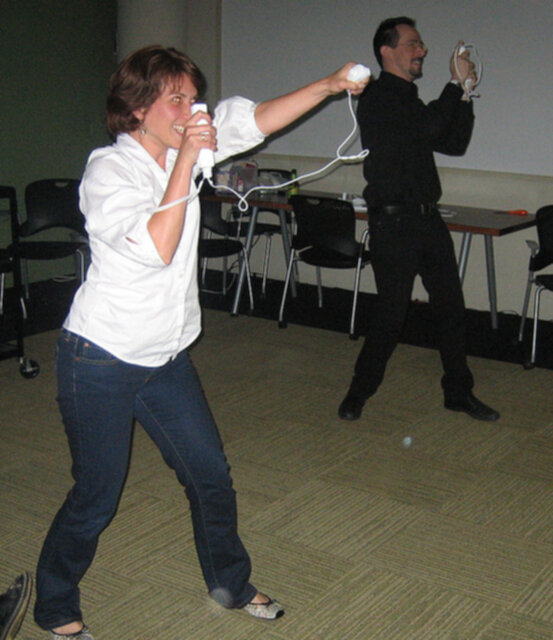}
 	
 	\noindent\textbf{Question:}is this inside ?

 	\noindent\textbf{MRR candidate set}
 	\begin{enumerate}
 		\item ﻿\textcolor{orange}{yes                   }\item ﻿\textcolor{orange}{yes it is                 }\item ﻿\textcolor{purple}{yes , it is                }\item ﻿\textcolor{purple}{yes i think so                }\end{enumerate}
 	\noindent\textbf{Top 10 from the remaining NDCG candidates}
 	\begin{enumerate}
 		\item yes , it seems to be              \item i think so , yes               \item this is inside                 \item it seems to be                \item it is indoors                 \item i believe so                 \item i think so                 \item looks like it                 \item no                   \item ues                   \end{enumerate}
 	﻿\includegraphics[width=1\linewidth]{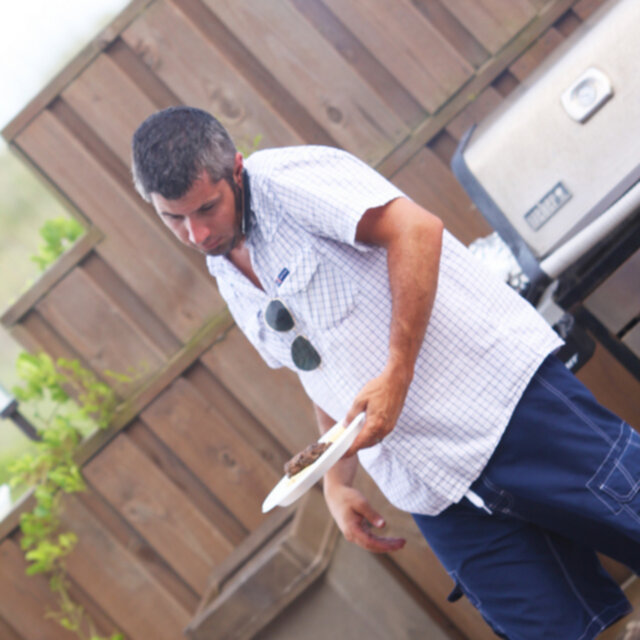}
 	
 	\noindent\textbf{Question:}does the food look appetizing ?

 	\noindent\textbf{MRR candidate set}
 	\begin{enumerate}
 		\item ﻿\textcolor{orange}{yes                   }\item ﻿\textcolor{orange}{no                   }\item ﻿\textcolor{purple}{not really                  }\item ﻿\textcolor{purple}{not at all                 }\item ﻿\textcolor{purple}{nope                   }\end{enumerate}
 	\noindent\textbf{Top 10 from the remaining NDCG candidates}
 	\begin{enumerate}
 		\item yes it does                 \item no , he does n't               \item yes , it does                \item i can not tell                \item ca n't tell                 \item a little                  \item not that i can see               \item sort of                  \item kinda                   \item i ca n't tell                \end{enumerate}
 	﻿\includegraphics[width=1\linewidth]{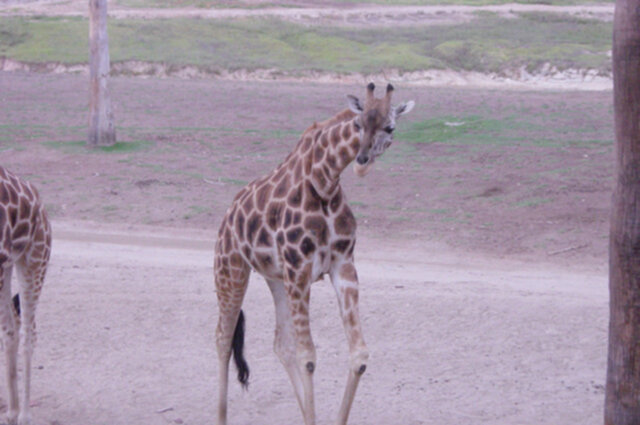}
 	
 	\noindent\textbf{Question:}is the image in color ?

 	\noindent\textbf{MRR candidate set}
 	\begin{enumerate}
 		\item ﻿\textcolor{orange}{yes                   }\item ﻿\textcolor{orange}{yes it is                 }\item ﻿\textcolor{purple}{yes , it is in color              }\item ﻿\textcolor{purple}{yes it is in color               }\item ﻿\textcolor{purple}{yes , it 's in color              }\item ﻿\textcolor{purple}{yes , the photo is in color             }\end{enumerate}
 	\noindent\textbf{Top 10 from the remaining NDCG candidates}
 	\begin{enumerate}
 		\item yes , it is                \item yes the photo is in color              \item yes , full color photo               \item yes it 's color                \item it is                  \item yes in color                 \item it is yes                 \item yess                   \item yea                   \item yyes                   \end{enumerate}
 	﻿\includegraphics[width=1\linewidth]{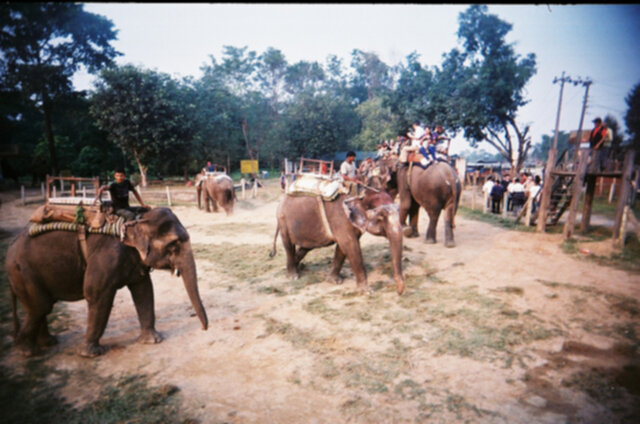}
 	
 	\noindent\textbf{Question:}is it daytime ?

 	\noindent\textbf{MRR candidate set}
 	\begin{enumerate}
 		\item ﻿\textcolor{orange}{yes                   }\item ﻿\textcolor{purple}{yes it is                 }\item ﻿\textcolor{red}{yes , it is daytime               }\item ﻿\textcolor{purple}{yes it is daytime                }\item ﻿\textcolor{purple}{yes , it is                }\end{enumerate}
 	\noindent\textbf{Top 10 from the remaining NDCG candidates}
 	\begin{enumerate}
 		\item it is                  \item yes , it looks to be              \item it is daytime                 \item yup                   \item it is day time                \item yes , i think so               \item i believe so                 \item yes it is , it looks sunny             \item daytime                   \item i think so                 \end{enumerate}
 	﻿\includegraphics[width=1\linewidth]{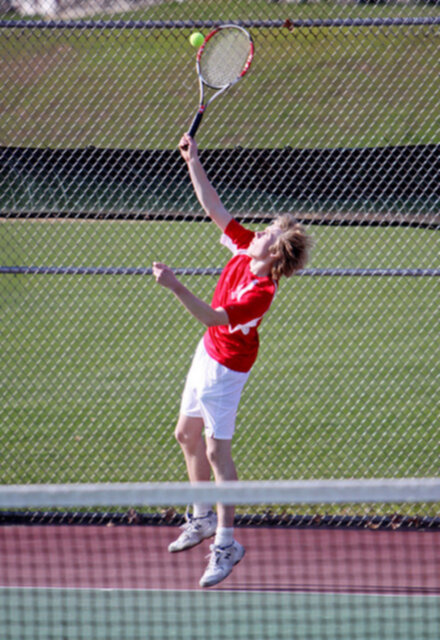}
 	
 	\noindent\textbf{Question:}is the photo in color ?

 	\noindent\textbf{MRR candidate set}
 	\begin{enumerate}
 		\item ﻿\textcolor{orange}{yes                   }\item ﻿\textcolor{orange}{yes it is                 }\item ﻿\textcolor{purple}{yes , it is in color              }\item ﻿\textcolor{purple}{yes it is in color               }\item ﻿\textcolor{purple}{yes it 's in color               }\end{enumerate}
 	\noindent\textbf{Top 10 from the remaining NDCG candidates}
 	\begin{enumerate}
 		\item yes , it is                \item yes the picture is in color              \item yeah , this photo is in color             \item yes in color                 \item it is                  \item it is in color                \item yup                   \item it is a beautiful color picture              \item ye                   \item y                   \end{enumerate}
 	﻿\includegraphics[width=1\linewidth]{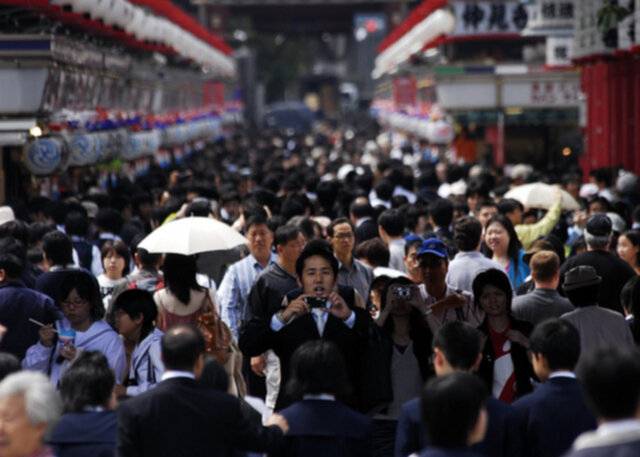}
 	
 	\noindent\textbf{Question:}any buildings visible ?

 	\noindent\textbf{MRR candidate set}
 	\begin{enumerate}
 		\item ﻿\textcolor{orange}{yes a few                 }\item ﻿\textcolor{orange}{yes                   }\item ﻿\textcolor{red}{yes , a few buildings               }\item ﻿\textcolor{purple}{yess                   }\item ﻿\textcolor{purple}{i think so                 }\end{enumerate}
 	\noindent\textbf{Top 10 from the remaining NDCG candidates}
 	\begin{enumerate}
 		\item yeah in the distance                \item lots                   \item in the distance                 \item yes there are some small ones              \item yes , 1 across the street              \item yes in the far background               \item yes , but i ca n't tell what kind           \item yes , it is                \item very far in the background               \item no                   \end{enumerate}
 	﻿\includegraphics[width=1\linewidth]{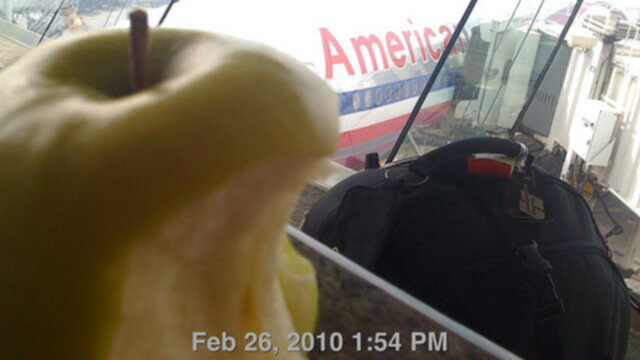}
 	
 	\noindent\textbf{Question:}is there anything made out of plastic ?

 	\noindent\textbf{MRR candidate set}
 	\begin{enumerate}
 		\item ﻿\textcolor{orange}{no                   }\item ﻿\textcolor{orange}{yes                   }\item ﻿\textcolor{purple}{nope                   }\item ﻿\textcolor{purple}{not that i can see               }\item ﻿\textcolor{purple}{i do n't think so               }\end{enumerate}
 	\noindent\textbf{Top 10 from the remaining NDCG candidates}
 	\begin{enumerate}
 		\item ca n't tell                 \item no that i can see               \item not sure                  \item ca n't see                 \item i ca n't tell                \item i can not tell                \item maybe                   \item i think so                 \item not really                  \item yes , looks like a box              \end{enumerate}
 	﻿\includegraphics[width=1\linewidth]{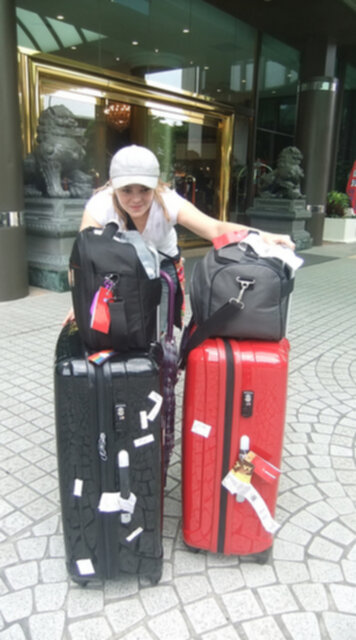}
 	
 	\noindent\textbf{Question:}how old is the woman ?

 	\noindent\textbf{MRR candidate set}
 	\begin{enumerate}
 		\item ﻿\textcolor{orange}{she looks like in her 30 's             }\item ﻿\textcolor{purple}{in her twenties i believe               }\item ﻿\textcolor{purple}{not sure                  }\end{enumerate}
 	\noindent\textbf{Top 10 from the remaining NDCG candidates}
 	\begin{enumerate}
 		\item 20 's                  \item early 20 's                 \item i ca n't tell                \item ca n't tell                 \item 20 's maybe                 \item looks to be late 20s               \item i can not tell                \item young adult                  \item 25                   \item early 30s maybe                 \end{enumerate}
 	﻿\includegraphics[width=1\linewidth]{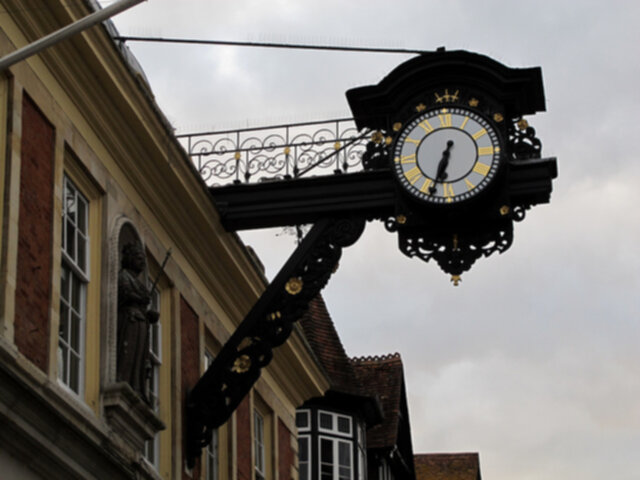}
 	
 	\noindent\textbf{Question:}is the image in color ?

 	\noindent\textbf{MRR candidate set}
 	\begin{enumerate}
 		\item ﻿\textcolor{orange}{yes                   }\item ﻿\textcolor{orange}{yes it is                 }\item ﻿\textcolor{orange}{yes , it is                }\item ﻿\textcolor{purple}{yes , it is in color              }\item ﻿\textcolor{purple}{yes it is in color               }\end{enumerate}
 	\noindent\textbf{Top 10 from the remaining NDCG candidates}
 	\begin{enumerate}
 		\item yes this is a color image              \item yes , it 's in color              \item it is yes                 \item yes this picture is in color              \item yes the photo is in color              \item yes in color                 \item yep                   \item it is in color                \item yes it 's color                \item ye                   \end{enumerate}
 	﻿\includegraphics[width=1\linewidth]{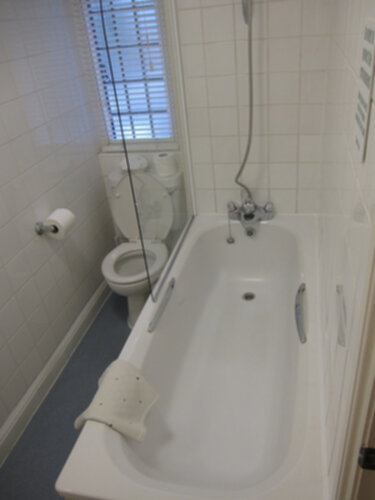}
 	
 	\noindent\textbf{Question:}is anyone in the bathroom ?

 	\noindent\textbf{MRR candidate set}
 	\begin{enumerate}
 		\item ﻿\textcolor{orange}{no                   }\item ﻿\textcolor{orange}{no the bathroom                 }\item ﻿\textcolor{purple}{not that i can see               }\item ﻿\textcolor{purple}{no people                  }\item ﻿\textcolor{purple}{nope                   }\item ﻿\textcolor{purple}{i do n't think so               }\end{enumerate}
 	\noindent\textbf{Top 10 from the remaining NDCG candidates}
 	\begin{enumerate}
 		\item no it appears empty                \item not it 's not visible               \item no he 's not                \item i would say no                \item no , there is no 1 in the kitchen           \item UNK                   \item don ’ t know                \item 0                   \item not really                  \item no people this is the sky              \end{enumerate}
 	﻿\includegraphics[width=1\linewidth]{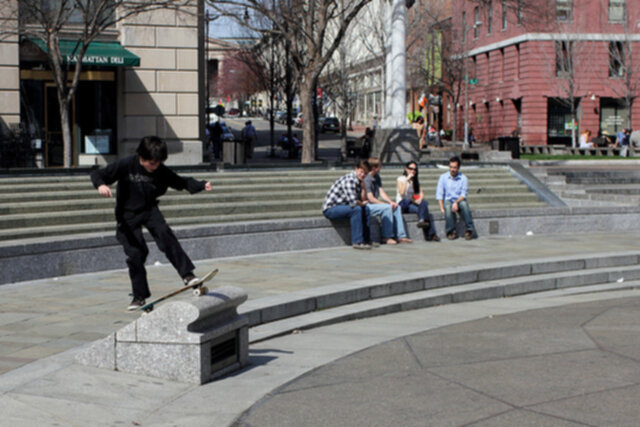}
 	
 	\noindent\textbf{Question:}is he wearing a hat ?

 	\noindent\textbf{MRR candidate set}
 	\begin{enumerate}
 		\item ﻿\textcolor{orange}{no                   }\item ﻿\textcolor{purple}{no he is not                }\item ﻿\textcolor{purple}{nope                   }\item ﻿\textcolor{purple}{i do n't see 1               }\end{enumerate}
 	\noindent\textbf{Top 10 from the remaining NDCG candidates}
 	\begin{enumerate}
 		\item i do n't think so               \item no , no hats                \item no baseball hat                 \item not that i can see               \item no ca n't see it               \item no he is not wearing a helmet             \item ca n't tell                 \item i ca n't tell                \item can not tell                 \item not really                  \end{enumerate}
 	﻿\includegraphics[width=1\linewidth]{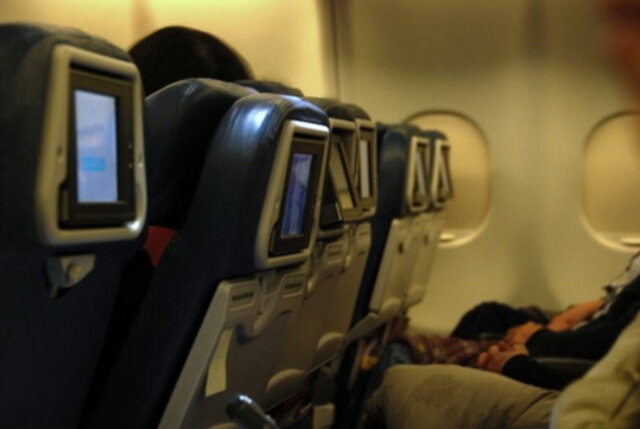}
 	
 	\noindent\textbf{Question:}are there any people ?

 	\noindent\textbf{MRR candidate set}
 	\begin{enumerate}
 		\item ﻿\textcolor{orange}{no                   }\item ﻿\textcolor{orange}{yes                   }\item ﻿\textcolor{purple}{not that i can see               }\item ﻿\textcolor{purple}{i do n't see any               }\item ﻿\textcolor{purple}{nope                   }\end{enumerate}
 	\noindent\textbf{Top 10 from the remaining NDCG candidates}
 	\begin{enumerate}
 		\item i ca n't see any , no             \item no people                  \item no people are visible                \item no , there are n't any people in the image          \item no there are n't                \item no people in the photo               \item 0 i can see                \item not that you can see               \item i ca n't see                \item on the screens there are               \end{enumerate}
 	﻿\includegraphics[width=1\linewidth]{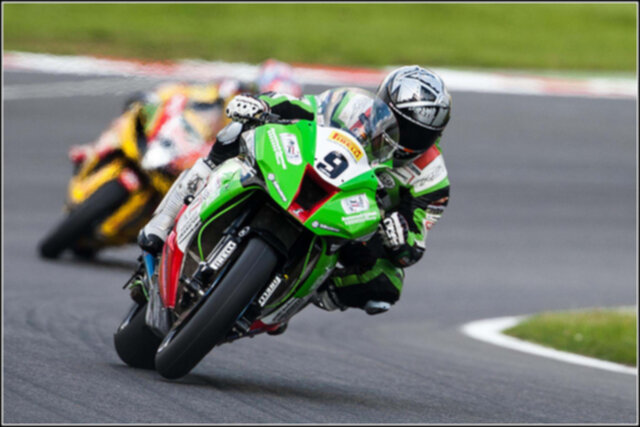}
 	
 	\noindent\textbf{Question:}do they have shoes on ?

 	\noindent\textbf{MRR candidate set}
 	\begin{enumerate}
 		\item ﻿\textcolor{orange}{yes                   }\item ﻿\textcolor{orange}{i can not tell                }\item ﻿\textcolor{red}{i see 1 shoe the other person is to blurry to tell        }\item ﻿\textcolor{purple}{it looks like it                }\item ﻿\textcolor{purple}{it 's hard to tell               }\item ﻿\textcolor{purple}{i ca n't tell                }\end{enumerate}
 	\noindent\textbf{Top 10 from the remaining NDCG candidates}
 	\begin{enumerate}
 		\item looks like it                 \item i guess so , it is hard to tell           \item yes i think so , you ca n't really see          \item ca n't tell                 \item i think so                 \item they are                  \item not sure                  \item no , they do not               \item ca n't see                 \item yes they are green                \end{enumerate}
 	﻿\includegraphics[width=1\linewidth]{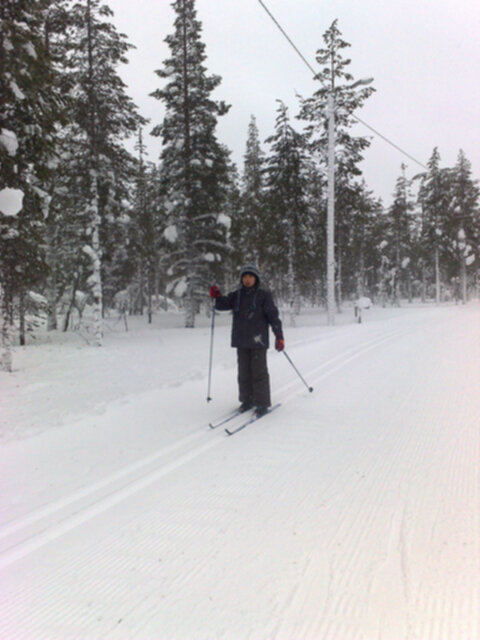}
 	
 	\noindent\textbf{Question:}does it look like a ski resort ?

 	\noindent\textbf{MRR candidate set}
 	\begin{enumerate}
 		\item ﻿\textcolor{orange}{no                   }\item ﻿\textcolor{red}{i can not tell                }\item ﻿\textcolor{purple}{i do n't think so               }\item ﻿\textcolor{purple}{no it does n't                }\item ﻿\textcolor{purple}{probably not                  }\item ﻿\textcolor{purple}{nope                   }\end{enumerate}
 	\noindent\textbf{Top 10 from the remaining NDCG candidates}
 	\begin{enumerate}
 		\item not really                  \item i ca n't tell                \item no i ca n't                \item i do n't think so but it 's hard to tell         \item not that i can see               \item not sure                  \item no , there is n't               \item maybe , i ca n't really tell             \item possibly                   \item not in the picture                \end{enumerate}
 	﻿\includegraphics[width=1\linewidth]{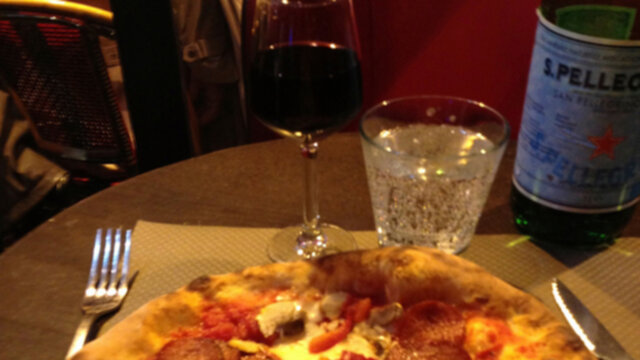}
 	
 	\noindent\textbf{Question:}is the picture in color ?

 	\noindent\textbf{MRR candidate set}
 	\begin{enumerate}
 		\item ﻿\textcolor{orange}{yes                   }\item ﻿\textcolor{orange}{yes it is                 }\item ﻿\textcolor{orange}{yes , it is                }\item ﻿\textcolor{purple}{yes it is in color               }\item ﻿\textcolor{purple}{yes it is a color image              }\end{enumerate}
 	\noindent\textbf{Top 10 from the remaining NDCG candidates}
 	\begin{enumerate}
 		\item yeah , this photo is in color             \item yes in color                 \item yes , it is fully colored              \item yyes                   \item full blown color                 \item i think so                 \item looks like it                 \item it is , but there is not much color in the photo due to the objects    \item somewhat , yes                 \item looks it                  \end{enumerate}
 	﻿\includegraphics[width=1\linewidth]{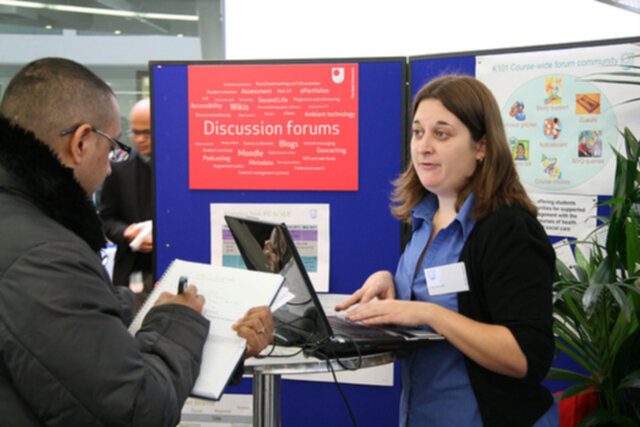}
 	
 	\noindent\textbf{Question:}is she wearing a blazer ?

 	\noindent\textbf{MRR candidate set}
 	\begin{enumerate}
 		\item ﻿\textcolor{orange}{no                   }\item ﻿\textcolor{orange}{no , sweater                 }\item ﻿\textcolor{purple}{nope                   }\item ﻿\textcolor{purple}{not that i can tell               }\item ﻿\textcolor{purple}{i do n't think so               }\end{enumerate}
 	\noindent\textbf{Top 10 from the remaining NDCG candidates}
 	\begin{enumerate}
 		\item no a sweater                 \item not that i can see               \item yes , a black 1               \item a long sleeve blouse                \item not sure                  \item yes                   \item can not see 1                \item do n't know                 \item 0 seen                  \item not really                  \end{enumerate}
 	﻿\includegraphics[width=1\linewidth]{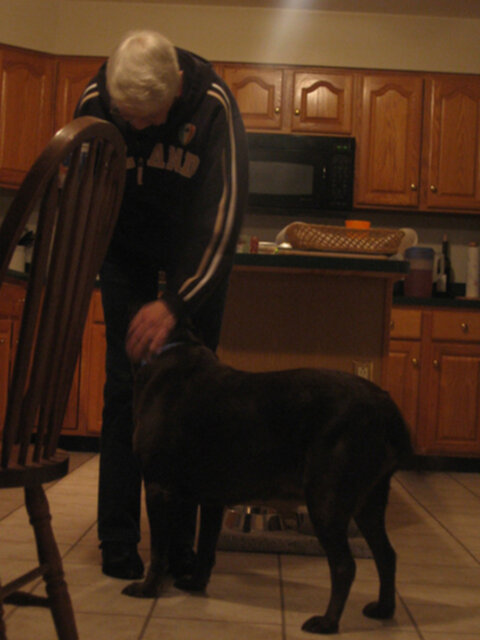}
 	
 	\noindent\textbf{Question:}does he look angry or otherwise UNK ?

 	\noindent\textbf{MRR candidate set}
 	\begin{enumerate}
 		\item ﻿\textcolor{orange}{no                   }\item ﻿\textcolor{orange}{ca n't tell                 }\item ﻿\textcolor{red}{you ca n't really see his face             }\item ﻿\textcolor{purple}{i ca n't tell                }\end{enumerate}
 	\noindent\textbf{Top 10 from the remaining NDCG candidates}
 	\begin{enumerate}
 		\item i ca n't see that               \item i can not tell                \item not real sure                 \item nope                   \item not really                  \item not sure                  \item i do n't think so               \item looks to be sleeping if anything              \item not that i can see               \item it looks determined , but not angry             \end{enumerate}
 	﻿\includegraphics[width=1\linewidth]{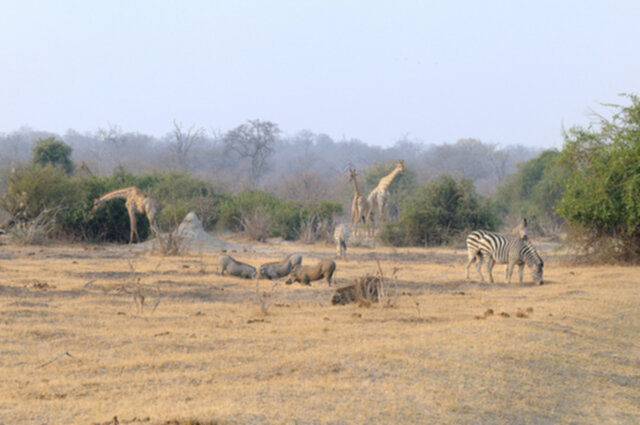}
 	
 	\noindent\textbf{Question:}are they UNK giraffes or UNK ?

 	\noindent\textbf{MRR candidate set}
 	\begin{enumerate}
 		\item ﻿\textcolor{red}{i can not tell                }\item ﻿\textcolor{red}{there are both                 }\item ﻿\textcolor{purple}{i ca n't tell                }\item ﻿\textcolor{purple}{not sure                  }\item ﻿\textcolor{purple}{ca n't tell                 }\end{enumerate}
 	\noindent\textbf{Top 10 from the remaining NDCG candidates}
 	\begin{enumerate}
 		\item some of each                 \item i do n't think so               \item hard to tell , but i do n't think so          \item it looks to be that way              \item no                   \item they are all various sizes               \item i think so                 \item both                   \item ca n't tell too close up              \item there are 2 UNK ones               \end{enumerate}
 	﻿\includegraphics[width=1\linewidth]{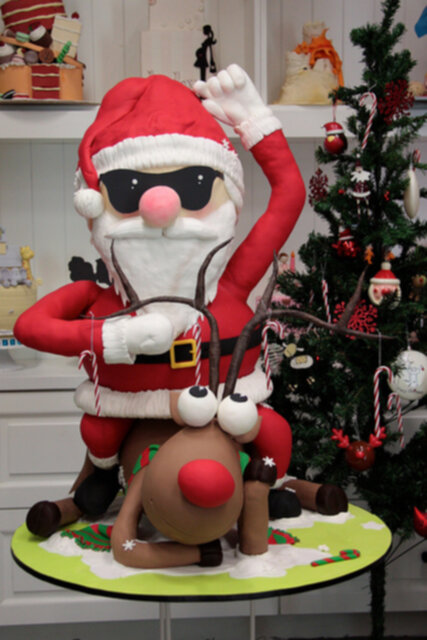}
 	
 	\noindent\textbf{Question:}what color is the reindeer ?

 	\noindent\textbf{MRR candidate set}
 	\begin{enumerate}
 		\item ﻿\textcolor{orange}{brown , with a red nose              }\item ﻿\textcolor{purple}{red                   }\item ﻿\textcolor{purple}{brown                   }\item ﻿\textcolor{purple}{brown and white                 }\item ﻿\textcolor{purple}{mostly white                  }\end{enumerate}
 	\noindent\textbf{Top 10 from the remaining NDCG candidates}
 	\begin{enumerate}
 		\item white                   \item golden                   \item white and tan                 \item brown , i think                \item white with a red stripe               \item brown , white and black               \item dark brown                  \item gold                   \item it is brownish gray                \item mostly white with some black               \end{enumerate}
 	﻿\includegraphics[width=1\linewidth]{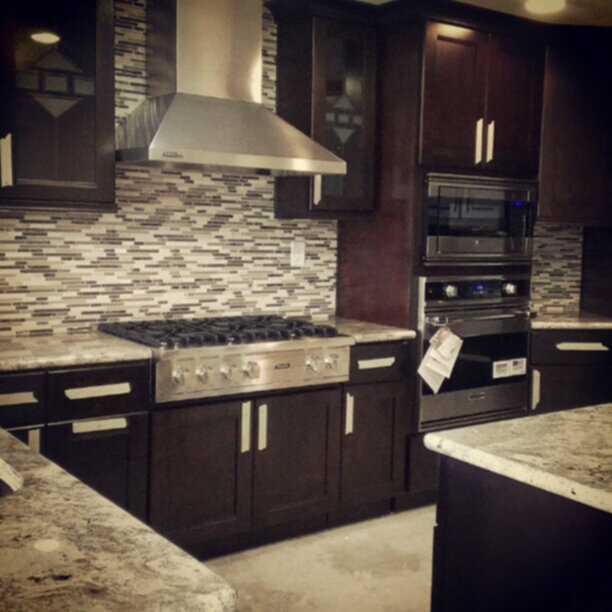}
 	
 	\noindent\textbf{Question:}can you tell what type they are , stone or laminate ?

 	\noindent\textbf{MRR candidate set}
 	\begin{enumerate}
 		\item ﻿\textcolor{orange}{they are stone                 }\item ﻿\textcolor{purple}{i can not                 }\item ﻿\textcolor{purple}{i can not tell                }\item ﻿\textcolor{purple}{i ca n't tell                }\item ﻿\textcolor{purple}{no i 'm not sure               }\item ﻿\textcolor{purple}{no , i can not               }\end{enumerate}
 	\noindent\textbf{Top 10 from the remaining NDCG candidates}
 	\begin{enumerate}
 		\item ca n't tell                 \item i can ' tell , it 's just white           \item nope                   \item no                   \item not sure                  \item not really                  \item no , i 'm bad at that             \item i ca n't te                \item no , i can only see the top            \item not that i can see               \end{enumerate}
 	﻿\includegraphics[width=1\linewidth]{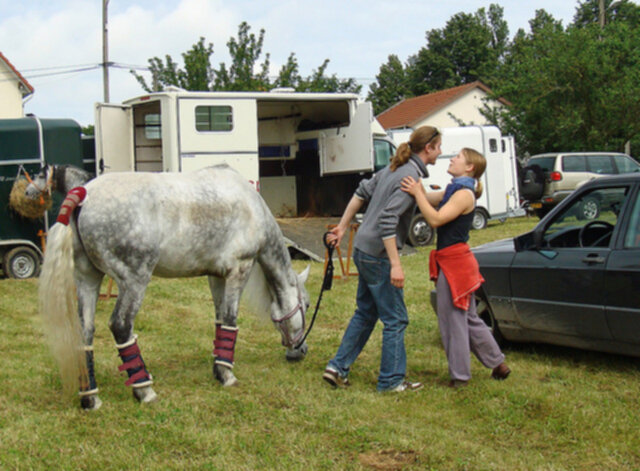}
 	
 	\noindent\textbf{Question:}any horses ?

 	\noindent\textbf{MRR candidate set}
 	\begin{enumerate}
 		\item ﻿\textcolor{orange}{no                   }\item ﻿\textcolor{red}{yes                   }\item ﻿\textcolor{purple}{yes , 1                 }\item ﻿\textcolor{purple}{1                   }\item ﻿\textcolor{purple}{nope                   }\end{enumerate}
 	\noindent\textbf{Top 10 from the remaining NDCG candidates}
 	\begin{enumerate}
 		\item 2                   \item i just see 2                \item yes , it looks like it              \item not that i can see               \item ye , 1 if not more              \item not that i see                \item i think so                 \item no there is n't                \item not visible                  \item barely part of 1                \end{enumerate}
 	﻿\includegraphics[width=1\linewidth]{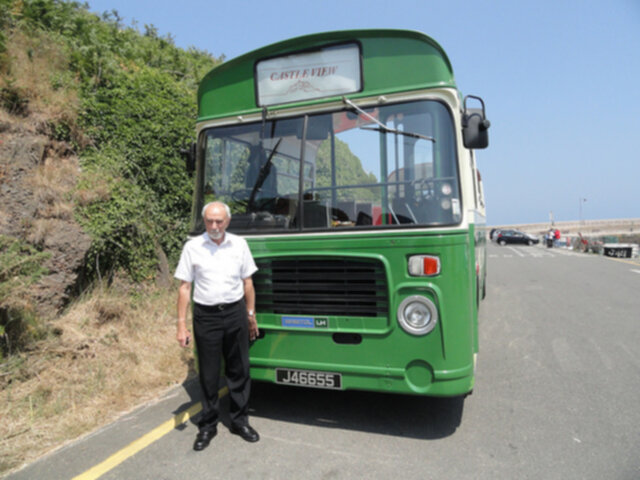}
 	
 	\noindent\textbf{Question:}are they male ?

 	\noindent\textbf{MRR candidate set}
 	\begin{enumerate}
 		\item ﻿\textcolor{orange}{yes                   }\item ﻿\textcolor{orange}{no                   }\item ﻿\textcolor{purple}{yes they are                 }\item ﻿\textcolor{purple}{i think so                 }\item ﻿\textcolor{purple}{i believe so                 }\item ﻿\textcolor{purple}{it seems like it                }\end{enumerate}
 	\noindent\textbf{Top 10 from the remaining NDCG candidates}
 	\begin{enumerate}
 		\item 1 is                  \item yes 1 is                 \item hard to tell                 \item yes !                  \item i ca n't tell                \item ca n't tell                 \item i can not tell                \item not sure                  \item no idea not visible                \item ca n't tell they are far away in background           \end{enumerate}
 	﻿\includegraphics[width=1\linewidth]{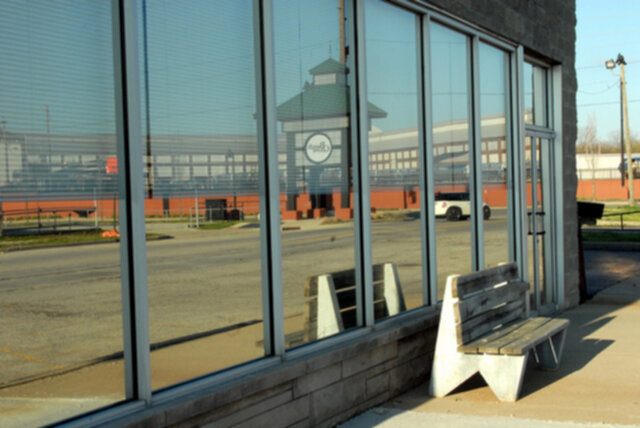}
 	
 	\noindent\textbf{Question:}is the bench in front of a building ?

 	\noindent\textbf{MRR candidate set}
 	\begin{enumerate}
 		\item ﻿\textcolor{red}{yes , it is                }\item ﻿\textcolor{orange}{yes                   }\item ﻿\textcolor{red}{yes in front of glass               }\item ﻿\textcolor{purple}{no it is n't                }\item ﻿\textcolor{purple}{nope                   }\end{enumerate}
 	\noindent\textbf{Top 10 from the remaining NDCG candidates}
 	\begin{enumerate}
 		\item no                   \item yes it is                 \item yes looks like it                \item not that i can see               \item looks like it                 \item appears so                  \item i think so                 \item i see a building on the left side , but only a bit of      \item i can not tell                \item i think so , it 's hard to tell           \end{enumerate}
 	﻿\includegraphics[width=1\linewidth]{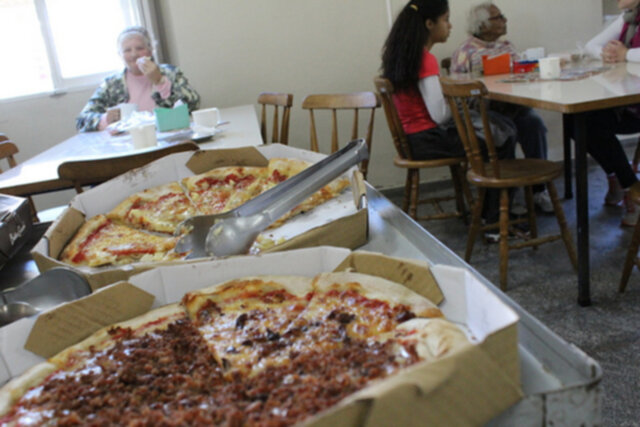}
 	
 	\noindent\textbf{Question:}how many pizzas ?

 	\noindent\textbf{MRR candidate set}
 	\begin{enumerate}
 		\item ﻿\textcolor{orange}{i can see 2 kinds of pizzas             }\item ﻿\textcolor{orange}{2                   }\item ﻿\textcolor{purple}{there are 3                 }\item ﻿\textcolor{purple}{there are 2                 }\item ﻿\textcolor{purple}{i see 2                 }\end{enumerate}
 	\noindent\textbf{Top 10 from the remaining NDCG candidates}
 	\begin{enumerate}
 		\item 2 in the scene                \item 3                   \item i think 4                 \item just 1                  \item i can not tell                \item 4                   \item it looks like a box of them and 2 sitting out         \item they are like 4                \item 1                   \item i ca n't tell                \end{enumerate}
 	﻿\includegraphics[width=1\linewidth]{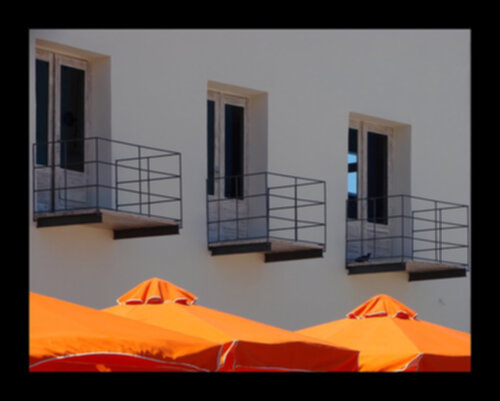}
 	
 	\noindent\textbf{Question:}what color is the building ?

 	\noindent\textbf{MRR candidate set}
 	\begin{enumerate}
 		\item ﻿\textcolor{orange}{white                   }\item ﻿\textcolor{purple}{grey                   }\item ﻿\textcolor{purple}{it is white                 }\item ﻿\textcolor{purple}{greyish                   }\item ﻿\textcolor{purple}{beige white                  }\end{enumerate}
 	\noindent\textbf{Top 10 from the remaining NDCG candidates}
 	\begin{enumerate}
 		\item white and black                 \item blue                   \item beige                   \item blue with a glass door               \item bright blue                  \item off white with some UNK of gray and red brown throughout         \item there is 1 building that is a cream color           \item it is glass                 \item crème                   \item it 's a brownish white color              \end{enumerate}
 	﻿\includegraphics[width=1\linewidth]{./VisualDialog_test2018/VisualDialog_test2018_000000113431.jpg}
 	
 	\noindent\textbf{Question:}is the photo in color ?

 	\noindent\textbf{MRR candidate set}
 	\begin{enumerate}
 		\item ﻿\textcolor{orange}{yes                   }\item ﻿\textcolor{orange}{yes it is                 }\item ﻿\textcolor{orange}{yes , it is                }\item ﻿\textcolor{purple}{yes , it is in color              }\item ﻿\textcolor{purple}{yes , the photo is in color             }\item ﻿\textcolor{purple}{yes it is in color               }\end{enumerate}
 	\noindent\textbf{Top 10 from the remaining NDCG candidates}
 	\begin{enumerate}
 		\item yes , the image is in color             \item yes in color                 \item yes , the picture is in color             \item the photo is in color               \item yes the picture is in color              \item it is                  \item sure is                  \item yes ,                  \item yup                   \item ye                   \end{enumerate}
 	﻿\includegraphics[width=1\linewidth]{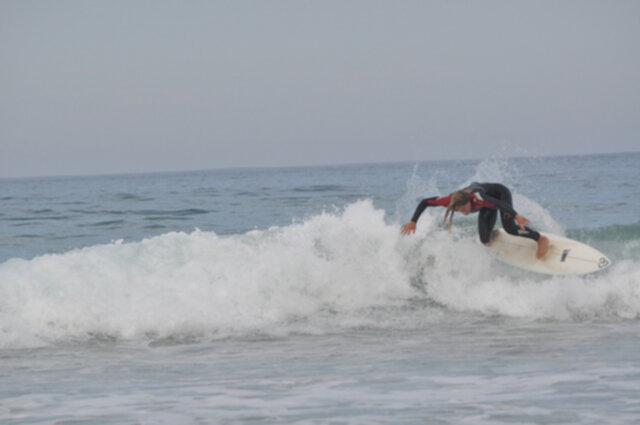}
 	
 	\noindent\textbf{Question:}is the day sunny ?

 	\noindent\textbf{MRR candidate set}
 	\begin{enumerate}
 		\item ﻿\textcolor{orange}{yes                   }\item ﻿\textcolor{orange}{no                   }\item ﻿\textcolor{purple}{not really                  }\item ﻿\textcolor{purple}{nope                   }\end{enumerate}
 	\noindent\textbf{Top 10 from the remaining NDCG candidates}
 	\begin{enumerate}
 		\item yes it is                 \item looks like it                 \item no , it is not               \item yes , it is                \item i think so                 \item UNK                   \item sort of                  \item i do n't think so               \item i suppose                  \item ca n't really see the sun think it 's pretty overcast         \end{enumerate}
 	﻿\includegraphics[width=1\linewidth]{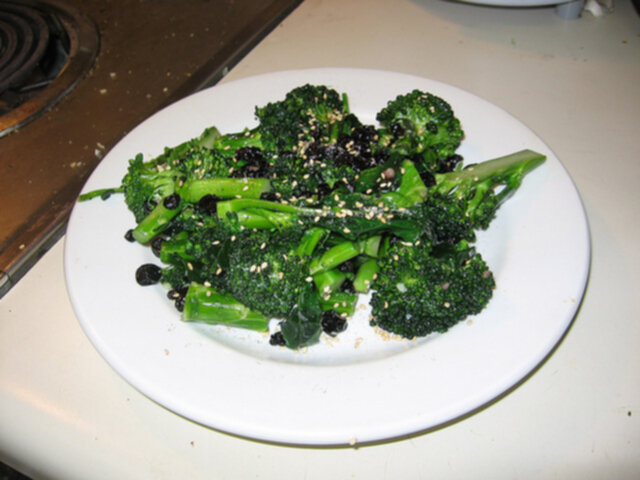}
 	
 	\noindent\textbf{Question:}are they on a white plate ?

 	\noindent\textbf{MRR candidate set}
 	\begin{enumerate}
 		\item ﻿\textcolor{orange}{yes                   }\item ﻿\textcolor{orange}{yes , it is                }\item ﻿\textcolor{purple}{yes they are                 }\item ﻿\textcolor{purple}{yes it is                 }\item ﻿\textcolor{purple}{i believe so                 }\end{enumerate}
 	\noindent\textbf{Top 10 from the remaining NDCG candidates}
 	\begin{enumerate}
 		\item it 's in color                \item yes , a brown 1               \item i think so                 \item UNK                   \item no it is in a paper boat             \item it 's okay                 \item white                   \item no , just on the table              \item yes , i can see some cars             \item it is in a box               \end{enumerate}
 	﻿\includegraphics[width=1\linewidth]{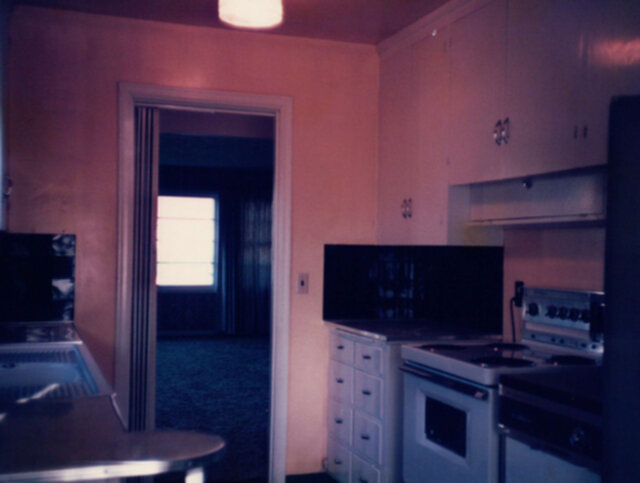}
 	
 	\noindent\textbf{Question:}is the sink made out of metal ?

 	\noindent\textbf{MRR candidate set}
 	\begin{enumerate}
 		\item ﻿\textcolor{orange}{yes                   }\item ﻿\textcolor{purple}{i can not tell                }\item ﻿\textcolor{purple}{i ca n't tell                }\item ﻿\textcolor{purple}{ca n't really tell                }\item ﻿\textcolor{purple}{ca n't tell                 }\end{enumerate}
 	\noindent\textbf{Top 10 from the remaining NDCG candidates}
 	\begin{enumerate}
 		\item not sure                  \item no                   \item yes it looks like it is              \item yes it is                 \item i think so                 \item looks like it                 \item yes , it is                \item it look like it                \item nope                   \item too far away to see that              \end{enumerate}
 	﻿\includegraphics[width=1\linewidth]{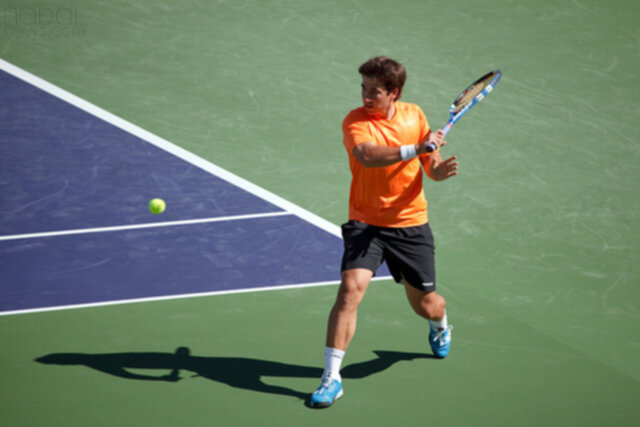}
 	
 	\noindent\textbf{Question:}is there people there apart from the man ?

 	\noindent\textbf{MRR candidate set}
 	\begin{enumerate}
 		\item ﻿\textcolor{orange}{no                   }\item ﻿\textcolor{purple}{no just 1 person                }\item ﻿\textcolor{purple}{not that i can see               }\item ﻿\textcolor{purple}{nope                   }\item ﻿\textcolor{purple}{i do n't see any               }\end{enumerate}
 	\noindent\textbf{Top 10 from the remaining NDCG candidates}
 	\begin{enumerate}
 		\item no other people                 \item i don ’ t see anyone at all            \item 0                   \item no people are in the picture              \item i do n't think so               \item no people                  \item no , there are n't any people visible            \item not really                  \item no , there are n't any people             \item no people in the room               \end{enumerate}
 	﻿\includegraphics[width=1\linewidth]{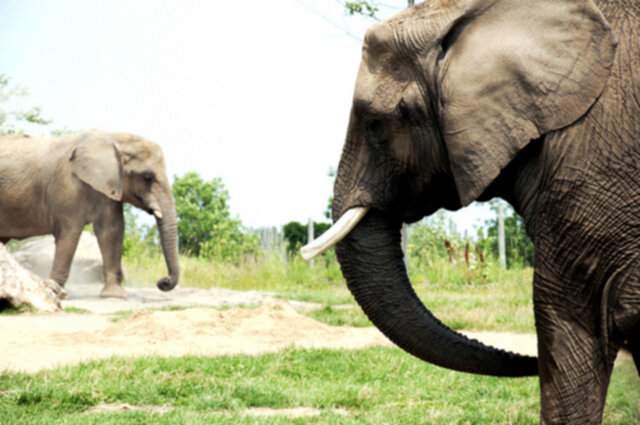}
 	
 	\noindent\textbf{Question:}are the elephants eating ?

 	\noindent\textbf{MRR candidate set}
 	\begin{enumerate}
 		\item ﻿\textcolor{orange}{no                   }\item ﻿\textcolor{orange}{yes                   }\item ﻿\textcolor{purple}{nope                   }\item ﻿\textcolor{purple}{no , do n't think so              }\end{enumerate}
 	\noindent\textbf{Top 10 from the remaining NDCG candidates}
 	\begin{enumerate}
 		\item no does n't look like it              \item i do n't think so               \item not that i can see               \item no standing                  \item it looks like it                \item only the 1 with its head down is            \item i think so                 \item yes they appear to be               \item yes , they appear to be              \item 1 of the 2 of them are             \end{enumerate}
 	﻿\includegraphics[width=1\linewidth]{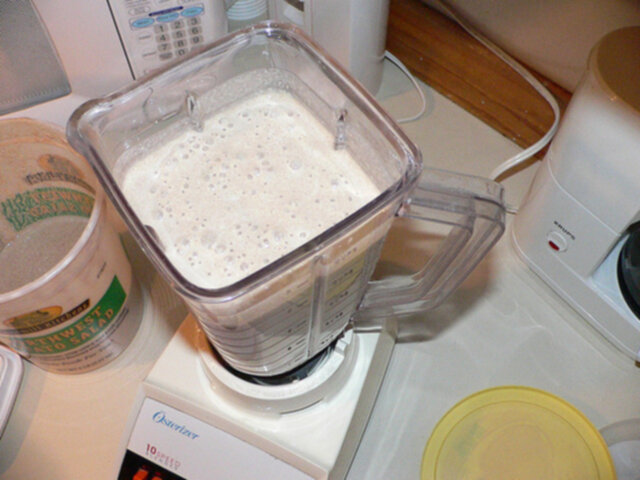}
 	
 	\noindent\textbf{Question:}is it a cold drink ?

 	\noindent\textbf{MRR candidate set}
 	\begin{enumerate}
 		\item ﻿\textcolor{orange}{no                   }\item ﻿\textcolor{orange}{yes                   }\item ﻿\textcolor{purple}{no it isn                 }\item ﻿\textcolor{purple}{nope                   }\end{enumerate}
 	\noindent\textbf{Top 10 from the remaining NDCG candidates}
 	\begin{enumerate}
 		\item looks that way                 \item i do n't think so               \item no not really                 \item i can not tell                \item looks like it                 \item looks like it is                \item no sure                  \item yes it is                 \item looks like                  \item it appears to be                \end{enumerate}
 	﻿\includegraphics[width=1\linewidth]{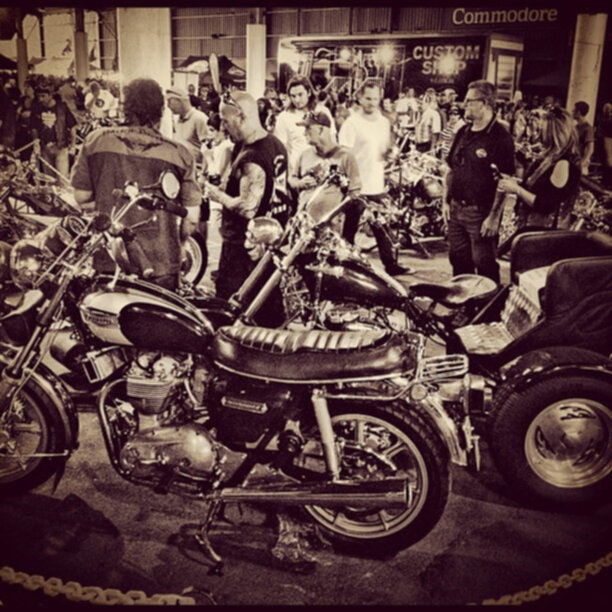}
 	
 	\noindent\textbf{Question:}do they look scary ?

 	\noindent\textbf{MRR candidate set}
 	\begin{enumerate}
 		\item ﻿\textcolor{orange}{no                   }\item ﻿\textcolor{purple}{not really                  }\item ﻿\textcolor{red}{yes kind of                 }\item ﻿\textcolor{orange}{yes                   }\item ﻿\textcolor{purple}{not particularly                  }\item ﻿\textcolor{purple}{nope                   }\end{enumerate}
 	\noindent\textbf{Top 10 from the remaining NDCG candidates}
 	\begin{enumerate}
 		\item not sure                  \item i do n't think so               \item not that i can see               \item i can not tell                \item hard to say                 \item maybe                   \item i ca n't tell                \item ca n't tell                 \item i ca n't see any               \item they look serious                 \end{enumerate}
 	﻿\includegraphics[width=1\linewidth]{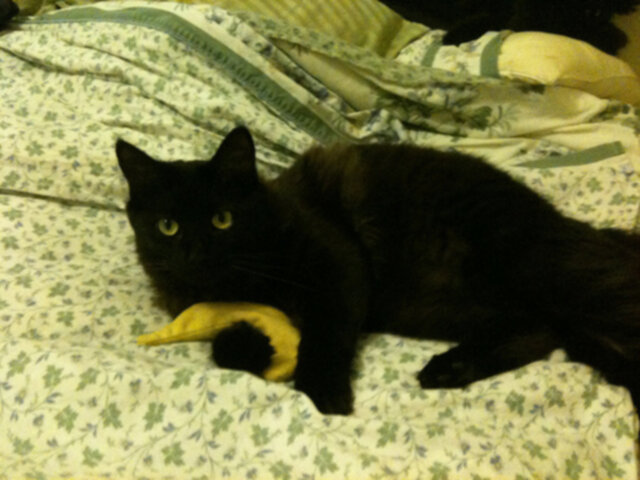}
 	
 	\noindent\textbf{Question:}are there any other animals in the picture ?

 	\noindent\textbf{MRR candidate set}
 	\begin{enumerate}
 		\item ﻿\textcolor{orange}{no                   }\item ﻿\textcolor{orange}{0                   }\item ﻿\textcolor{purple}{nope                   }\item ﻿\textcolor{purple}{0 at all                 }\item ﻿\textcolor{purple}{not that i can see               }\end{enumerate}
 	\noindent\textbf{Top 10 from the remaining NDCG candidates}
 	\begin{enumerate}
 		\item no , there are 0               \item i ca n't see any               \item no , just the 1               \item i do n't think so               \item nope , just 1                \item 0 sen                  \item no animals                  \item no animals either                 \item no other people                 \item no people                  \end{enumerate}
 	﻿\includegraphics[width=1\linewidth]{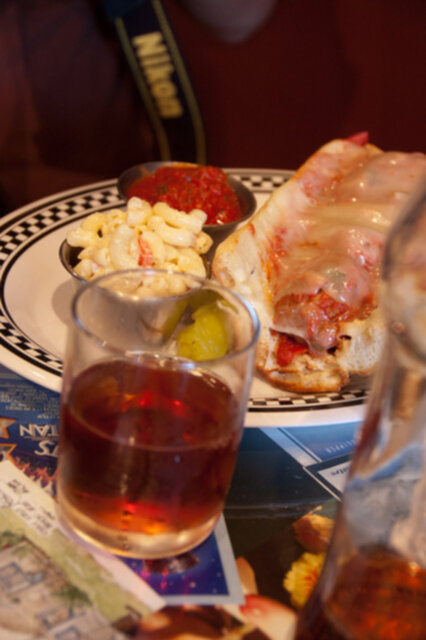}
 	
 	\noindent\textbf{Question:}does it look delicious ?

 	\noindent\textbf{MRR candidate set}
 	\begin{enumerate}
 		\item ﻿\textcolor{orange}{yes                   }\item ﻿\textcolor{purple}{yes it does                 }\item ﻿\textcolor{red}{yes , it is                }\item ﻿\textcolor{purple}{looks okay                  }\item ﻿\textcolor{purple}{it does yes                 }\item ﻿\textcolor{purple}{yes , making me hungry               }\end{enumerate}
 	\noindent\textbf{Top 10 from the remaining NDCG candidates}
 	\begin{enumerate}
 		\item it 's okay                 \item not really                  \item no not really                 \item UNK                   \item no it does not                \item no                   \item i think so                 \item no really                  \item nope                   \item no it 's not                \end{enumerate}
 	﻿\includegraphics[width=1\linewidth]{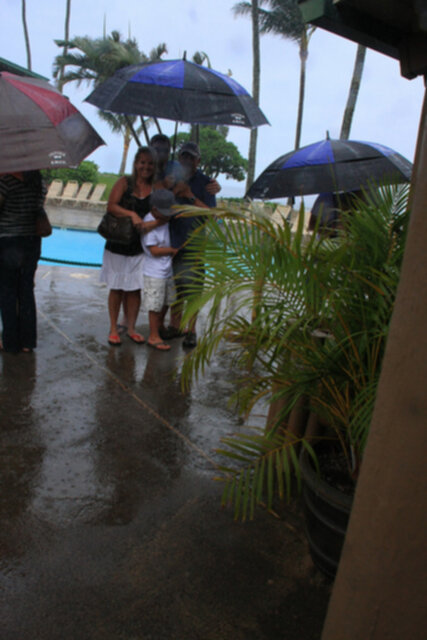}
 	
 	\noindent\textbf{Question:}are the kids playing with anything ?

 	\noindent\textbf{MRR candidate set}
 	\begin{enumerate}
 		\item ﻿\textcolor{orange}{holding umbrellas                  }\item ﻿\textcolor{orange}{no                   }\item ﻿\textcolor{purple}{yes                   }\item ﻿\textcolor{purple}{looks like a boy has something in his hand           }\item ﻿\textcolor{purple}{nope                   }\item ﻿\textcolor{purple}{not that i can see               }\end{enumerate}
 	\noindent\textbf{Top 10 from the remaining NDCG candidates}
 	\begin{enumerate}
 		\item no , all are kids               \item i do n't think so               \item i can not tell                \item not really                  \item yes i think so                \item nothing visible                  \item yes , they are                \item i ca n't tell that               \item yes , a bunch of umbrellas              \item yes they are                 \end{enumerate}
 	﻿\includegraphics[width=1\linewidth]{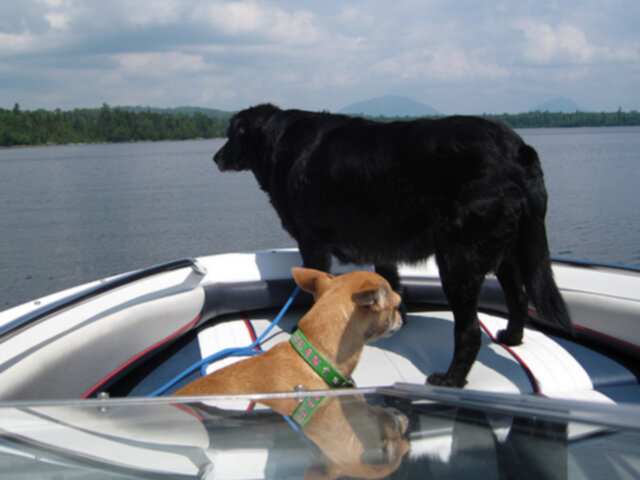}
 	
 	\noindent\textbf{Question:}are they older or younger ?

 	\noindent\textbf{MRR candidate set}
 	\begin{enumerate}
 		\item ﻿\textcolor{orange}{they look older                 }\item ﻿\textcolor{orange}{i think younger                 }\item ﻿\textcolor{red}{young                   }\item ﻿\textcolor{purple}{ca n't tell                 }\item ﻿\textcolor{purple}{do n't know                 }\item ﻿\textcolor{purple}{i am not sure                }\item ﻿\textcolor{purple}{i can not tell                }\item ﻿\textcolor{purple}{i ca n't tell                }\end{enumerate}
 	\noindent\textbf{Top 10 from the remaining NDCG candidates}
 	\begin{enumerate}
 		\item not sure                  \item looks like 1 of each               \item medium                   \item they look like adults mostly               \item ca n't see their faces               \item 1 looks old while the other looks mid 30s           \item very old                  \item they look fairly big                \item no                   \item i ca n't tell , but judging by the UNK , probably        \end{enumerate}
 	﻿\includegraphics[width=1\linewidth]{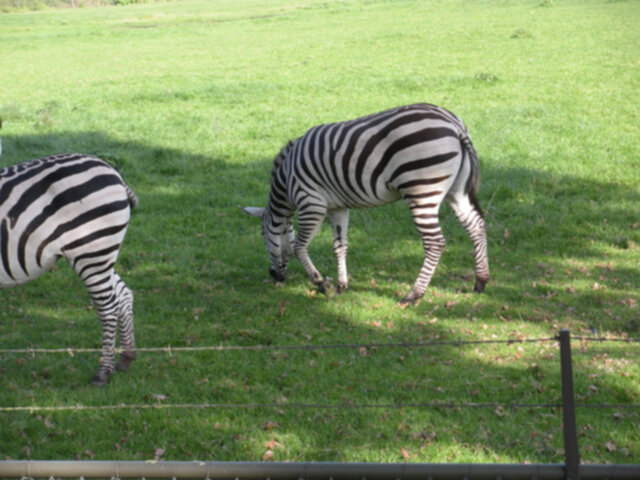}
 	
 	\noindent\textbf{Question:}is it a metal fence ?

 	\noindent\textbf{MRR candidate set}
 	\begin{enumerate}
 		\item ﻿\textcolor{orange}{yes                   }\item ﻿\textcolor{orange}{wire                   }\item ﻿\textcolor{purple}{yes it is                 }\item ﻿\textcolor{purple}{it appears to be                }\item ﻿\textcolor{purple}{yes , it is                }\end{enumerate}
 	\noindent\textbf{Top 10 from the remaining NDCG candidates}
 	\begin{enumerate}
 		\item it is                  \item yes , it looks like it is             \item it looks like it                \item looks like it is                \item i think so it                \item it has some lines , but yes             \item i think so                 \item yes they are                 \item yes , i can see a fence             \item yes i think so , hard to tell            \end{enumerate}
 	﻿\includegraphics[width=1\linewidth]{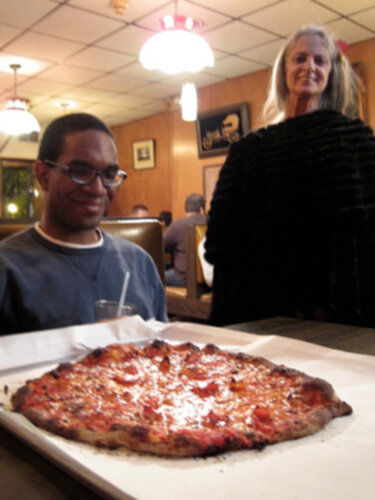}
 	
 	\noindent\textbf{Question:}are there pictures on the walls ?

 	\noindent\textbf{MRR candidate set}
 	\begin{enumerate}
 		\item ﻿\textcolor{orange}{yes                   }\item ﻿\textcolor{orange}{no                   }\item ﻿\textcolor{purple}{yes there are                 }\item ﻿\textcolor{purple}{yes for sure                 }\end{enumerate}
 	\noindent\textbf{Top 10 from the remaining NDCG candidates}
 	\begin{enumerate}
 		\item i think so                 \item yes , 1 that i can see             \item not that i can see               \item nope                   \item no pictures on the wall               \item yeah , 3                 \item 1                   \item might be but it is hard to see            \item no , i do n't see any             \item yes , but not too many              \end{enumerate}
 	﻿\includegraphics[width=1\linewidth]{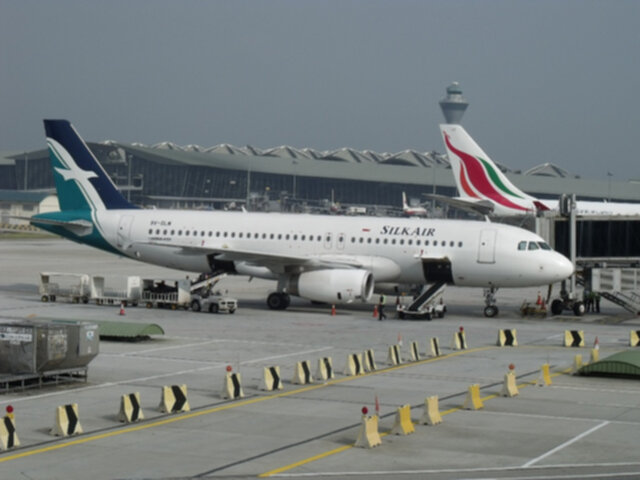}
 	
 	\noindent\textbf{Question:}can you see clouds ?

 	\noindent\textbf{MRR candidate set}
 	\begin{enumerate}
 		\item ﻿\textcolor{orange}{i see no clouds the sky is fairly grey though          }\item ﻿\textcolor{orange}{no clouds are visible                }\item ﻿\textcolor{orange}{yes i can see the sky cloudy             }\item ﻿\textcolor{purple}{no                   }\item ﻿\textcolor{purple}{no , it looks cloudy               }\end{enumerate}
 	\noindent\textbf{Top 10 from the remaining NDCG candidates}
 	\begin{enumerate}
 		\item no ,                  \item there seems to be a few clouds             \item nope                   \item yes they are clouds                \item very few                  \item not that i can see               \item not really                  \item i do n't think so               \item yes i can                 \item 0                   \end{enumerate}
 	﻿\includegraphics[width=1\linewidth]{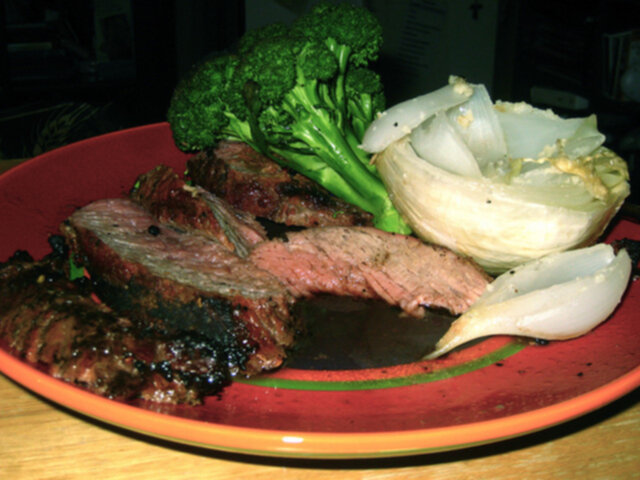}
 	
 	\noindent\textbf{Question:}where are the onions ?

 	\noindent\textbf{MRR candidate set}
 	\begin{enumerate}
 		\item ﻿\textcolor{orange}{on the right of the plate              }\item ﻿\textcolor{purple}{you can not really tell               }\item ﻿\textcolor{purple}{i can not tell                }\item ﻿\textcolor{purple}{not sure                  }\item ﻿\textcolor{purple}{i ca n't tell                }\end{enumerate}
 	\noindent\textbf{Top 10 from the remaining NDCG candidates}
 	\begin{enumerate}
 		\item yes                   \item ca n't tell                 \item UNK                   \item sitting at the table                \item it looks like a box of them and 2 sitting out         \item there is squash , shredded carrots , and shredded cabbage          \item not that i can see               \item the UNK of heaven                \item yes , it is                \item i think os                 \end{enumerate}
 	﻿\includegraphics[width=1\linewidth]{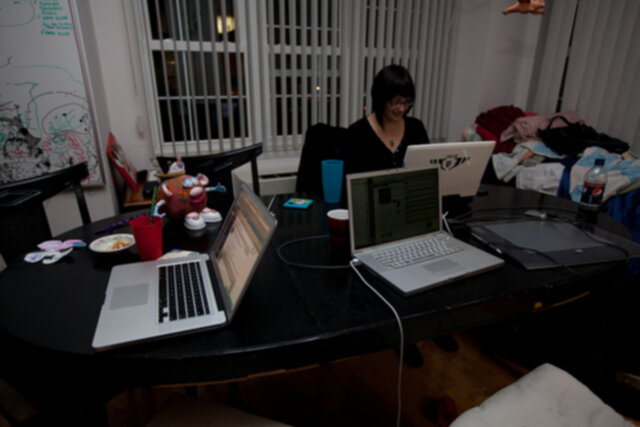}
 	
 	\noindent\textbf{Question:}does she have brown hair ?

 	\noindent\textbf{MRR candidate set}
 	\begin{enumerate}
 		\item ﻿\textcolor{orange}{yes                   }\item ﻿\textcolor{purple}{yes , she does                }\item ﻿\textcolor{red}{yes , it is                }\item ﻿\textcolor{purple}{i believe so                 }\item ﻿\textcolor{purple}{i think so                 }\item ﻿\textcolor{purple}{oh yeah                  }\end{enumerate}
 	\noindent\textbf{Top 10 from the remaining NDCG candidates}
 	\begin{enumerate}
 		\item black                   \item yes and black                 \item no                   \item no , her eyes are brown              \item nope                   \item it looks to be maybe of shoulder length            \item ca n't tell                 \item i ca n't tell                \item yes i do                 \item i can not tell                \end{enumerate}
 	﻿\includegraphics[width=1\linewidth]{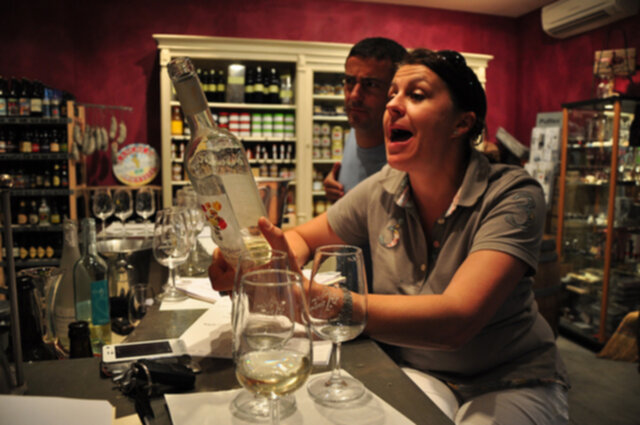}
 	
 	\noindent\textbf{Question:}are there other people ?

 	\noindent\textbf{MRR candidate set}
 	\begin{enumerate}
 		\item ﻿\textcolor{orange}{there is a gentleman beside her              }\item ﻿\textcolor{orange}{no                   }\item ﻿\textcolor{orange}{yes , there are 2 other people in the background          }\item ﻿\textcolor{purple}{no , there is not               }\item ﻿\textcolor{purple}{not that i can see               }\item ﻿\textcolor{purple}{no there are not                }\item ﻿\textcolor{purple}{nope                   }\end{enumerate}
 	\noindent\textbf{Top 10 from the remaining NDCG candidates}
 	\begin{enumerate}
 		\item no other people in the photo              \item 0 that i can see in the picture            \item no , there are no other people com            \item 0 i can see                \item no there 's not                \item not from what i can see              \item no , i ca n't see anyone             \item not within view                 \item yes 2 people can be seen in the background           \item i do n't think so               \end{enumerate}
 	﻿\includegraphics[width=1\linewidth]{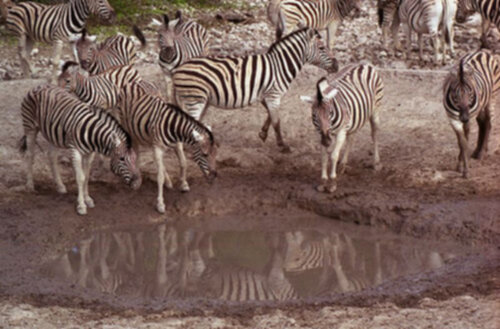}
 	
 	\noindent\textbf{Question:}how many zebras are there ?

 	\noindent\textbf{MRR candidate set}
 	\begin{enumerate}
 		\item ﻿\textcolor{orange}{9                   }\item ﻿\textcolor{orange}{i count twelve                 }\item ﻿\textcolor{purple}{10                   }\item ﻿\textcolor{purple}{7                   }\item ﻿\textcolor{purple}{8 or so                 }\item ﻿\textcolor{purple}{too many to count                }\end{enumerate}
 	\noindent\textbf{Top 10 from the remaining NDCG candidates}
 	\begin{enumerate}
 		\item about ten , of different species              \item eleven                   \item at least 20                 \item 11 or 12                 \item looks like around a dozen               \item a lot                  \item over 20                  \item about 12-14                  \item about 25                  \item i see 6 i think               \end{enumerate}
 	﻿\includegraphics[width=1\linewidth]{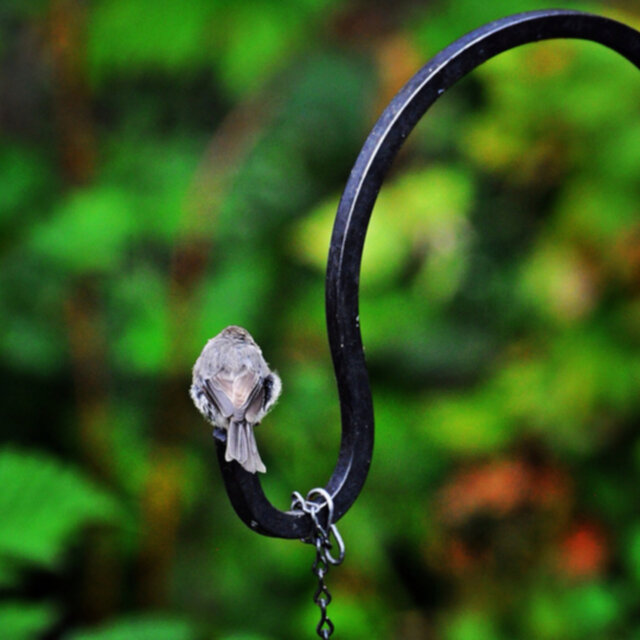}
 	
 	\noindent\textbf{Question:}is the yard large ?

 	\noindent\textbf{MRR candidate set}
 	\begin{enumerate}
 		\item ﻿\textcolor{orange}{i can not tell                }\item ﻿\textcolor{orange}{yes                   }\item ﻿\textcolor{purple}{i ca n't tell                }\item ﻿\textcolor{purple}{ca n't tell                 }\item ﻿\textcolor{purple}{can not tell too close up              }\item ﻿\textcolor{purple}{not sure                  }\end{enumerate}
 	\noindent\textbf{Top 10 from the remaining NDCG candidates}
 	\begin{enumerate}
 		\item from what i can see               \item seems like it very close up photo             \item i think so                 \item maybe                   \item no                   \item no , but hard to tell from the close up          \item yes it is                 \item yes , it is                \item not really                  \item yes it 's large                \end{enumerate}
 	﻿\includegraphics[width=1\linewidth]{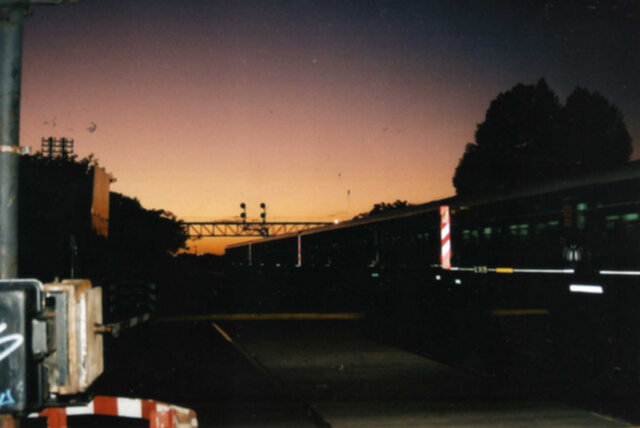}
 	
 	\noindent\textbf{Question:}does the picture match the caption ?

 	\noindent\textbf{MRR candidate set}
 	\begin{enumerate}
 		\item ﻿\textcolor{orange}{yes                   }\item ﻿\textcolor{purple}{no                   }\item ﻿\textcolor{red}{yes , it does                }\item ﻿\textcolor{red}{yes it does match the caption              }\item ﻿\textcolor{purple}{yes it does                 }\item ﻿\textcolor{purple}{no it is not                }\end{enumerate}
 	\noindent\textbf{Top 10 from the remaining NDCG candidates}
 	\begin{enumerate}
 		\item nope                   \item no ,                  \item not really                  \item it does                  \item yes , the picture is in color             \item yes , it 's in color              \item it is                  \item yes , it is                \item yes it is                 \item i think so                 \end{enumerate}
 	﻿\includegraphics[width=1\linewidth]{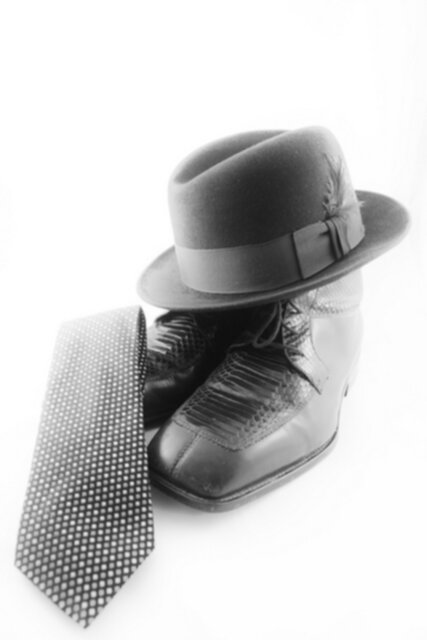}
 	
 	\noindent\textbf{Question:}are there any lights ?

 	\noindent\textbf{MRR candidate set}
 	\begin{enumerate}
 		\item ﻿\textcolor{orange}{no                   }\item ﻿\textcolor{orange}{i think it 's light coming from a camera he is holding        }\item ﻿\textcolor{orange}{yes                   }\item ﻿\textcolor{purple}{not that i can see               }\item ﻿\textcolor{purple}{0 that i can see               }\item ﻿\textcolor{purple}{nope                   }\end{enumerate}
 	\noindent\textbf{Top 10 from the remaining NDCG candidates}
 	\begin{enumerate}
 		\item not seen                  \item i can not see them               \item i do n't think so               \item yes , well lit                \item i ca n't tell                \item 0 at all                 \item i think so                 \item i can not tell                \item yes there are                 \item 0                   \end{enumerate}
 	﻿\includegraphics[width=1\linewidth]{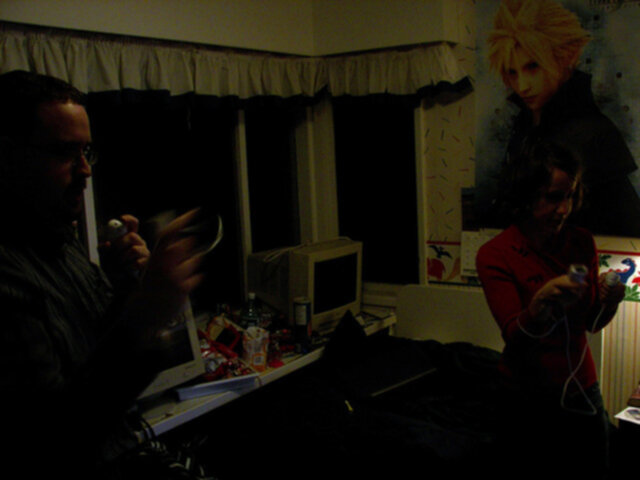}
 	
 	\noindent\textbf{Question:}how many kids are there ?

 	\noindent\textbf{MRR candidate set}
 	\begin{enumerate}
 		\item ﻿\textcolor{orange}{2                   }\item ﻿\textcolor{red}{there are 2 people but they are n't kids           }\item ﻿\textcolor{purple}{i think 2                 }\item ﻿\textcolor{purple}{only 2                  }\item ﻿\textcolor{purple}{there is only 1                }\end{enumerate}
 	\noindent\textbf{Top 10 from the remaining NDCG candidates}
 	\begin{enumerate}
 		\item 2 , possibly 3                \item 1                   \item 3                   \item looks like 3                 \item i can see 2 clearly and there may be part of a third but i 'm not sure  \item there is 1                 \item only see 1 , too close up to see much          \item 3 total 1 looks like the guide             \item there are tow                 \item i see 1 arm besides them              \end{enumerate}
 	﻿\includegraphics[width=1\linewidth]{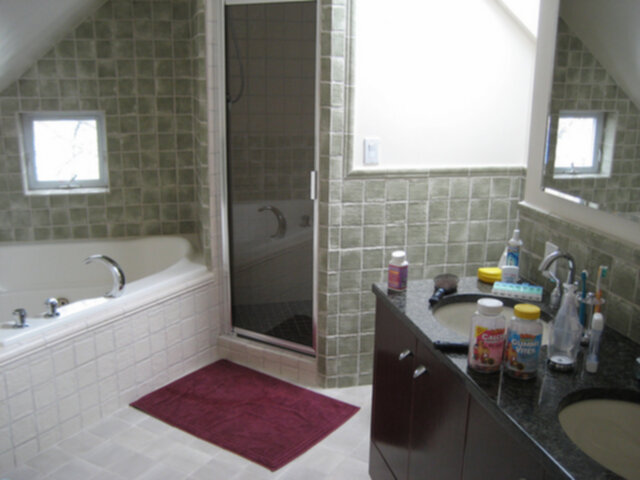}
 	
 	\noindent\textbf{Question:}are there any bath mats ?

 	\noindent\textbf{MRR candidate set}
 	\begin{enumerate}
 		\item ﻿\textcolor{orange}{no                   }\item ﻿\textcolor{red}{yes there is a bath mat              }\item ﻿\textcolor{red}{i can not tell                }\item ﻿\textcolor{purple}{nope                   }\item ﻿\textcolor{purple}{no , it does n't appear so             }\end{enumerate}
 	\noindent\textbf{Top 10 from the remaining NDCG candidates}
 	\begin{enumerate}
 		\item yes                   \item yes there are                 \item no ,                  \item yes there is 1                \item there are                  \item not that i can see               \item not that you can see               \item just 1                  \item i do n't think so               \item 0 that i can see               \end{enumerate}
 	﻿\includegraphics[width=1\linewidth]{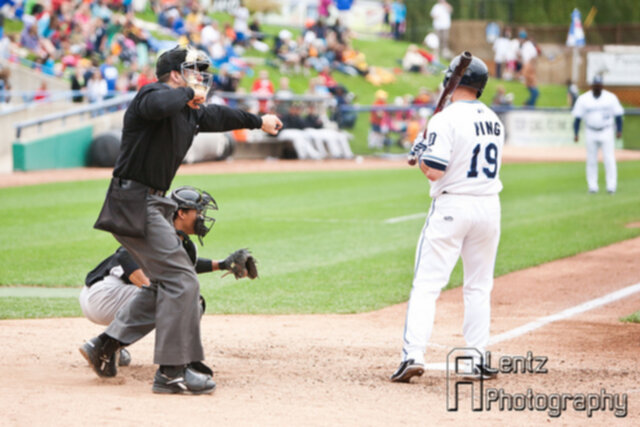}
 	
 	\noindent\textbf{Question:}can you see the sky ?

 	\noindent\textbf{MRR candidate set}
 	\begin{enumerate}
 		\item ﻿\textcolor{orange}{no                   }\item ﻿\textcolor{purple}{no i can not                }\item ﻿\textcolor{purple}{no i ca n't                }\item ﻿\textcolor{purple}{no , i can not               }\item ﻿\textcolor{purple}{nope                   }\end{enumerate}
 	\noindent\textbf{Top 10 from the remaining NDCG candidates}
 	\begin{enumerate}
 		\item no you ca n't                \item no i ca n't see the sky             \item no i do not                \item no , the sky is not visible             \item not really                  \item no i can UNK                \item just a little bit                \item n                   \item i do n't think so               \item a little bit                 \end{enumerate}
 	﻿\includegraphics[width=1\linewidth]{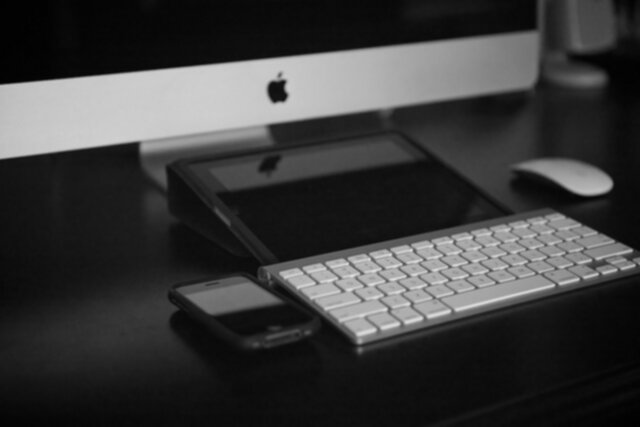}
 	
 	\noindent\textbf{Question:}is there a chair ?

 	\noindent\textbf{MRR candidate set}
 	\begin{enumerate}
 		\item ﻿\textcolor{orange}{no , i can only see the close up of what ’ s on the desk    }\item ﻿\textcolor{orange}{no                   }\item ﻿\textcolor{purple}{not that i can see               }\item ﻿\textcolor{purple}{i do n't see 1               }\item ﻿\textcolor{purple}{no there is n't                }\end{enumerate}
 	\noindent\textbf{Top 10 from the remaining NDCG candidates}
 	\begin{enumerate}
 		\item not that can be seen               \item not visible                  \item nope                   \item do not see 1                \item do n't see any                \item 0 that i can see               \item ca n't tell                 \item i do n't think so               \item i can not tell                \item not sure                  \end{enumerate}
 	﻿\includegraphics[width=1\linewidth]{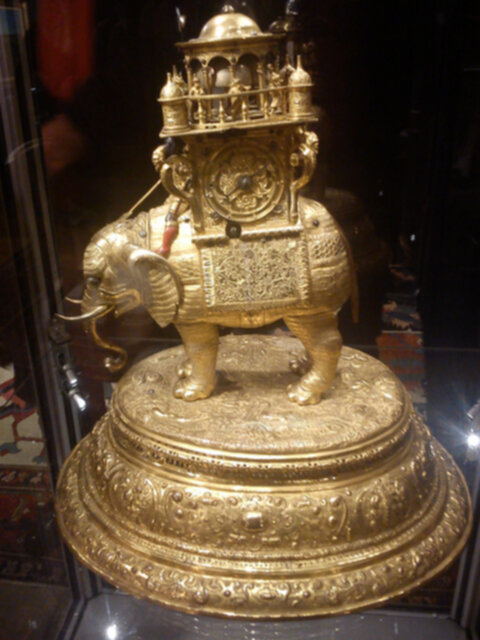}
 	
 	\noindent\textbf{Question:}are there any people in the image ?

 	\noindent\textbf{MRR candidate set}
 	\begin{enumerate}
 		\item ﻿\textcolor{orange}{no                   }\item ﻿\textcolor{purple}{no people                  }\item ﻿\textcolor{purple}{nope                   }\item ﻿\textcolor{purple}{no i do not see people              }\item ﻿\textcolor{purple}{not that i can see               }\end{enumerate}
 	\noindent\textbf{Top 10 from the remaining NDCG candidates}
 	\begin{enumerate}
 		\item noo                   \item 0 are visible                 \item 0 i can see                \item 0                   \item nobody around                  \item 0 sen                  \item do n't really see any               \item mp                   \item i do n't think so               \item not really                  \end{enumerate}
 	﻿\includegraphics[width=1\linewidth]{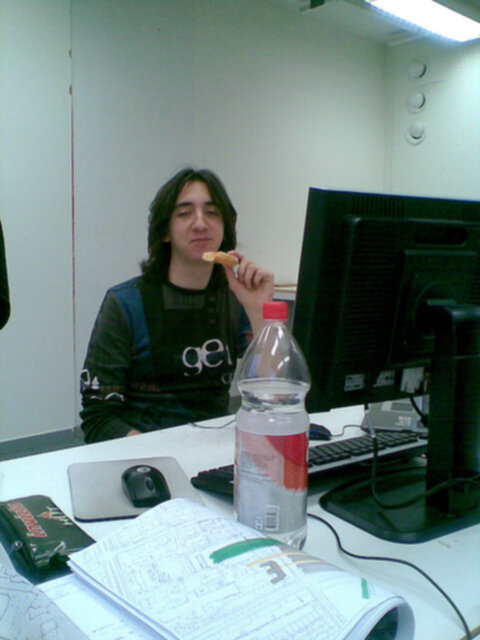}
 	
 	\noindent\textbf{Question:}is this a living room ?

 	\noindent\textbf{MRR candidate set}
 	\begin{enumerate}
 		\item ﻿\textcolor{orange}{no                   }\item ﻿\textcolor{red}{yes                   }\item ﻿\textcolor{purple}{ca n't tell                 }\item ﻿\textcolor{purple}{i do n't think so               }\item ﻿\textcolor{purple}{i ca n't tell                }\item ﻿\textcolor{purple}{nope                   }\end{enumerate}
 	\noindent\textbf{Top 10 from the remaining NDCG candidates}
 	\begin{enumerate}
 		\item i do not think so               \item ca n't really tell                \item can not tell                 \item i can not tell                \item i do n't know                \item i ca n't tell for sure              \item not sure                  \item i don ’ t know               \item hard to tell                 \item not that i can see               \end{enumerate}
 	﻿\includegraphics[width=1\linewidth]{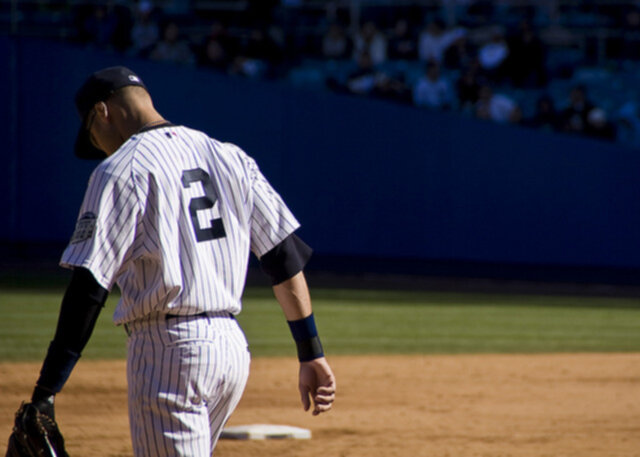}
 	
 	\noindent\textbf{Question:}what color is the jersey ?

 	\noindent\textbf{MRR candidate set}
 	\begin{enumerate}
 		\item ﻿\textcolor{orange}{pinstripe white                  }\item ﻿\textcolor{orange}{white                   }\item ﻿\textcolor{purple}{it is white                 }\item ﻿\textcolor{purple}{blue and white                 }\item ﻿\textcolor{purple}{it 's white                 }\end{enumerate}
 	\noindent\textbf{Top 10 from the remaining NDCG candidates}
 	\begin{enumerate}
 		\item grey                   \item photo is in black and white              \item the picture is in black and white             \item the pants are gray and his jersey is navy blue          \item its a black and white picture so i 'm not sure         \item we can say that                \item it 's blue                 \item grey and UNK cloudy                \item beige and blue                 \item the die is a grey and white i believe           \end{enumerate}
 	﻿\includegraphics[width=1\linewidth]{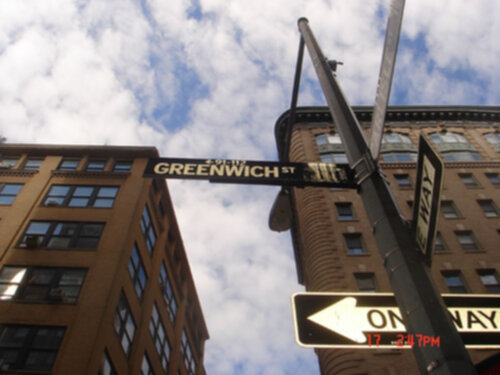}
 	
 	\noindent\textbf{Question:}are they tall ?

 	\noindent\textbf{MRR candidate set}
 	\begin{enumerate}
 		\item ﻿\textcolor{orange}{yes , they are tall               }\item ﻿\textcolor{orange}{yes                   }\item ﻿\textcolor{purple}{yes they are                 }\item ﻿\textcolor{purple}{yes they definitely are                }\end{enumerate}
 	\noindent\textbf{Top 10 from the remaining NDCG candidates}
 	\begin{enumerate}
 		\item yes a few stories at least              \item looks like                  \item yes , some are                \item i think so                 \item not really                  \item yes , it is                \item no                   \item yes it is                 \item nope                   \item possibly                   \end{enumerate}
 	﻿\includegraphics[width=1\linewidth]{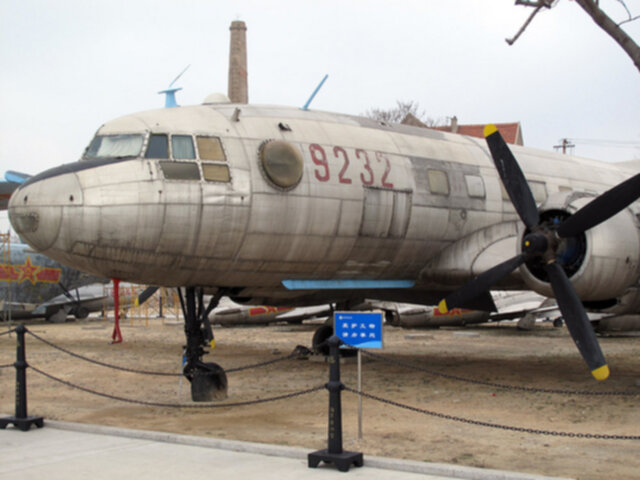}
 	
 	\noindent\textbf{Question:}does it look rusty ?

 	\noindent\textbf{MRR candidate set}
 	\begin{enumerate}
 		\item ﻿\textcolor{orange}{no                   }\item ﻿\textcolor{orange}{yes                   }\item ﻿\textcolor{purple}{not at all                 }\item ﻿\textcolor{purple}{nope                   }\end{enumerate}
 	\noindent\textbf{Top 10 from the remaining NDCG candidates}
 	\begin{enumerate}
 		\item not really                  \item no it is not                \item kind of                  \item no but looks like it did previously             \item yes , it looks UNK               \item not that i can see               \item yes , it does                \item i do n't think so               \item yeah                   \item sort of                  \end{enumerate}
 	﻿\includegraphics[width=1\linewidth]{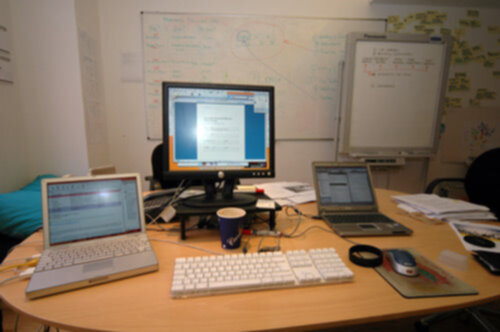}
 	
 	\noindent\textbf{Question:}is this inside ?

 	\noindent\textbf{MRR candidate set}
 	\begin{enumerate}
 		\item ﻿\textcolor{orange}{yes                   }\item ﻿\textcolor{orange}{yes it is                 }\item ﻿\textcolor{orange}{yes , it is                }\item ﻿\textcolor{purple}{yes it appears to be               }\item ﻿\textcolor{purple}{i think so                 }\end{enumerate}
 	\noindent\textbf{Top 10 from the remaining NDCG candidates}
 	\begin{enumerate}
 		\item yes it 's indoors                \item i believe so                 \item yes* and yes                 \item appear to be                 \item most likely indoors                 \item it might be                 \item i believe so but it is hard to tell for sure         \item most                   \item yes it 's clean                \item maybe                   \end{enumerate}
 	﻿\includegraphics[width=1\linewidth]{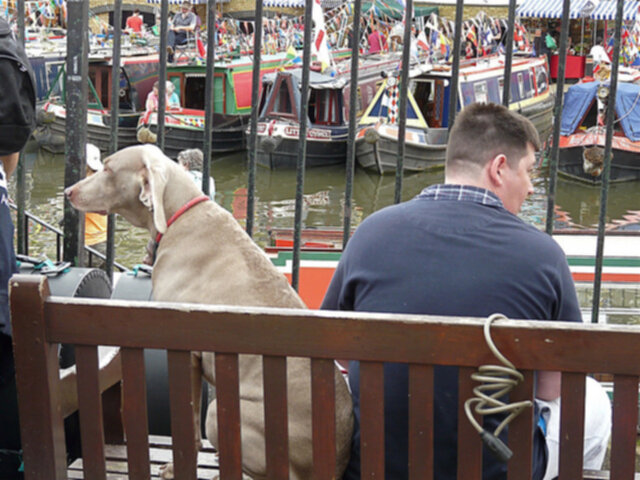}
 	
 	\noindent\textbf{Question:}how many other people are there ?

 	\noindent\textbf{MRR candidate set}
 	\begin{enumerate}
 		\item ﻿\textcolor{red}{2 people                  }\item ﻿\textcolor{red}{a crowd                  }\item ﻿\textcolor{purple}{lots and lots                 }\item ﻿\textcolor{purple}{very many                  }\item ﻿\textcolor{purple}{approximately 20                  }\end{enumerate}
 	\noindent\textbf{Top 10 from the remaining NDCG candidates}
 	\begin{enumerate}
 		\item it looks like over 20               \item 15 maybe                  \item 10                   \item about a 100                 \item about 10                  \item ca n't tell                 \item 8 or so                 \item 12 or so                 \item around 10                  \item i can not tell                \end{enumerate}
 	﻿\includegraphics[width=1\linewidth]{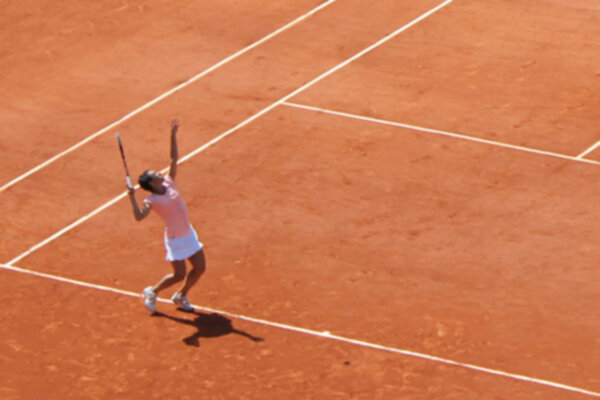}
 	
 	\noindent\textbf{Question:}is she wearing shorts ?

 	\noindent\textbf{MRR candidate set}
 	\begin{enumerate}
 		\item ﻿\textcolor{orange}{no a white skirt                }\item ﻿\textcolor{orange}{yes                   }\item ﻿\textcolor{orange}{no , she is wearing a skirt             }\item ﻿\textcolor{purple}{no                   }\item ﻿\textcolor{purple}{yes she is                 }\item ﻿\textcolor{purple}{i think so                 }\end{enumerate}
 	\noindent\textbf{Top 10 from the remaining NDCG candidates}
 	\begin{enumerate}
 		\item white                   \item hard to tell , but i think so            \item nope                   \item yes , it is                \item can                   \item yes , they are                \item yes he is                 \item yes it is                 \item he                   \item maybe                   \end{enumerate}
 	﻿\includegraphics[width=1\linewidth]{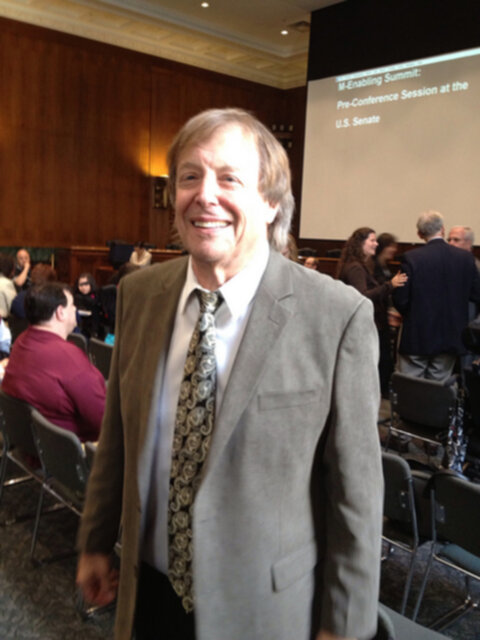}
 	
 	\noindent\textbf{Question:}what type of flooring ?

 	\noindent\textbf{MRR candidate set}
 	\begin{enumerate}
 		\item ﻿\textcolor{orange}{carpet                   }\item ﻿\textcolor{purple}{tile                   }\item ﻿\textcolor{purple}{i can not tell                }\item ﻿\textcolor{purple}{i ca n't tell                }\item ﻿\textcolor{purple}{ca n't tell                 }\end{enumerate}
 	\noindent\textbf{Top 10 from the remaining NDCG candidates}
 	\begin{enumerate}
 		\item not sure                  \item i don ’ t know for sure             \item maybe tan                  \item ca n't tell from the picture              \item no idea                  \item brown                   \item grey                   \item seem grayish from this angle               \item green                   \item color of what                 \end{enumerate}
 	﻿\includegraphics[width=1\linewidth]{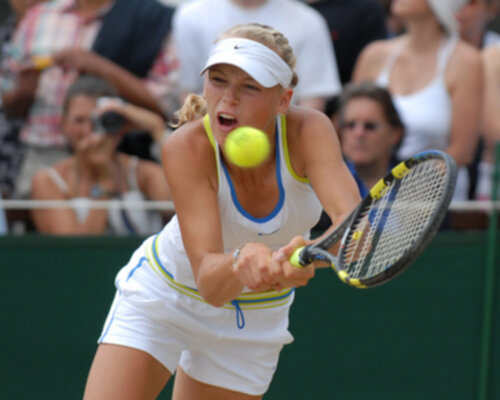}
 	
 	\noindent\textbf{Question:}cab you tell the women 's race ?

 	\noindent\textbf{MRR candidate set}
 	\begin{enumerate}
 		\item ﻿\textcolor{orange}{caucasian                   }\item ﻿\textcolor{orange}{white                   }\item ﻿\textcolor{purple}{no i can not                }\item ﻿\textcolor{purple}{no i ca n't                }\end{enumerate}
 	\noindent\textbf{Top 10 from the remaining NDCG candidates}
 	\begin{enumerate}
 		\item no                   \item no i do not                \item nope                   \item yes                   \item i can not tell                \item not sure                  \item not really                  \item yes , i can                \item yes i can                 \item i ca n't tell                \end{enumerate}
 	﻿\includegraphics[width=1\linewidth]{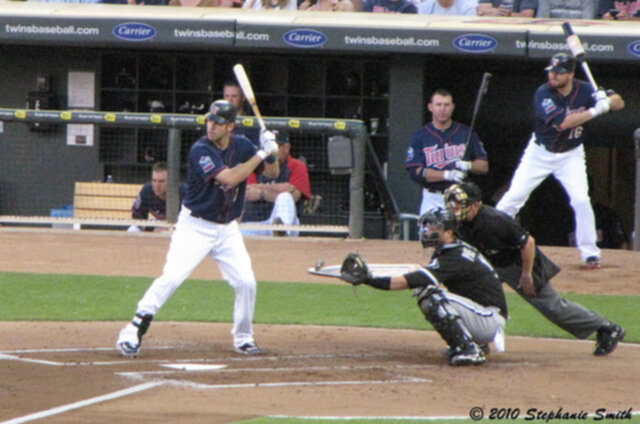}
 	
 	\noindent\textbf{Question:}can you see the scoreboard ?

 	\noindent\textbf{MRR candidate set}
 	\begin{enumerate}
 		\item ﻿\textcolor{orange}{no                   }\item ﻿\textcolor{orange}{yes                   }\item ﻿\textcolor{purple}{i can not                 }\item ﻿\textcolor{purple}{nope                   }\item ﻿\textcolor{purple}{no i can not                }\end{enumerate}
 	\noindent\textbf{Top 10 from the remaining NDCG candidates}
 	\begin{enumerate}
 		\item no it is cut off               \item not really                  \item n                   \item i do n't think so               \item not that i can see               \item i can not tell                \item not sure                  \item barely                   \item i ca n't tell                \item ca n't tell                 \end{enumerate}
 	﻿\includegraphics[width=1\linewidth]{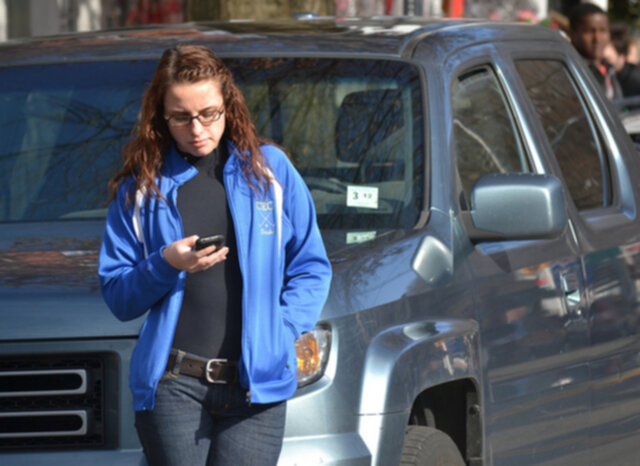}
 	
 	\noindent\textbf{Question:}is she wearing sneakers ?

 	\noindent\textbf{MRR candidate set}
 	\begin{enumerate}
 		\item ﻿\textcolor{purple}{no                   }\item ﻿\textcolor{orange}{not visible                  }\item ﻿\textcolor{red}{i ca n't see her feet              }\item ﻿\textcolor{purple}{ca n't tell                 }\item ﻿\textcolor{red}{i can not tell                }\item ﻿\textcolor{purple}{i ca n't tell                }\end{enumerate}
 	\noindent\textbf{Top 10 from the remaining NDCG candidates}
 	\begin{enumerate}
 		\item ca n't know                 \item nope                   \item i ca n't tell from the picture             \item can                   \item not sure                  \item i do n't think so               \item not that i can see               \item ca n't see the bottom part              \item maybe i am not sure               \item yes                   \end{enumerate}
 	﻿\includegraphics[width=1\linewidth]{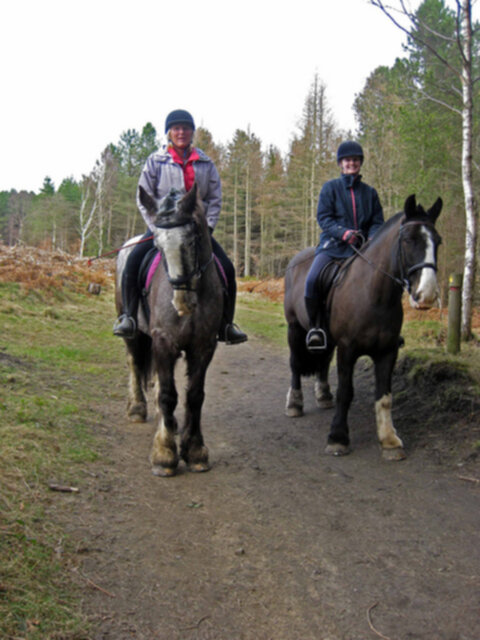}
 	
 	\noindent\textbf{Question:}is it sunny outside ?

 	\noindent\textbf{MRR candidate set}
 	\begin{enumerate}
 		\item ﻿\textcolor{orange}{yes                   }\item ﻿\textcolor{orange}{no                   }\item ﻿\textcolor{purple}{not really                  }\item ﻿\textcolor{purple}{no it is not                }\end{enumerate}
 	\noindent\textbf{Top 10 from the remaining NDCG candidates}
 	\begin{enumerate}
 		\item no , it is cloudy               \item it appears sunny                 \item no , it is not sunny outside             \item overcast                   \item nope                   \item it looks cloudy                 \item no cloudy                  \item it 's somewhat overcast                \item yes it is sunny                \item yes , looks sunny                \end{enumerate}
 	﻿\includegraphics[width=1\linewidth]{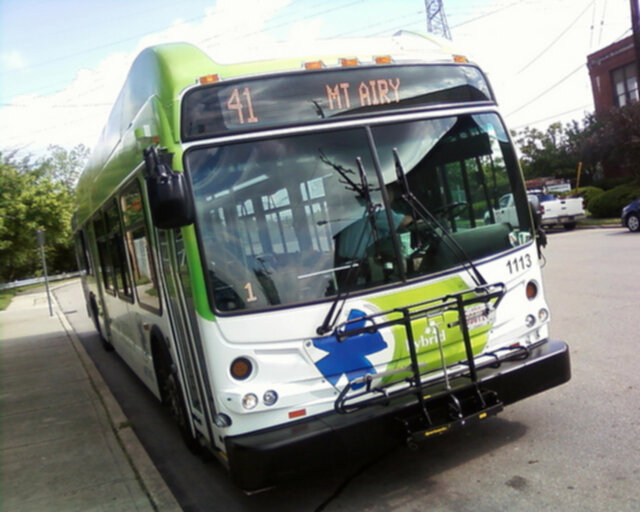}
 	
 	\noindent\textbf{Question:}are they in a parking lot or on the side of the road ?

 	\noindent\textbf{MRR candidate set}
 	\begin{enumerate}
 		\item ﻿\textcolor{orange}{on the side of the road              }\item ﻿\textcolor{orange}{yes                   }\item ﻿\textcolor{purple}{ca n't tell                 }\item ﻿\textcolor{purple}{i ca n't tell                }\item ﻿\textcolor{purple}{i can not tell                }\item ﻿\textcolor{purple}{not sure                  }\end{enumerate}
 	\noindent\textbf{Top 10 from the remaining NDCG candidates}
 	\begin{enumerate}
 		\item it 's hard to tell , but i can not see any        \item yes , many of them are lined up on the sidewalk         \item looks like it , ca n't tell             \item 1 is , yes                \item no , just 1                \item yes , it is                \item i think so                 \item no , they are not               \item on the ground                 \item yes it is a side view only             \end{enumerate}
 	﻿\includegraphics[width=1\linewidth]{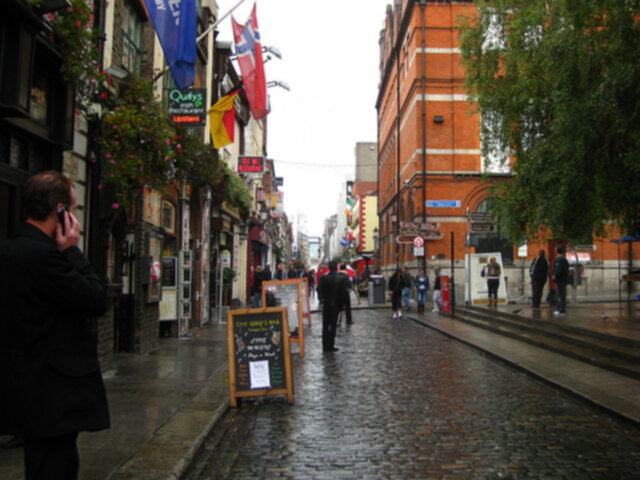}
 	
 	\noindent\textbf{Question:}is he holding it to his ear or using UNK ?

 	\noindent\textbf{MRR candidate set}
 	\begin{enumerate}
 		\item ﻿\textcolor{orange}{it 's up to his ear              }\item ﻿\textcolor{purple}{ca n't tell                 }\item ﻿\textcolor{purple}{i ca n't tell                }\item ﻿\textcolor{purple}{i can not tell                }\end{enumerate}
 	\noindent\textbf{Top 10 from the remaining NDCG candidates}
 	\begin{enumerate}
 		\item ca n't see                 \item not sure                  \item not that i can see               \item can you see                 \item no                   \item you ca n't tell                \item nope                   \item not really it 's kind of far away            \item both                   \item yes                   \end{enumerate}
 	﻿\includegraphics[width=1\linewidth]{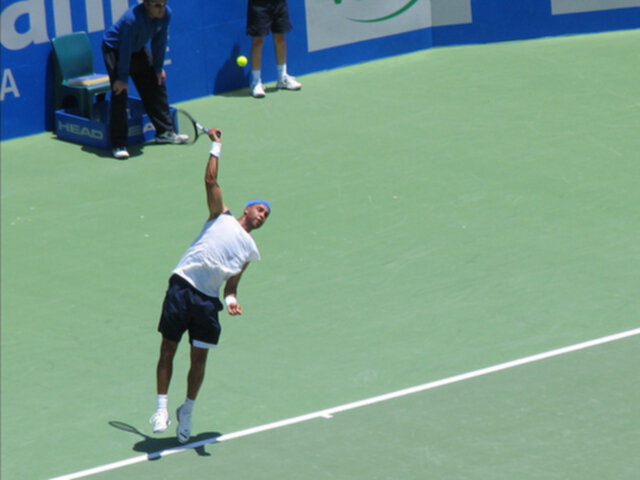}
 	
 	\noindent\textbf{Question:}is the man wearing a visor ?

 	\noindent\textbf{MRR candidate set}
 	\begin{enumerate}
 		\item ﻿\textcolor{orange}{no                   }\item ﻿\textcolor{orange}{yes                   }\item ﻿\textcolor{red}{no heat , but maybe more of a sweatband           }\item ﻿\textcolor{purple}{he is not                 }\item ﻿\textcolor{purple}{nope                   }\item ﻿\textcolor{purple}{no he is not                }\end{enumerate}
 	\noindent\textbf{Top 10 from the remaining NDCG candidates}
 	\begin{enumerate}
 		\item i do n't think so               \item yes , he does                \item yes he is                 \item not that i can see               \item yes !                  \item i believe so                 \item yep                   \item ca n't see that                \item can                   \item i think so                 \end{enumerate}
 	﻿\includegraphics[width=1\linewidth]{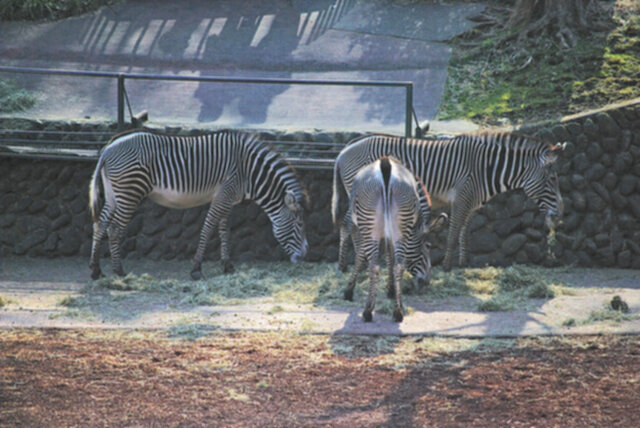}
 	
 	\noindent\textbf{Question:}could this be a zoo ?

 	\noindent\textbf{MRR candidate set}
 	\begin{enumerate}
 		\item ﻿\textcolor{orange}{yes                   }\item ﻿\textcolor{orange}{maybe                   }\item ﻿\textcolor{purple}{i think so                 }\item ﻿\textcolor{purple}{i would think so                }\item ﻿\textcolor{purple}{ca n't tell if it is              }\end{enumerate}
 	\noindent\textbf{Top 10 from the remaining NDCG candidates}
 	\begin{enumerate}
 		\item looks like it                 \item it looks like it                \item ca n't tell                 \item possibly , not sure                \item i ca n't tell                \item possibly , ca n't really tell              \item it looks like it is               \item not sure                  \item they appear to be                \item it 's hard to tell               \end{enumerate}
 	﻿\includegraphics[width=1\linewidth]{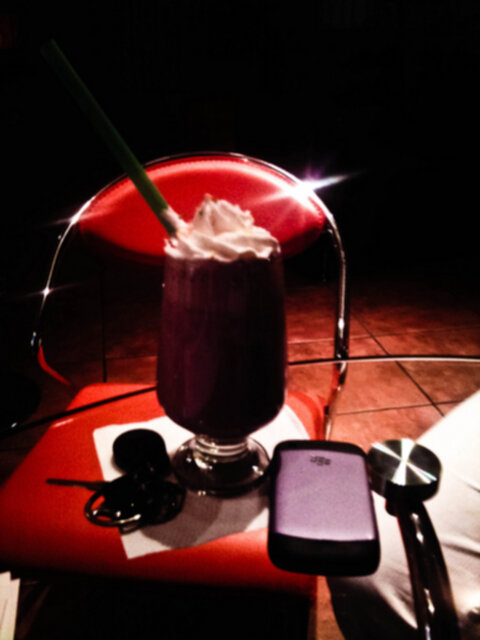}
 	
 	\noindent\textbf{Question:}is this a smartphone ?

 	\noindent\textbf{MRR candidate set}
 	\begin{enumerate}
 		\item ﻿\textcolor{orange}{yes                   }\item ﻿\textcolor{orange}{no                   }\item ﻿\textcolor{purple}{ca n't tell                 }\item ﻿\textcolor{purple}{i ca n't tell                }\item ﻿\textcolor{purple}{i do n't know                }\end{enumerate}
 	\noindent\textbf{Top 10 from the remaining NDCG candidates}
 	\begin{enumerate}
 		\item i can not tell                \item not sure                  \item hard to tell                 \item nope                   \item it does n't appear to be              \item i do not think so               \item no it is n't                \item i do n't think so               \item ca n't tell , maybe               \item no is not                 \end{enumerate}
 	﻿\includegraphics[width=1\linewidth]{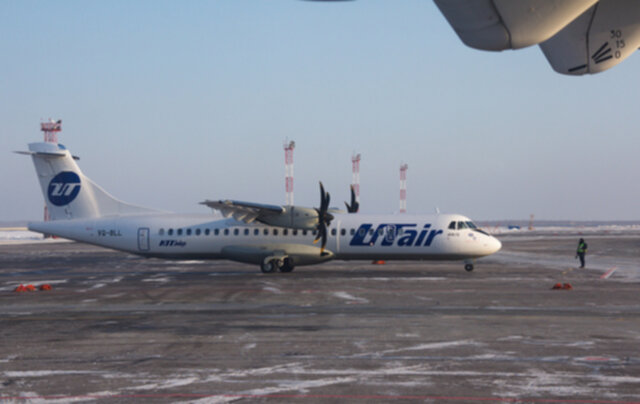}
 	
 	\noindent\textbf{Question:}are there words on the plane ?

 	\noindent\textbf{MRR candidate set}
 	\begin{enumerate}
 		\item ﻿\textcolor{orange}{yes , the plane says UNK              }\item ﻿\textcolor{purple}{yes                   }\item ﻿\textcolor{orange}{yes , there are words               }\item ﻿\textcolor{purple}{yes , there are                }\item ﻿\textcolor{purple}{yes there are                 }\item ﻿\textcolor{purple}{yes there is                 }\end{enumerate}
 	\noindent\textbf{Top 10 from the remaining NDCG candidates}
 	\begin{enumerate}
 		\item yes , it says bp               \item yes , quite a few               \item some yes                  \item yes , it is                \item i think so                 \item it appears to be a website i see words and an image but ca n't make them out  \item yes , i see some from the window            \item yes , many windows                \item yes , water                 \item yes it is                 \end{enumerate}
 	﻿\includegraphics[width=1\linewidth]{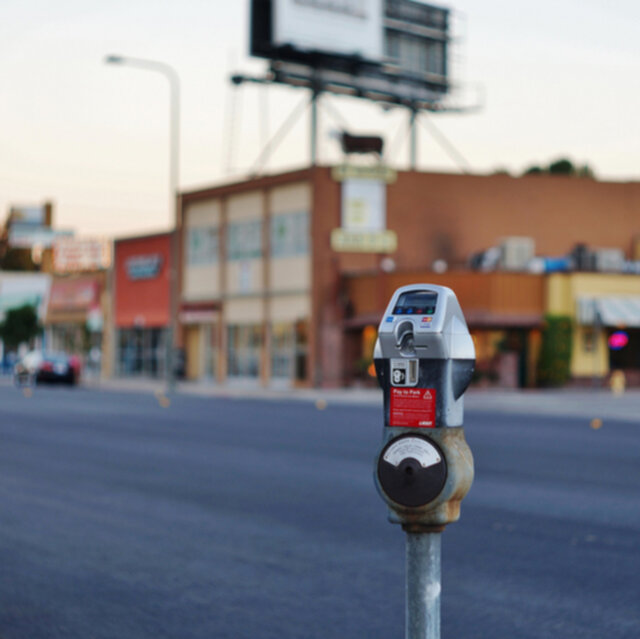}
 	
 	\noindent\textbf{Question:}can you see any store signs ?

 	\noindent\textbf{MRR candidate set}
 	\begin{enumerate}
 		\item ﻿\textcolor{orange}{no , it is blurry               }\item ﻿\textcolor{orange}{no                   }\item ﻿\textcolor{purple}{yes                   }\item ﻿\textcolor{purple}{no i do n't think so              }\item ﻿\textcolor{purple}{nope                   }\item ﻿\textcolor{purple}{not that i can see               }\end{enumerate}
 	\noindent\textbf{Top 10 from the remaining NDCG candidates}
 	\begin{enumerate}
 		\item there are 0 of those               \item in the distance                 \item yes in the background                \item not really                  \item yes , several in the background              \item i do n't think so               \item yes some                  \item barely                   \item yes quite a few in the background             \item 1 partially off in the distance              \end{enumerate}
 	﻿\includegraphics[width=1\linewidth]{./VisualDialog_test2018/VisualDialog_test2018_000000168655.jpg}
 	
 	\noindent\textbf{Question:}can you see inside the bus ?

 	\noindent\textbf{MRR candidate set}
 	\begin{enumerate}
 		\item ﻿\textcolor{orange}{no , just the back of the bus            }\item ﻿\textcolor{orange}{no                   }\item ﻿\textcolor{purple}{i can not                 }\item ﻿\textcolor{red}{i can not tell                }\item ﻿\textcolor{purple}{no i ca n't                }\item ﻿\textcolor{purple}{nope                   }\end{enumerate}
 	\noindent\textbf{Top 10 from the remaining NDCG candidates}
 	\begin{enumerate}
 		\item no '                  \item no , i do n't               \item not really                  \item yes                   \item barely                   \item i do n't think so               \item yes , i can                \item not it 's not visible               \item not that i can see               \item i ca n't tell                \end{enumerate}
 	﻿\includegraphics[width=1\linewidth]{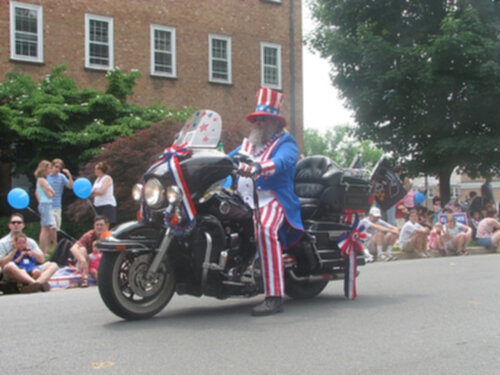}
 	
 	\noindent\textbf{Question:}is the man wearing a mask ?

 	\noindent\textbf{MRR candidate set}
 	\begin{enumerate}
 		\item ﻿\textcolor{orange}{no                   }\item ﻿\textcolor{red}{yes                   }\item ﻿\textcolor{purple}{no he does not                }\item ﻿\textcolor{purple}{no , he is not               }\item ﻿\textcolor{purple}{nope                   }\end{enumerate}
 	\noindent\textbf{Top 10 from the remaining NDCG candidates}
 	\begin{enumerate}
 		\item i do n't think so               \item does n't look like it               \item not that i can see               \item not really                  \item no , he 's wearing a helmet             \item yes indeed                  \item he is                  \item yes he is                 \item i do n't think he is , hard to tell with most of his head down    \item it looks like a helmet               \end{enumerate}
 	﻿\includegraphics[width=1\linewidth]{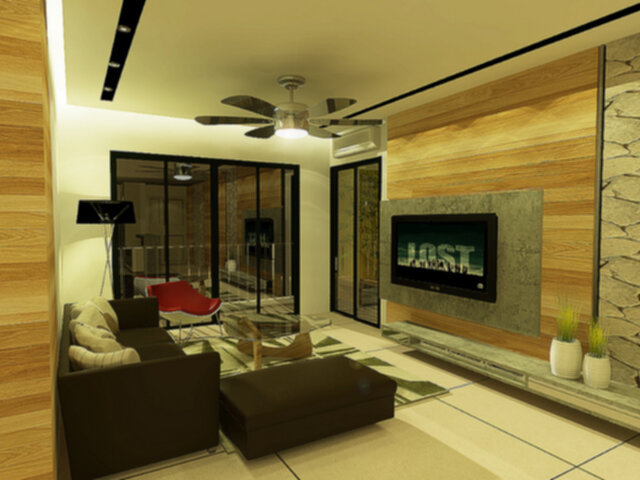}
 	
 	\noindent\textbf{Question:}does the room look big ?

 	\noindent\textbf{MRR candidate set}
 	\begin{enumerate}
 		\item ﻿\textcolor{red}{it is decent size                }\item ﻿\textcolor{red}{yes , it does                }\item ﻿\textcolor{purple}{yes                   }\item ﻿\textcolor{purple}{i 'd say so                }\item ﻿\textcolor{purple}{yes , very                 }\end{enumerate}
 	\noindent\textbf{Top 10 from the remaining NDCG candidates}
 	\begin{enumerate}
 		\item looks like it                 \item it looks like                 \item not too big                 \item no , it is pretty small              \item yes , it is                \item average                   \item it could be                 \item i think it                 \item not really                  \item very                   \end{enumerate}
 	﻿\includegraphics[width=1\linewidth]{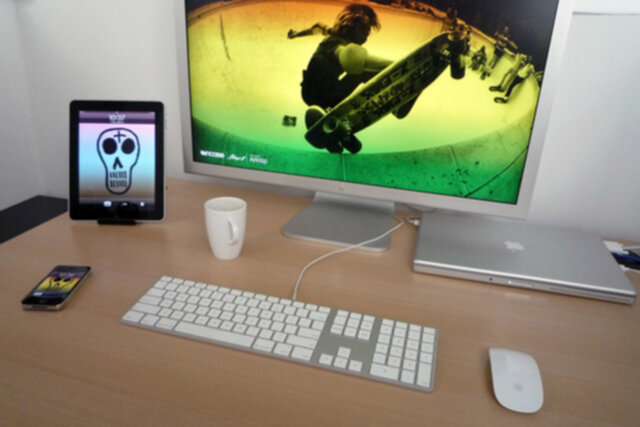}
 	
 	\noindent\textbf{Question:}is the laptop on a desk ?

 	\noindent\textbf{MRR candidate set}
 	\begin{enumerate}
 		\item ﻿\textcolor{orange}{yes                   }\item ﻿\textcolor{orange}{a table                  }\item ﻿\textcolor{purple}{yes it is                 }\item ﻿\textcolor{purple}{yes , it is                }\item ﻿\textcolor{purple}{it appears to be                }\item ﻿\textcolor{purple}{yes it looks like it is              }\end{enumerate}
 	\noindent\textbf{Top 10 from the remaining NDCG candidates}
 	\begin{enumerate}
 		\item looks like it                 \item i think so                 \item on a portable desk thing , like a lap desk          \item yes it is on                \item yes , just 1                \item no , it looks like it is on a table          \item no on a stand                \item no a stand                 \item yes , he she is               \item yes , there is something below the tv            \end{enumerate}
 	﻿\includegraphics[width=1\linewidth]{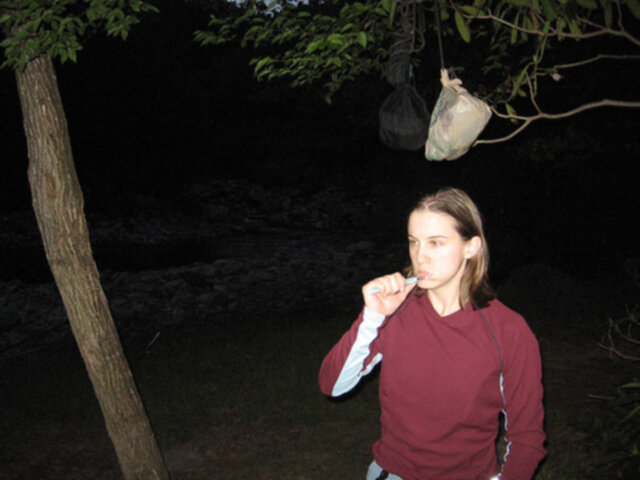}
 	
 	\noindent\textbf{Question:}what color is her hair ?

 	\noindent\textbf{MRR candidate set}
 	\begin{enumerate}
 		\item ﻿\textcolor{orange}{brown                   }\item ﻿\textcolor{orange}{blonde                   }\item ﻿\textcolor{purple}{dark blonde                  }\item ﻿\textcolor{purple}{it 's brown                 }\item ﻿\textcolor{purple}{her hair is brown                }\item ﻿\textcolor{purple}{she has brown hair                }\end{enumerate}
 	\noindent\textbf{Top 10 from the remaining NDCG candidates}
 	\begin{enumerate}
 		\item dark brown                  \item dirty blonde                  \item dark , brown                 \item i think it brown                \item brown , i think                \item it 's dirty blonde                \item black                   \item red                   \item UNK rose UNK                 \item looks dirty                  \end{enumerate}
 	﻿\includegraphics[width=1\linewidth]{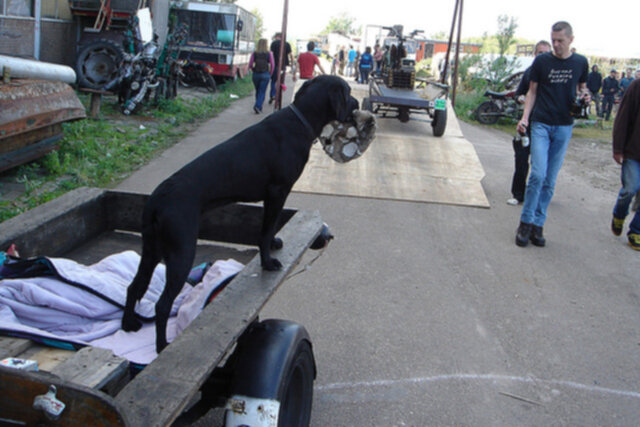}
 	
 	\noindent\textbf{Question:}what color is the dog ?

 	\noindent\textbf{MRR candidate set}
 	\begin{enumerate}
 		\item ﻿\textcolor{orange}{black                   }\item ﻿\textcolor{purple}{black lab                  }\item ﻿\textcolor{purple}{the dog is black                }\item ﻿\textcolor{purple}{it is black                 }\item ﻿\textcolor{purple}{dark                   }\end{enumerate}
 	\noindent\textbf{Top 10 from the remaining NDCG candidates}
 	\begin{enumerate}
 		\item black and white                 \item dark brown , beautiful                \item it 's dark                 \item not sure                  \item black and brown                 \item i can not tell                \item 1 is black , and the other is brown           \item it 's in a shadow but maybe black            \item black and white feet                \item i ca n't tell                \end{enumerate}
 	﻿\includegraphics[width=1\linewidth]{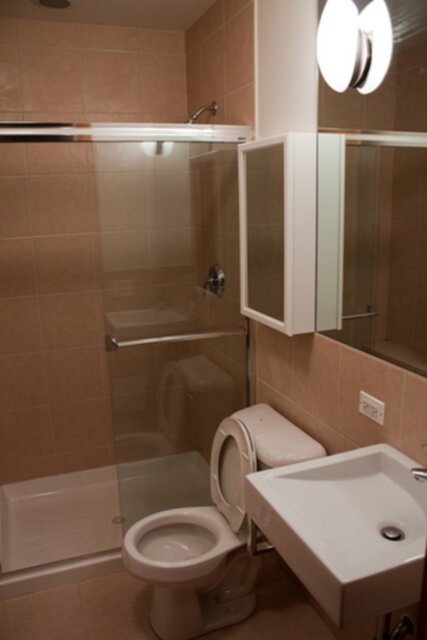}
 	
 	\noindent\textbf{Question:}any walls ?

 	\noindent\textbf{MRR candidate set}
 	\begin{enumerate}
 		\item ﻿\textcolor{orange}{yes                   }\item ﻿\textcolor{red}{yes , it is                }\item ﻿\textcolor{purple}{yes a large 1                }\item ﻿\textcolor{purple}{yess                   }\end{enumerate}
 	\noindent\textbf{Top 10 from the remaining NDCG candidates}
 	\begin{enumerate}
 		\item yes 1                  \item yes it is                 \item hello it 's mostly brown               \item UNK                   \item yes , to the left               \item i think so                 \item no                   \item barely part of 1                \item is pink                  \item not that i can see               \end{enumerate}
 	﻿\includegraphics[width=1\linewidth]{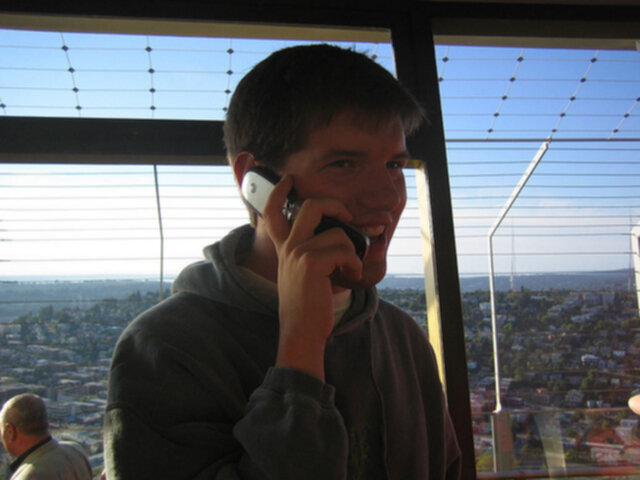}
 	
 	\noindent\textbf{Question:}are the buildings tall ?

 	\noindent\textbf{MRR candidate set}
 	\begin{enumerate}
 		\item ﻿\textcolor{orange}{it is hard to tell from the picture            }\item ﻿\textcolor{orange}{i can not tell                }\item ﻿\textcolor{red}{in the background barely visible               }\item ﻿\textcolor{purple}{i ca n't tell                }\item ﻿\textcolor{purple}{hard to tell                 }\item ﻿\textcolor{purple}{ca n't tell                 }\end{enumerate}
 	\noindent\textbf{Top 10 from the remaining NDCG candidates}
 	\begin{enumerate}
 		\item they do not appear to be              \item possible hard to tell                \item yes , they appear to be              \item not sure                  \item yes                   \item possibly                   \item no                   \item i do n't think so               \item not really                  \item i think so                 \end{enumerate}
 	﻿\includegraphics[width=1\linewidth]{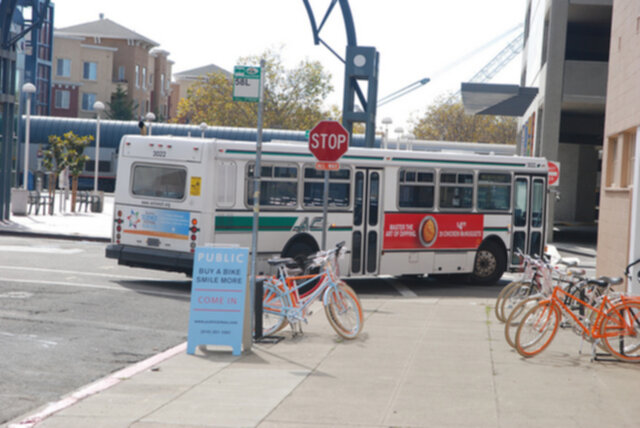}
 	
 	\noindent\textbf{Question:}can you see any trees ?

 	\noindent\textbf{MRR candidate set}
 	\begin{enumerate}
 		\item ﻿\textcolor{orange}{yes                   }\item ﻿\textcolor{purple}{yes in the background                }\item ﻿\textcolor{purple}{yes , a few in the background             }\item ﻿\textcolor{purple}{yes , in the background               }\item ﻿\textcolor{purple}{yep                   }\end{enumerate}
 	\noindent\textbf{Top 10 from the remaining NDCG candidates}
 	\begin{enumerate}
 		\item yes some                  \item in the distance , a few              \item in the background                 \item there are several in the background              \item there are a few large trees near by and several in background        \item yes , there are a lot of trees            \item ues                   \item some                   \item yes , far in the background              \item yes 1                  \end{enumerate}
 	﻿\includegraphics[width=1\linewidth]{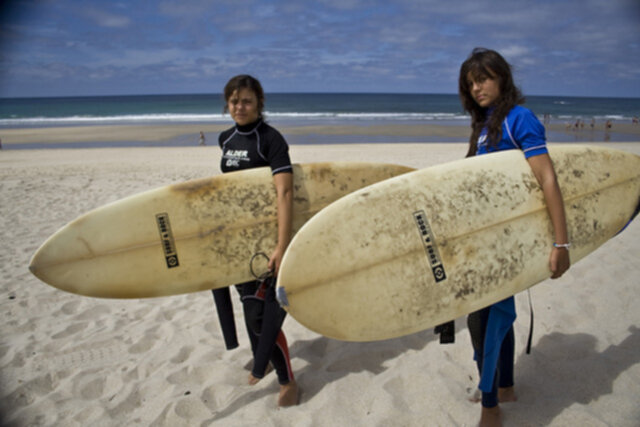}
 	
 	\noindent\textbf{Question:}what is the gender of the people ?

 	\noindent\textbf{MRR candidate set}
 	\begin{enumerate}
 		\item ﻿\textcolor{orange}{both boys                  }\item ﻿\textcolor{red}{female                   }\item ﻿\textcolor{purple}{hard to say maybe male and female             }\item ﻿\textcolor{purple}{not sure                  }\end{enumerate}
 	\noindent\textbf{Top 10 from the remaining NDCG candidates}
 	\begin{enumerate}
 		\item they look caucasian                 \item looks male                  \item yes                   \item it appears so                 \item ca n't tell                 \item they are both posing for the picture 1 posing as if he 's going to hit the tennis ball the\item i can not tell                \item no 1 is on 1st , 2nd and 3rd are not visible        \item i ca n't tell                \item yes , it is                \end{enumerate}
 	﻿\includegraphics[width=1\linewidth]{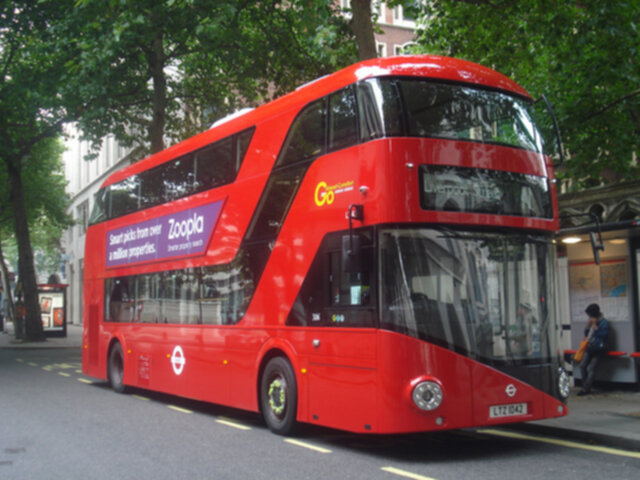}
 	
 	\noindent\textbf{Question:}does the bus have a number ?

 	\noindent\textbf{MRR candidate set}
 	\begin{enumerate}
 		\item ﻿\textcolor{orange}{no                   }\item ﻿\textcolor{orange}{yes                   }\item ﻿\textcolor{purple}{i ca n't see 1               }\item ﻿\textcolor{purple}{no that i can see               }\item ﻿\textcolor{purple}{yes it does                 }\end{enumerate}
 	\noindent\textbf{Top 10 from the remaining NDCG candidates}
 	\begin{enumerate}
 		\item yes ,                  \item a number pad yes                \item not that i can see               \item nope                   \item do n't see 1                \item i think so                 \item does n't look like it               \item can not tell not visible               \item no it does n't                \item 2                   \end{enumerate}
 	﻿\includegraphics[width=1\linewidth]{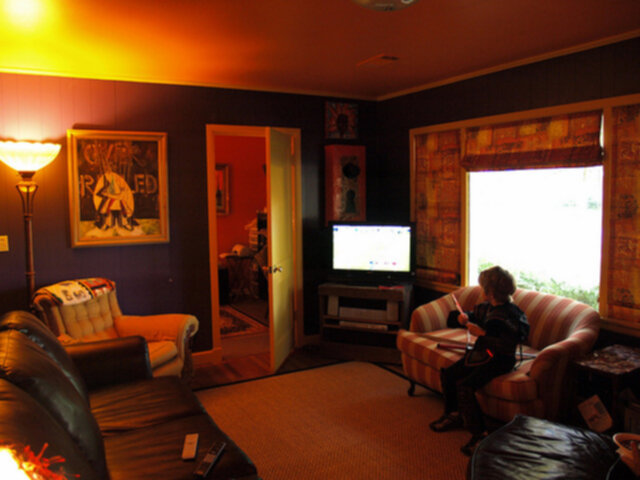}
 	
 	\noindent\textbf{Question:}what kind of toy is the boy playing with ?

 	\noindent\textbf{MRR candidate set}
 	\begin{enumerate}
 		\item ﻿\textcolor{orange}{can not tell which toy it is             }\item ﻿\textcolor{orange}{look to be a wii video game             }\item ﻿\textcolor{purple}{i can not tell                }\item ﻿\textcolor{purple}{i ca n't tell                }\item ﻿\textcolor{purple}{ca n't tell                 }\item ﻿\textcolor{purple}{i m not sure                }\end{enumerate}
 	\noindent\textbf{Top 10 from the remaining NDCG candidates}
 	\begin{enumerate}
 		\item you can not see who he is playing with           \item i do n't know                \item a small remote controlled 1               \item not sure                  \item i see 1 , but its far             \item a tiny bird                 \item kind of a derby i think it 's called           \item i ca n't see the ball              \item a bowl                  \item maybe a UNK                 \end{enumerate}
 	﻿\includegraphics[width=1\linewidth]{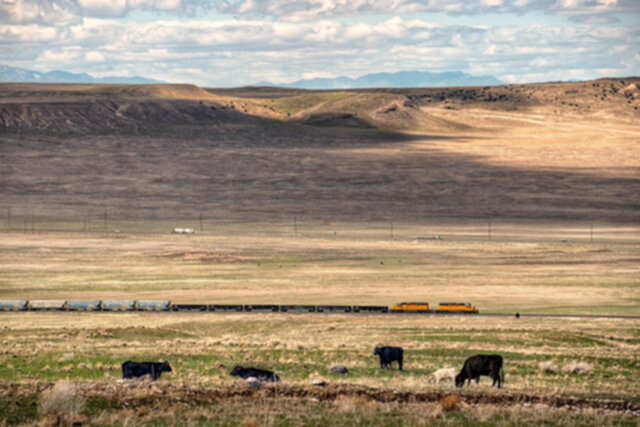}
 	
 	\noindent\textbf{Question:}is the sky cloudy ?

 	\noindent\textbf{MRR candidate set}
 	\begin{enumerate}
 		\item ﻿\textcolor{orange}{yes                   }\item ﻿\textcolor{purple}{yes , blue skies some clouds              }\item ﻿\textcolor{purple}{yes it is                 }\item ﻿\textcolor{purple}{yes , it is                }\end{enumerate}
 	\noindent\textbf{Top 10 from the remaining NDCG candidates}
 	\begin{enumerate}
 		\item yes '                  \item yes but it is blue sky              \item yeah pretty much                 \item yes , it is very clear              \item yes very clear                 \item yes partially                  \item i think so                 \item it appears overcast                 \item a little bit                 \item relatively                   \end{enumerate}
 	﻿\includegraphics[width=1\linewidth]{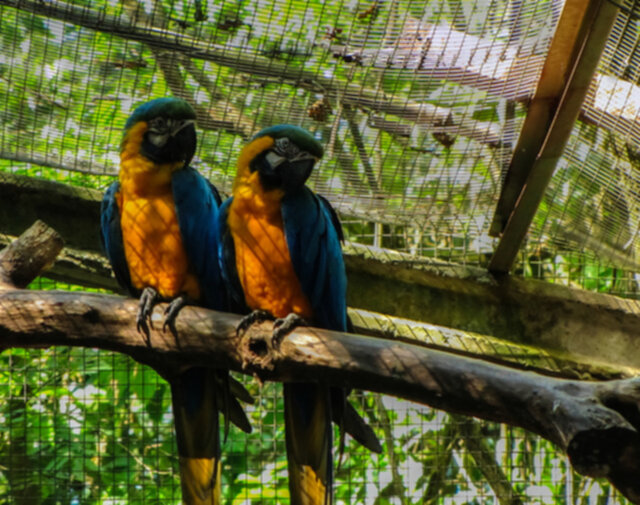}
 	
 	\noindent\textbf{Question:}is the cage made of metal ?

 	\noindent\textbf{MRR candidate set}
 	\begin{enumerate}
 		\item ﻿\textcolor{orange}{yes                   }\item ﻿\textcolor{red}{metal and wood - looks like a zoo enclosure           }\item ﻿\textcolor{purple}{yes it is                 }\item ﻿\textcolor{purple}{it appears to be                }\item ﻿\textcolor{purple}{yes , it is                }\end{enumerate}
 	\noindent\textbf{Top 10 from the remaining NDCG candidates}
 	\begin{enumerate}
 		\item yes it is ,                \item i think so                 \item yep                   \item wood                   \item hmm i think so                \item no                   \item i ca n't really tell               \item they are and have metal parts too             \item i ca n't tell                \item i can not tell                \end{enumerate}
 	﻿\includegraphics[width=1\linewidth]{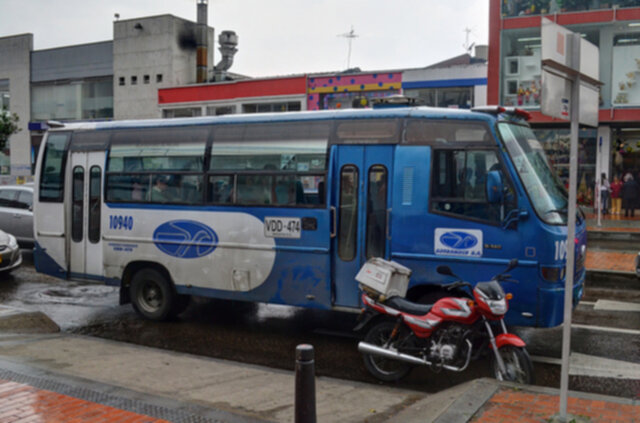}
 	
 	\noindent\textbf{Question:}do any of them have their hair up ?

 	\noindent\textbf{MRR candidate set}
 	\begin{enumerate}
 		\item ﻿\textcolor{orange}{no                   }\item ﻿\textcolor{purple}{not that i can see               }\item ﻿\textcolor{purple}{no they do n't                }\item ﻿\textcolor{purple}{i ca n't tell                }\item ﻿\textcolor{purple}{nope                   }\end{enumerate}
 	\noindent\textbf{Top 10 from the remaining NDCG candidates}
 	\begin{enumerate}
 		\item ca n't tell                 \item i do n't think so               \item i can not tell                \item i do n't know                \item not sure                  \item too far away in photo to tell             \item hard to tell since they 're far away            \item the picture does not show 1              \item 0                   \item maybe , not sure                \end{enumerate}
 	﻿\includegraphics[width=1\linewidth]{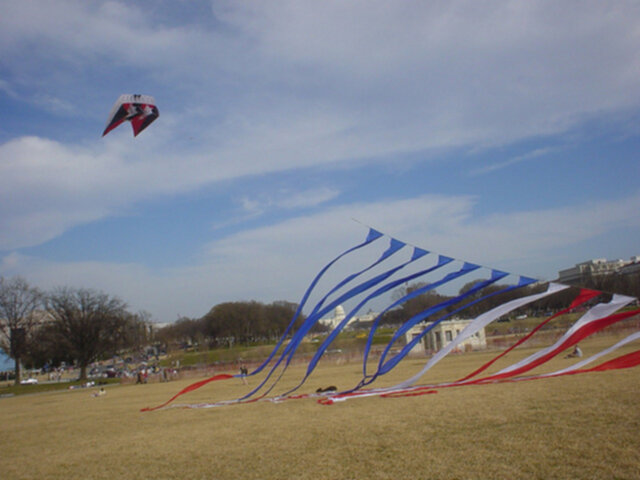}
 	
 	\noindent\textbf{Question:}is there other kites in the sky or just the 1 ?

 	\noindent\textbf{MRR candidate set}
 	\begin{enumerate}
 		\item ﻿\textcolor{orange}{there is 1 other kite               }\item ﻿\textcolor{purple}{1                   }\item ﻿\textcolor{purple}{i can only see 1 in this picture            }\item ﻿\textcolor{purple}{not that i can see               }\end{enumerate}
 	\noindent\textbf{Top 10 from the remaining NDCG candidates}
 	\begin{enumerate}
 		\item no there are n't                \item no                   \item nope                   \item yes a few                 \item yes                   \item just beside it                 \item yes , both                 \item no , no 1 else in the photo            \item in the air                 \item no just her                 \end{enumerate}
 	﻿\includegraphics[width=1\linewidth]{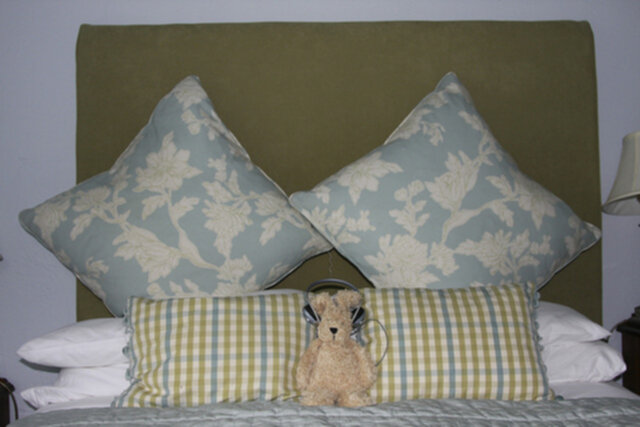}
 	
 	\noindent\textbf{Question:}is it a close up of the bear ?

 	\noindent\textbf{MRR candidate set}
 	\begin{enumerate}
 		\item ﻿\textcolor{orange}{yes                   }\item ﻿\textcolor{purple}{yes , it is                }\item ﻿\textcolor{purple}{yes , pretty close                }\item ﻿\textcolor{purple}{yes it is                 }\item ﻿\textcolor{purple}{it is , yes                }\end{enumerate}
 	\noindent\textbf{Top 10 from the remaining NDCG candidates}
 	\begin{enumerate}
 		\item sort of                  \item i think so                 \item it look like it                \item a little bit , yes               \item no                   \item not really                  \item yes there is                 \item nope                   \item yes , it is very clear              \item a few inches or so               \end{enumerate}
 	﻿\includegraphics[width=1\linewidth]{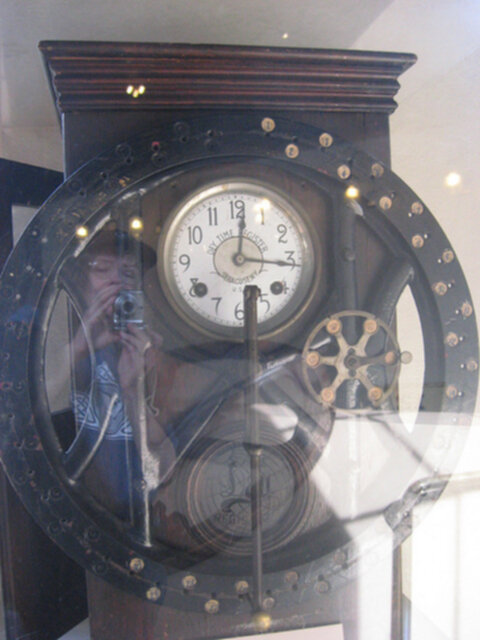}
 	
 	\noindent\textbf{Question:}is the shelf wood ?

 	\noindent\textbf{MRR candidate set}
 	\begin{enumerate}
 		\item ﻿\textcolor{orange}{yes                   }\item ﻿\textcolor{red}{i can not tell                }\item ﻿\textcolor{purple}{yes , it is                }\item ﻿\textcolor{purple}{i think so                 }\item ﻿\textcolor{purple}{yes , i think so               }\item ﻿\textcolor{purple}{yes it is                 }\end{enumerate}
 	\noindent\textbf{Top 10 from the remaining NDCG candidates}
 	\begin{enumerate}
 		\item no                   \item looks like it                 \item no it 's not                \item ca n't tell                 \item i ca n't tell                \item i think it is , but hard to say for sure         \item not sure                  \item it is hard to tell               \item yes it is brown                \item i do n't think so               \end{enumerate}
 	﻿\includegraphics[width=1\linewidth]{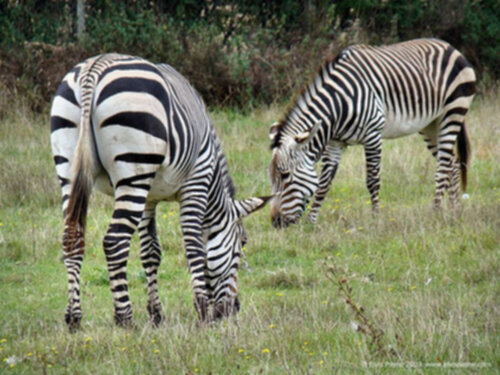}
 	
 	\noindent\textbf{Question:}can you see any roads ?

 	\noindent\textbf{MRR candidate set}
 	\begin{enumerate}
 		\item ﻿\textcolor{orange}{no                   }\item ﻿\textcolor{orange}{nope                   }\item ﻿\textcolor{purple}{no i ca n't                }\item ﻿\textcolor{purple}{i do n't                 }\item ﻿\textcolor{purple}{not that i can see               }\end{enumerate}
 	\noindent\textbf{Top 10 from the remaining NDCG candidates}
 	\begin{enumerate}
 		\item no you can not                \item 0                   \item np                   \item i do n't think so               \item not really                  \item nobody around                  \item no , there are no signs              \item no buildings                  \item no , it 's sunny               \item no it 's too far away              \end{enumerate}
 	﻿\includegraphics[width=1\linewidth]{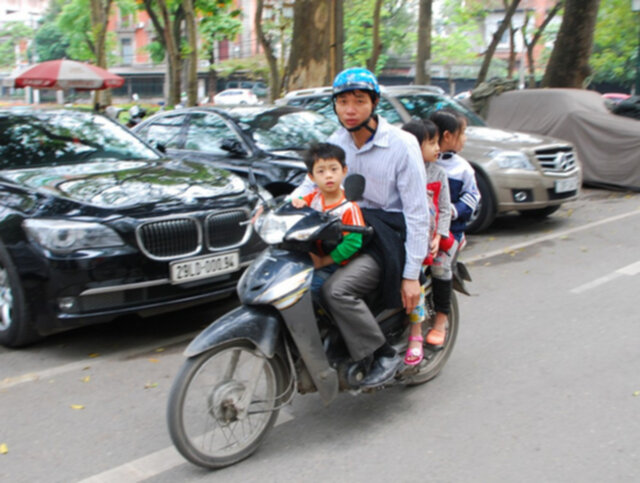}
 	
 	\noindent\textbf{Question:}is it sunny ?

 	\noindent\textbf{MRR candidate set}
 	\begin{enumerate}
 		\item ﻿\textcolor{orange}{yes                   }\item ﻿\textcolor{red}{yes , it is                }\item ﻿\textcolor{purple}{yes it is sunny                }\item ﻿\textcolor{purple}{it looks overcast , but i ca n't see the sky         }\item ﻿\textcolor{purple}{it appears to be sunny               }\item ﻿\textcolor{purple}{yes , its daytime and it looks sunny            }\item ﻿\textcolor{purple}{no , it 's looks overcast              }\end{enumerate}
 	\noindent\textbf{Top 10 from the remaining NDCG candidates}
 	\begin{enumerate}
 		\item no it looks cloudy                \item does n't really seem to be              \item yes it is                 \item no                   \item it seems like it                \item looks like it is yes               \item appears to be                 \item it 's overcast                 \item not really                  \item i believe it is                \end{enumerate}
 	﻿\includegraphics[width=1\linewidth]{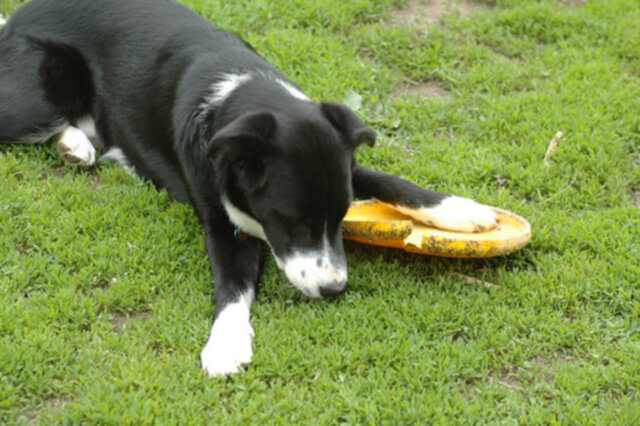}
 	
 	\noindent\textbf{Question:}do see any cars ?

 	\noindent\textbf{MRR candidate set}
 	\begin{enumerate}
 		\item ﻿\textcolor{orange}{no there are no cars               }\item ﻿\textcolor{orange}{no cars                  }\item ﻿\textcolor{purple}{no                   }\item ﻿\textcolor{purple}{no , not at all               }\end{enumerate}
 	\noindent\textbf{Top 10 from the remaining NDCG candidates}
 	\begin{enumerate}
 		\item nope                   \item there is 0                 \item 0 that i can see               \item not that i can see               \item 0 visible                  \item 0                   \item no , ca n't see the street             \item not really                  \item i do n't think so               \item no , she is not               \end{enumerate}
 	﻿\includegraphics[width=1\linewidth]{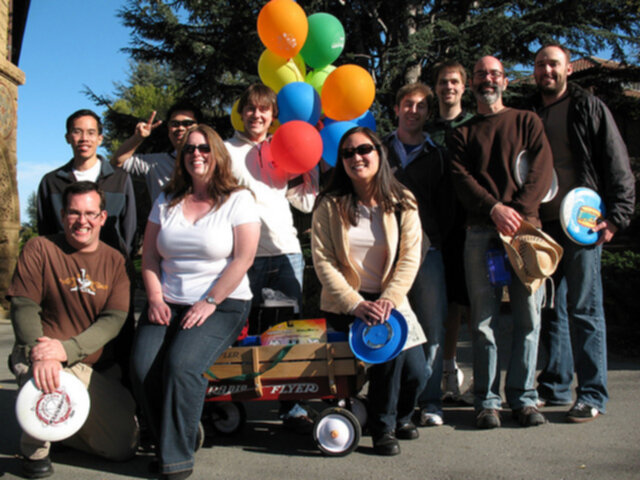}
 	
 	\noindent\textbf{Question:}are they outside ?

 	\noindent\textbf{MRR candidate set}
 	\begin{enumerate}
 		\item ﻿\textcolor{orange}{yes                   }\item ﻿\textcolor{purple}{yes , they 're outside               }\item ﻿\textcolor{purple}{yes , they are                }\item ﻿\textcolor{purple}{yes they are                 }\item ﻿\textcolor{purple}{it looks like it                }\end{enumerate}
 	\noindent\textbf{Top 10 from the remaining NDCG candidates}
 	\begin{enumerate}
 		\item it seems like it                \item i think so                 \item i believe so                 \item yes , it is                \item yes it is outside                \item probably                   \item looks like                  \item yes , on a street corner              \item it looks outside                 \item yes it is                 \end{enumerate}
 	﻿\includegraphics[width=1\linewidth]{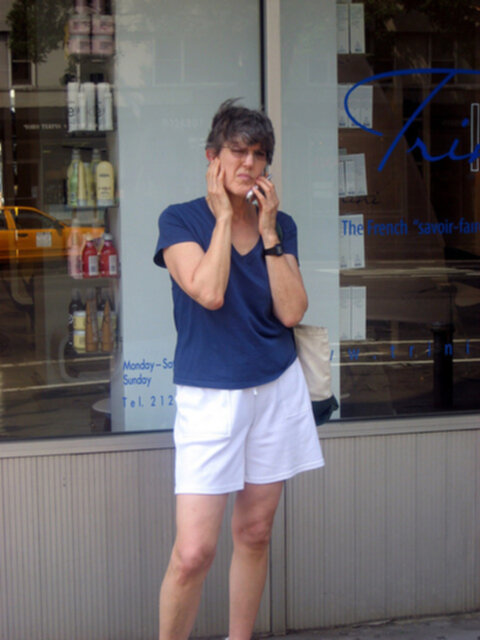}
 	
 	\noindent\textbf{Question:}what color are her shorts ?

 	\noindent\textbf{MRR candidate set}
 	\begin{enumerate}
 		\item ﻿\textcolor{orange}{white                   }\item ﻿\textcolor{orange}{they are white                 }\item ﻿\textcolor{orange}{her pants are light blue               }\item ﻿\textcolor{purple}{blue                   }\item ﻿\textcolor{purple}{salmon                   }\end{enumerate}
 	\noindent\textbf{Top 10 from the remaining NDCG candidates}
 	\begin{enumerate}
 		\item looks like summer                 \item looks blue                  \item khaki                   \item UNK bright blue                 \item i think blue , hard to tell             \item mauve                   \item grey                   \item yes                   \item blue yellow and white                \item blue and black                 \end{enumerate}
 	﻿\includegraphics[width=1\linewidth]{./VisualDialog_test2018/VisualDialog_test2018_000000287909.jpg}
 	
 	\noindent\textbf{Question:}is it sunny outdoors ?

 	\noindent\textbf{MRR candidate set}
 	\begin{enumerate}
 		\item ﻿\textcolor{orange}{yes                   }\item ﻿\textcolor{orange}{yes it is                 }\item ﻿\textcolor{purple}{yes , it is                }\item ﻿\textcolor{purple}{yes , it is a sunny day             }\item ﻿\textcolor{purple}{yes sunny                  }\end{enumerate}
 	\noindent\textbf{Top 10 from the remaining NDCG candidates}
 	\begin{enumerate}
 		\item it is                  \item very                   \item yes , the weather looks nice              \item yes the sun is shining               \item yes appears so                 \item it appears to be sunny               \item it seems to be sunny               \item seems like it                 \item yes , but i can not see the sun           \item yrd                   \end{enumerate}
 	﻿\includegraphics[width=1\linewidth]{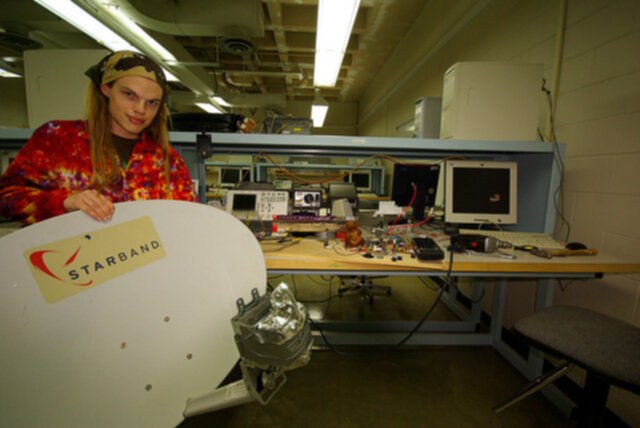}
 	
 	\noindent\textbf{Question:}do you see any lights ?

 	\noindent\textbf{MRR candidate set}
 	\begin{enumerate}
 		\item ﻿\textcolor{orange}{no                   }\item ﻿\textcolor{orange}{yes                   }\item ﻿\textcolor{purple}{yes i see 2 lights               }\item ﻿\textcolor{purple}{i think so                 }\item ﻿\textcolor{purple}{yes many                  }\end{enumerate}
 	\noindent\textbf{Top 10 from the remaining NDCG candidates}
 	\begin{enumerate}
 		\item 1 in the background                \item no i do n't                \item no , i do n't               \item nope                   \item yes , 1 big 1               \item yes , some on the floor              \item i do n't think so               \item 'no i do n't                \item not that i can see               \item yes above the sink                \end{enumerate}
 	﻿\includegraphics[width=1\linewidth]{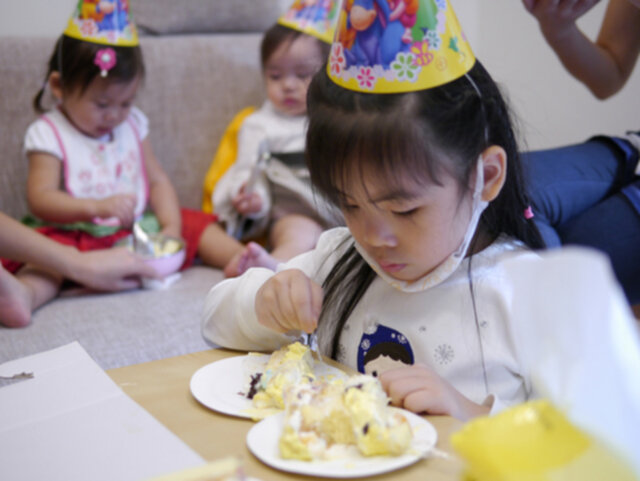}
 	
 	\noindent\textbf{Question:}are they outside ?

 	\noindent\textbf{MRR candidate set}
 	\begin{enumerate}
 		\item ﻿\textcolor{orange}{no                   }\item ﻿\textcolor{orange}{yes                   }\item ﻿\textcolor{purple}{i do n't think so               }\item ﻿\textcolor{purple}{i ca n't tell                }\item ﻿\textcolor{purple}{nope                   }\end{enumerate}
 	\noindent\textbf{Top 10 from the remaining NDCG candidates}
 	\begin{enumerate}
 		\item i am not sure                \item ca n't tell                 \item hard to tell                 \item no , it 's indoors               \item i can not tell                \item i do n't think they are based on their clothing          \item not that i can see               \item not sure                  \item ca n't tell - maybe               \item i think so                 \end{enumerate}
 	﻿\includegraphics[width=1\linewidth]{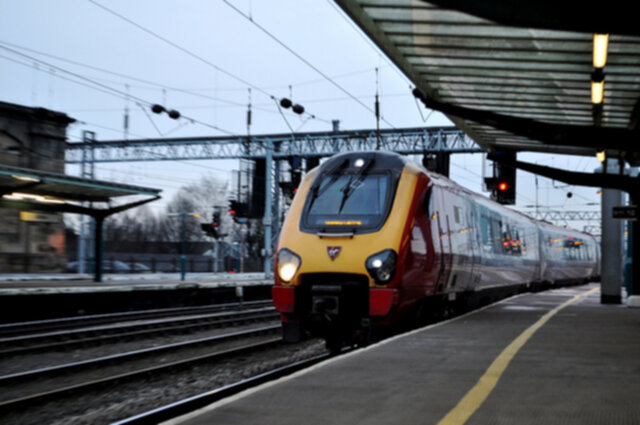}
 	
 	\noindent\textbf{Question:}is it a passenger train ?

 	\noindent\textbf{MRR candidate set}
 	\begin{enumerate}
 		\item ﻿\textcolor{orange}{yes                   }\item ﻿\textcolor{orange}{no                   }\item ﻿\textcolor{purple}{yes it is                 }\item ﻿\textcolor{purple}{yes , it is                }\item ﻿\textcolor{purple}{yes , i think so               }\end{enumerate}
 	\noindent\textbf{Top 10 from the remaining NDCG candidates}
 	\begin{enumerate}
 		\item yes , it looks like it              \item it appears to be                \item i think so                 \item it looks like it , hard to tell            \item yes , it 's big               \item could be                  \item i ca n't tell for sure , but i think so         \item maybe                   \item UNK                   \item 2                   \end{enumerate}
 	﻿\includegraphics[width=1\linewidth]{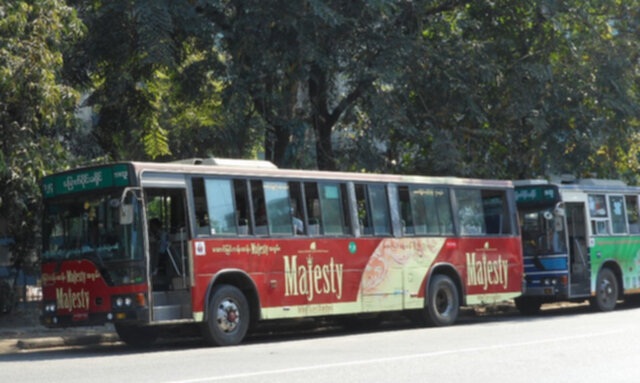}
 	
 	\noindent\textbf{Question:}is the bus doors open or closed ?

 	\noindent\textbf{MRR candidate set}
 	\begin{enumerate}
 		\item ﻿\textcolor{orange}{open                   }\item ﻿\textcolor{red}{closed                   }\item ﻿\textcolor{red}{i can not tell                }\item ﻿\textcolor{purple}{i ca n't tell                }\item ﻿\textcolor{purple}{it 's open                 }\end{enumerate}
 	\noindent\textbf{Top 10 from the remaining NDCG candidates}
 	\begin{enumerate}
 		\item ca n't tell                 \item no doors                  \item both are open                 \item not sure                  \item ca n't really tell                \item yes it is open                \item no                   \item yes                   \item no , closed                 \item seems to be                 \end{enumerate}
 	﻿\includegraphics[width=1\linewidth]{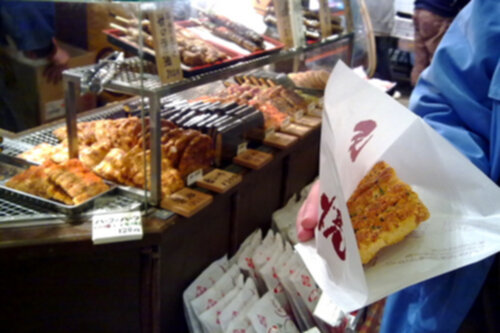}
 	
 	\noindent\textbf{Question:}describe the person holding the treat ?

 	\noindent\textbf{MRR candidate set}
 	\begin{enumerate}
 		\item ﻿\textcolor{orange}{ca n't see the person               }\item ﻿\textcolor{red}{i can only see their arm they are wearing blue          }\item ﻿\textcolor{red}{all i can see is a hand but probably a male         }\item ﻿\textcolor{purple}{i can not tell                }\item ﻿\textcolor{purple}{ca n't tell from the picture              }\item ﻿\textcolor{purple}{i am unable to tell               }\end{enumerate}
 	\noindent\textbf{Top 10 from the remaining NDCG candidates}
 	\begin{enumerate}
 		\item i can not see                \item i ca n't tell                \item not visible not seen                \item i ca n't tell his back is towards me           \item ca n't tell                 \item he looks confused                 \item not sure                  \item woman                   \item UNK smiling                  \item the shirt is striped in black and white            \end{enumerate}
 	﻿\includegraphics[width=1\linewidth]{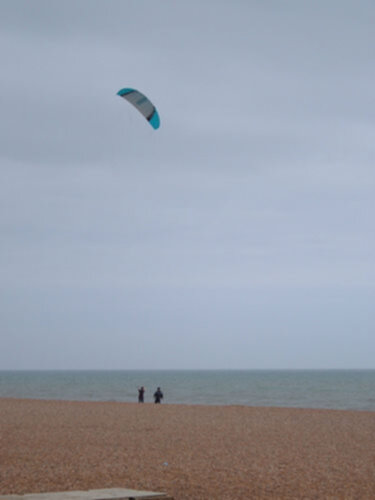}
 	
 	\noindent\textbf{Question:}how many people are there ?

 	\noindent\textbf{MRR candidate set}
 	\begin{enumerate}
 		\item ﻿\textcolor{orange}{2                   }\item ﻿\textcolor{purple}{3                   }\item ﻿\textcolor{purple}{there are 2                 }\item ﻿\textcolor{purple}{i think i see only 2              }\item ﻿\textcolor{purple}{2 in the picture                }\end{enumerate}
 	\noindent\textbf{Top 10 from the remaining NDCG candidates}
 	\begin{enumerate}
 		\item there are 3                 \item at least 4                 \item 4                   \item i can see 4 or 5              \item there are tow                 \item 1                   \item more than 3                 \item i can not tell                \item not sure                  \item i ca n't tell                \end{enumerate}
 	﻿\includegraphics[width=1\linewidth]{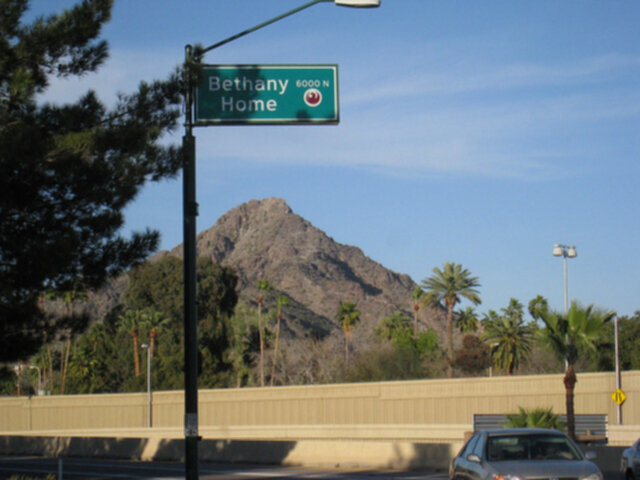}
 	
 	\noindent\textbf{Question:}is it day time ?

 	\noindent\textbf{MRR candidate set}
 	\begin{enumerate}
 		\item ﻿\textcolor{orange}{yes                   }\item ﻿\textcolor{orange}{yes it is                 }\item ﻿\textcolor{orange}{yes , it is                }\item ﻿\textcolor{purple}{it is                  }\item ﻿\textcolor{purple}{yep                   }\end{enumerate}
 	\noindent\textbf{Top 10 from the remaining NDCG candidates}
 	\begin{enumerate}
 		\item yes '                  \item yes , it is day               \item yes , daytime                 \item yes it is daytime                \item it is day time                \item yes it is , it 's sunny             \item yes , i believe so               \item it is daytime                 \item i think so                 \item it 's daytime                 \end{enumerate}
 	﻿\includegraphics[width=1\linewidth]{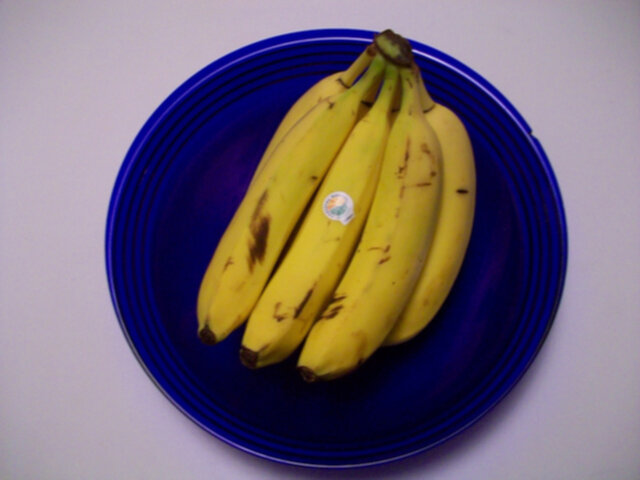}
 	
 	\noindent\textbf{Question:}are there any utensils on the table ?

 	\noindent\textbf{MRR candidate set}
 	\begin{enumerate}
 		\item ﻿\textcolor{orange}{no                   }\item ﻿\textcolor{orange}{there are no utensils                }\item ﻿\textcolor{purple}{nope                   }\item ﻿\textcolor{purple}{no , there are not               }\item ﻿\textcolor{purple}{no , not that i can see             }\item ﻿\textcolor{purple}{no there is not                }\end{enumerate}
 	\noindent\textbf{Top 10 from the remaining NDCG candidates}
 	\begin{enumerate}
 		\item no , there are 0               \item not that i can see               \item 0 that i can see               \item 0 that can be seen               \item 0                   \item 0 are visible                 \item not seen                  \item i do n't think so               \item not really that i can see              \item no other thing                 \end{enumerate}
 	﻿\includegraphics[width=1\linewidth]{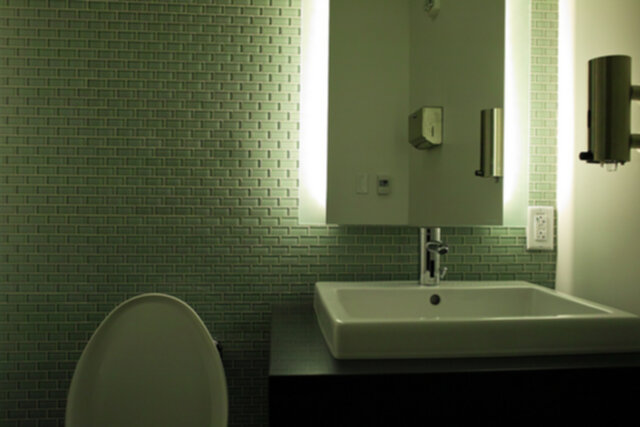}
 	
 	\noindent\textbf{Question:}can you see the floor ?

 	\noindent\textbf{MRR candidate set}
 	\begin{enumerate}
 		\item ﻿\textcolor{orange}{no                   }\item ﻿\textcolor{orange}{yes                   }\item ﻿\textcolor{purple}{nope                   }\item ﻿\textcolor{purple}{i ca n't                 }\item ﻿\textcolor{purple}{no i ca n't                }\end{enumerate}
 	\noindent\textbf{Top 10 from the remaining NDCG candidates}
 	\begin{enumerate}
 		\item no i can not                \item no , the floor is n't visible             \item no , not at all               \item no you can not                \item not really                  \item barely                   \item i do n't think so               \item not that i can see               \item a little bit                 \item yes , i can                \end{enumerate}
 	﻿\includegraphics[width=1\linewidth]{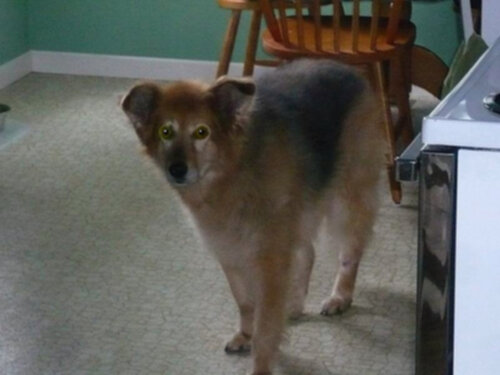}
 	
 	\noindent\textbf{Question:}what color is the dog ?

 	\noindent\textbf{MRR candidate set}
 	\begin{enumerate}
 		\item ﻿\textcolor{red}{brown                   }\item ﻿\textcolor{red}{it is black and tan               }\item ﻿\textcolor{purple}{it is mostly brown with some tan             }\item ﻿\textcolor{purple}{brown and white                 }\item ﻿\textcolor{purple}{light brown                  }\item ﻿\textcolor{purple}{brown with white                 }\end{enumerate}
 	\noindent\textbf{Top 10 from the remaining NDCG candidates}
 	\begin{enumerate}
 		\item tan color                  \item it is reddish-brown                 \item brown and black                 \item brown , black , tan , and white            \item black and brown                 \item i think its a golden retriever , so light yellow          \item browns and blacks                 \item white and brown                 \item no people                  \item UNK                   \end{enumerate}
 	﻿\includegraphics[width=1\linewidth]{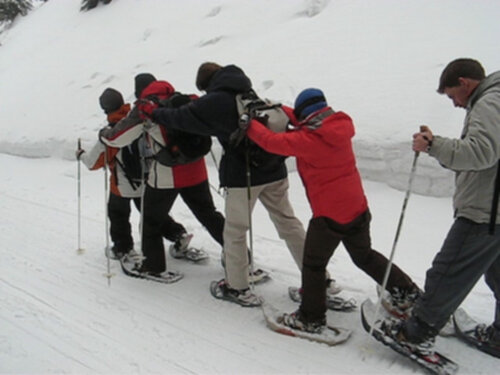}
 	
 	\noindent\textbf{Question:}are they wearing ski goggles ?

 	\noindent\textbf{MRR candidate set}
 	\begin{enumerate}
 		\item ﻿\textcolor{red}{i can not tell                }\item ﻿\textcolor{orange}{no                   }\item ﻿\textcolor{red}{i just see there backs i ca n't see their faces         }\item ﻿\textcolor{purple}{yes                   }\item ﻿\textcolor{purple}{i ca n't tell                }\item ﻿\textcolor{purple}{no , they are n't wearing goggles             }\item ﻿\textcolor{purple}{ca n't tell                 }\end{enumerate}
 	\noindent\textbf{Top 10 from the remaining NDCG candidates}
 	\begin{enumerate}
 		\item ca n't see faces see backs              \item i am not sure                \item no they are not                \item i believe so                 \item yes they are                 \item nope                   \item not that i can see               \item i do n't think so               \item again , i can not see their face            \item they look like it                \end{enumerate}
 	﻿\includegraphics[width=1\linewidth]{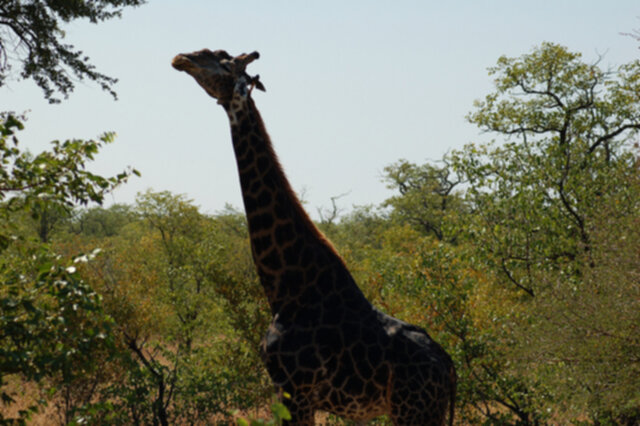}
 	
 	\noindent\textbf{Question:}do the trees have leaves ?

 	\noindent\textbf{MRR candidate set}
 	\begin{enumerate}
 		\item ﻿\textcolor{orange}{yes                   }\item ﻿\textcolor{orange}{some do                  }\item ﻿\textcolor{purple}{yes , they are                }\item ﻿\textcolor{purple}{yes , i think so               }\end{enumerate}
 	\noindent\textbf{Top 10 from the remaining NDCG candidates}
 	\begin{enumerate}
 		\item yes very fresh looking                \item the 1 i can see does , yes            \item long                   \item i think so                 \item at least 2 of them do              \item i 'm sure it can               \item yes , it is                \item yes there 's clearance                \item mostly brown color                 \item maybe                   \end{enumerate}
 	﻿\includegraphics[width=1\linewidth]{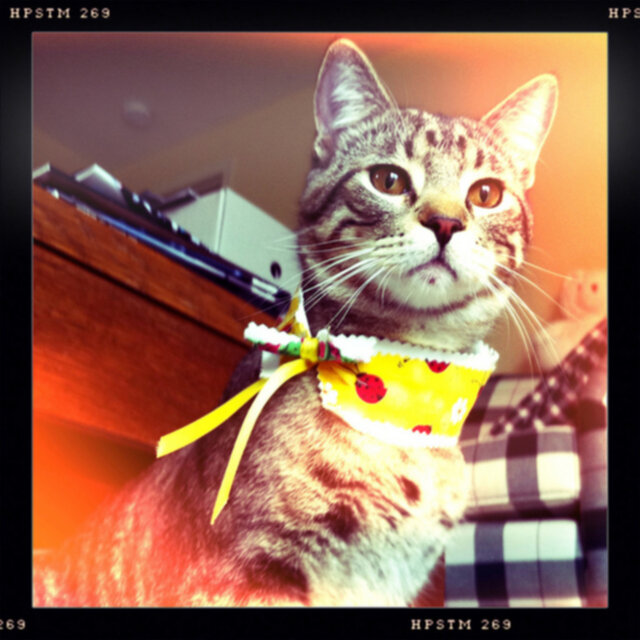}
 	
 	\noindent\textbf{Question:}how big is the table ?

 	\noindent\textbf{MRR candidate set}
 	\begin{enumerate}
 		\item ﻿\textcolor{orange}{i can not tell                }\item ﻿\textcolor{orange}{i ca n't tell                }\item ﻿\textcolor{red}{hard to say , it 's too close up           }\item ﻿\textcolor{purple}{ca n't tell                 }\item ﻿\textcolor{purple}{close up ca n't tell               }\end{enumerate}
 	\noindent\textbf{Top 10 from the remaining NDCG candidates}
 	\begin{enumerate}
 		\item ca n't tell from the picture              \item looks pretty big to me               \item pretty small                  \item not sure                  \item pretty large                  \item quite large                  \item medium size                  \item medium sized                  \item big                   \item it 's more narrow than most              \end{enumerate}
 	﻿\includegraphics[width=1\linewidth]{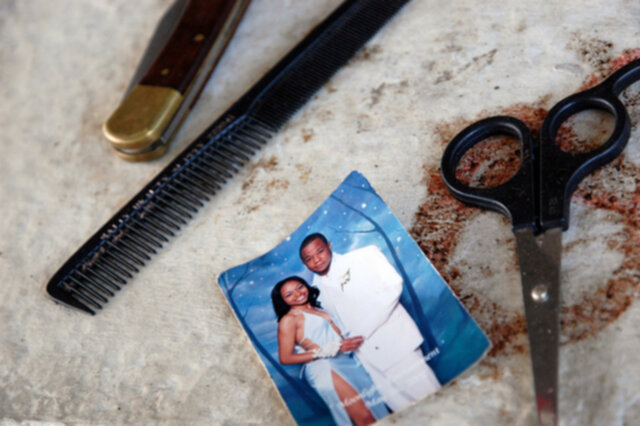}
 	
 	\noindent\textbf{Question:}what kind of scissors ?

 	\noindent\textbf{MRR candidate set}
 	\begin{enumerate}
 		\item ﻿\textcolor{orange}{hair cutting scissors                 }\item ﻿\textcolor{purple}{silver                   }\item ﻿\textcolor{purple}{i can not tell                }\item ﻿\textcolor{purple}{i ca n't tell                }\item ﻿\textcolor{purple}{not sure                  }\end{enumerate}
 	\noindent\textbf{Top 10 from the remaining NDCG candidates}
 	\begin{enumerate}
 		\item black                   \item ca n't tell                 \item larger                   \item wood                   \item grey                   \item gray and white                 \item bear                   \item blue                   \item brown                   \item ca n't tell , too close              \end{enumerate}
 	﻿\includegraphics[width=1\linewidth]{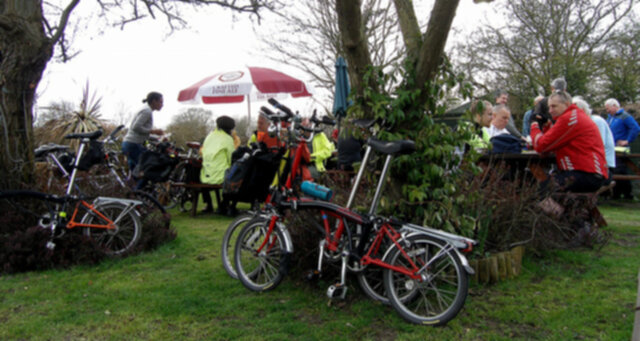}
 	
 	\noindent\textbf{Question:}any signs ?

 	\noindent\textbf{MRR candidate set}
 	\begin{enumerate}
 		\item ﻿\textcolor{orange}{no                   }\item ﻿\textcolor{orange}{yes                   }\item ﻿\textcolor{orange}{0                   }\item ﻿\textcolor{purple}{not that i can see               }\item ﻿\textcolor{purple}{nope                   }\item ﻿\textcolor{purple}{not that i see                }\end{enumerate}
 	\noindent\textbf{Top 10 from the remaining NDCG candidates}
 	\begin{enumerate}
 		\item 0 visible                  \item not visible                  \item nno                   \item yes a couple                 \item i do n't think so               \item yes a few                 \item yes , 1                 \item 1                   \item not really                  \item ca n't see                 \end{enumerate}
 	﻿\includegraphics[width=1\linewidth]{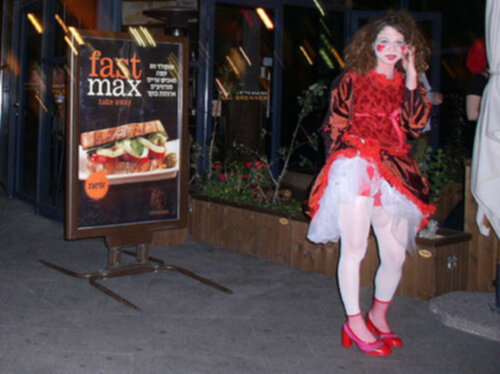}
 	
 	\noindent\textbf{Question:}is it american ?

 	\noindent\textbf{MRR candidate set}
 	\begin{enumerate}
 		\item ﻿\textcolor{orange}{do n't know                 }\item ﻿\textcolor{orange}{no                   }\item ﻿\textcolor{orange}{yes                   }\item ﻿\textcolor{purple}{ca n't tell                 }\item ﻿\textcolor{purple}{i ca n't tell                }\item ﻿\textcolor{purple}{can not tell                 }\item ﻿\textcolor{purple}{i can not tell                }\end{enumerate}
 	\noindent\textbf{Top 10 from the remaining NDCG candidates}
 	\begin{enumerate}
 		\item not sure                  \item i do n't think so               \item nope                   \item i think so , i ca n't tell            \item maybe                   \item no it is n't                \item i believe so , but ca n't say for sure          \item i think so                 \item perhaps                   \item it appears to be but it is hard to confirm          \end{enumerate}
 	﻿\includegraphics[width=1\linewidth]{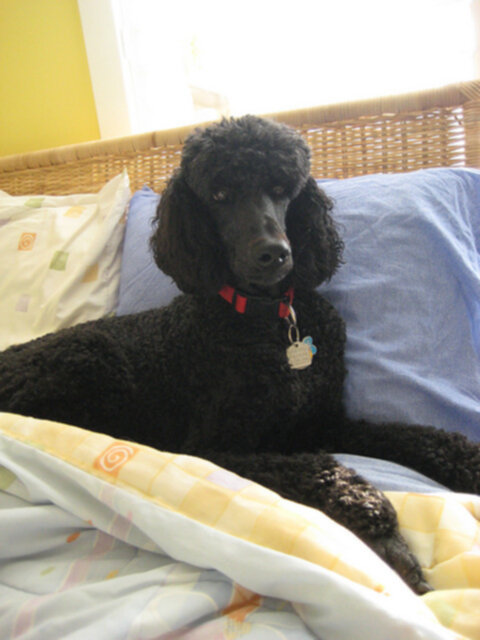}
 	
 	\noindent\textbf{Question:}is the bed outside ?

 	\noindent\textbf{MRR candidate set}
 	\begin{enumerate}
 		\item ﻿\textcolor{orange}{no                   }\item ﻿\textcolor{red}{i can not tell                }\item ﻿\textcolor{purple}{no , it does n't appear to be            }\item ﻿\textcolor{purple}{no it is not                }\item ﻿\textcolor{purple}{no it 's not                }\item ﻿\textcolor{purple}{nope                   }\end{enumerate}
 	\noindent\textbf{Top 10 from the remaining NDCG candidates}
 	\begin{enumerate}
 		\item do n't think so                \item i do n't think so               \item does n't look like it               \item don ’ t see a bed              \item no , it is inside               \item not that i can see               \item inside                   \item i ca n't tell                \item i do n't think so , it 's hard to tell         \item it is not from what i see             \end{enumerate}
 	﻿\includegraphics[width=1\linewidth]{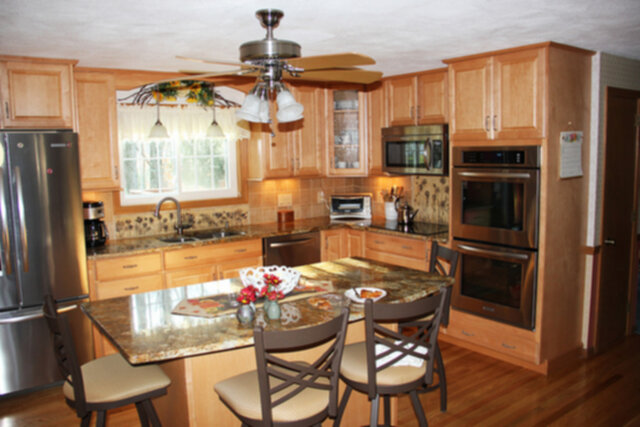}
 	
 	\noindent\textbf{Question:}is there windows ?

 	\noindent\textbf{MRR candidate set}
 	\begin{enumerate}
 		\item ﻿\textcolor{orange}{yes there is                 }\item ﻿\textcolor{orange}{yes                   }\item ﻿\textcolor{purple}{a few                  }\item ﻿\textcolor{purple}{there is a window which has bright sunlight streaming in          }\item ﻿\textcolor{purple}{yes , it looks like               }\end{enumerate}
 	\noindent\textbf{Top 10 from the remaining NDCG candidates}
 	\begin{enumerate}
 		\item there are about 3                \item yes , it is                \item i think so                 \item yes there are tree 's               \item yes it is                 \item only 2 are                 \item yes , there are some storefronts              \item in the distance                 \item 1                   \item nope                   \end{enumerate}
 	﻿\includegraphics[width=1\linewidth]{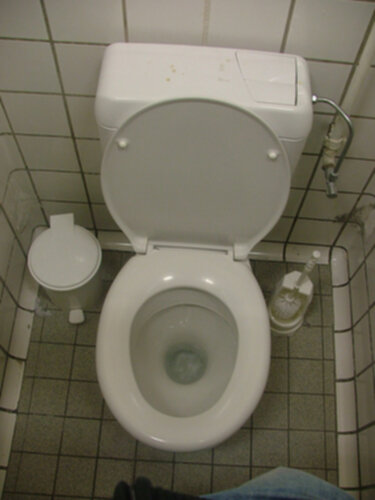}
 	
 	\noindent\textbf{Question:}is there a sink ?

 	\noindent\textbf{MRR candidate set}
 	\begin{enumerate}
 		\item ﻿\textcolor{orange}{no sink                  }\item ﻿\textcolor{purple}{not that i can see               }\item ﻿\textcolor{purple}{not in the image                }\item ﻿\textcolor{purple}{nope                   }\item ﻿\textcolor{purple}{i ca n't see 1               }\item ﻿\textcolor{purple}{there is not                 }\end{enumerate}
 	\noindent\textbf{Top 10 from the remaining NDCG candidates}
 	\begin{enumerate}
 		\item no there is n't                \item no                   \item not that i see                \item not seen in the image               \item yes , there appears to be one , but i can not actually see it     \item not visible                  \item ca n't see it in the picture             \item yes there is a faucet no actual sink            \item ca n't see 1                \item yes there is                 \end{enumerate}
 	﻿\includegraphics[width=1\linewidth]{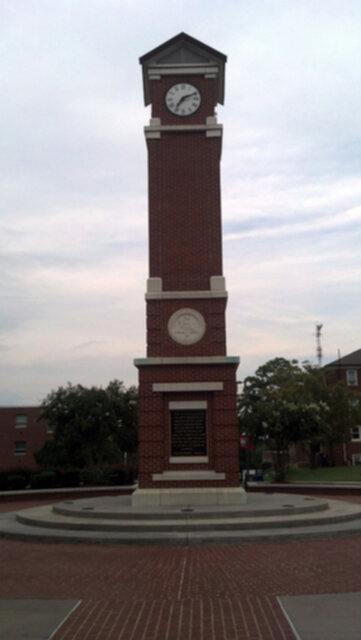}
 	
 	\noindent\textbf{Question:}is the clock tower large ?

 	\noindent\textbf{MRR candidate set}
 	\begin{enumerate}
 		\item ﻿\textcolor{orange}{yes                   }\item ﻿\textcolor{purple}{yes , huge                 }\item ﻿\textcolor{purple}{yes it is                 }\item ﻿\textcolor{purple}{yes , it is                }\end{enumerate}
 	\noindent\textbf{Top 10 from the remaining NDCG candidates}
 	\begin{enumerate}
 		\item yes it 's pretty big               \item not really                  \item i think so                 \item kinda                   \item medium size                  \item no                   \item normal size                  \item medium sized                  \item medium                   \item it 's wide but not tall              \end{enumerate}
 	﻿\includegraphics[width=1\linewidth]{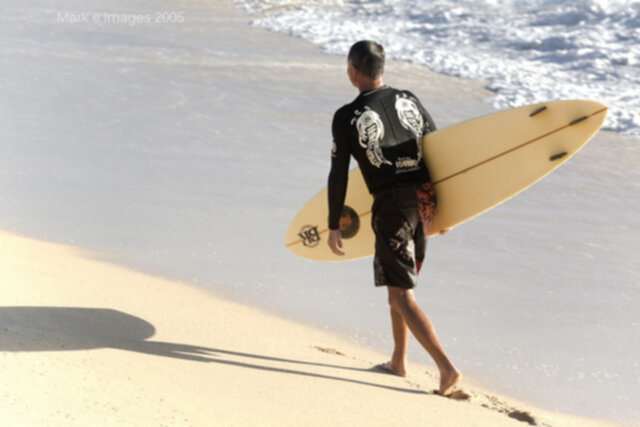}
 	
 	\noindent\textbf{Question:}can you see the water ?

 	\noindent\textbf{MRR candidate set}
 	\begin{enumerate}
 		\item ﻿\textcolor{orange}{yes                   }\item ﻿\textcolor{red}{no                   }\item ﻿\textcolor{purple}{yes i can                 }\item ﻿\textcolor{purple}{yes , i can                }\end{enumerate}
 	\noindent\textbf{Top 10 from the remaining NDCG candidates}
 	\begin{enumerate}
 		\item yews                   \item i think so                 \item yes a little bit                \item yes it is blue                \item a little bit                 \item partially                   \item looks to be                 \item barely                   \item a bit                  \item yes , it is                \end{enumerate}
 	﻿\includegraphics[width=1\linewidth]{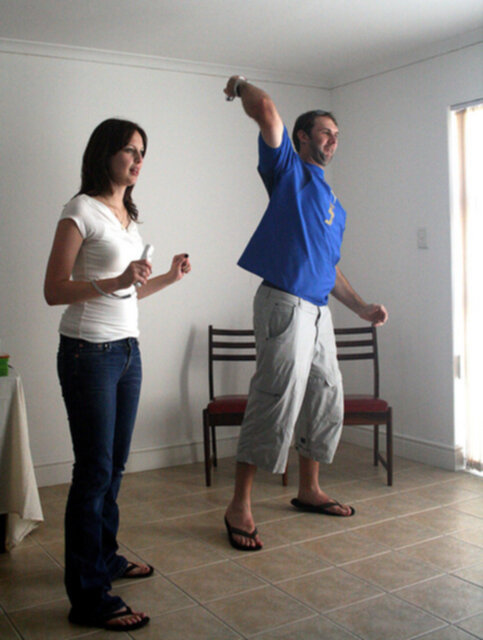}
 	
 	\noindent\textbf{Question:}are the man and woman the only people in the photo ?

 	\noindent\textbf{MRR candidate set}
 	\begin{enumerate}
 		\item ﻿\textcolor{orange}{yes                   }\item ﻿\textcolor{orange}{no                   }\item ﻿\textcolor{purple}{no , they are not               }\item ﻿\textcolor{purple}{yes both are                 }\item ﻿\textcolor{purple}{nope                   }\end{enumerate}
 	\noindent\textbf{Top 10 from the remaining NDCG candidates}
 	\begin{enumerate}
 		\item i think so                 \item no other people that i can see             \item yes that is all                \item i do n't think so               \item just the woman and the little girl             \item yes , it is                \item yes it is                 \item not that i can see               \item yes it 's look like it was take late at night         \item not really                  \end{enumerate}
 	﻿\includegraphics[width=1\linewidth]{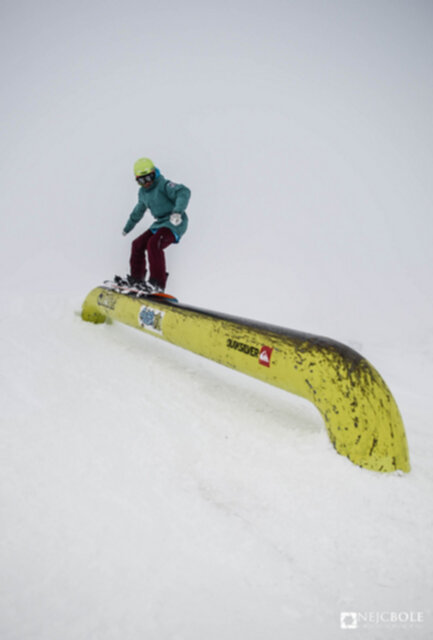}
 	
 	\noindent\textbf{Question:}is anyone watching him ?

 	\noindent\textbf{MRR candidate set}
 	\begin{enumerate}
 		\item ﻿\textcolor{orange}{no                   }\item ﻿\textcolor{purple}{i do not see anyone else in the picture           }\item ﻿\textcolor{purple}{not that i can see               }\item ﻿\textcolor{purple}{i ca n't tell                }\item ﻿\textcolor{purple}{nope                   }\end{enumerate}
 	\noindent\textbf{Top 10 from the remaining NDCG candidates}
 	\begin{enumerate}
 		\item i do n't think so               \item i can not tell                \item ca n't tell                 \item no he 's not                \item hard to see                 \item no he is not                \item not sure                  \item can                   \item possibly                   \item i believe so , but ca n't say for sure          \end{enumerate}
 	﻿\includegraphics[width=1\linewidth]{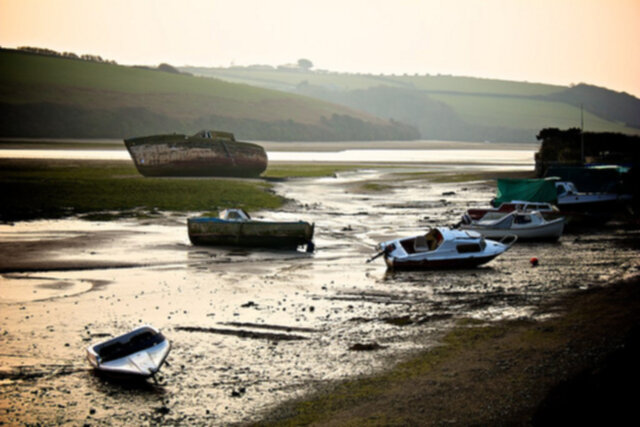}
 	
 	\noindent\textbf{Question:}are they big boats or small boats ?

 	\noindent\textbf{MRR candidate set}
 	\begin{enumerate}
 		\item ﻿\textcolor{orange}{small                   }\item ﻿\textcolor{orange}{they look small                 }\item ﻿\textcolor{purple}{they are all different sizes               }\item ﻿\textcolor{red}{to scale they are large , medium and small UNK          }\item ﻿\textcolor{purple}{medium                   }\end{enumerate}
 	\noindent\textbf{Top 10 from the remaining NDCG candidates}
 	\begin{enumerate}
 		\item mix                   \item both                   \item on the small side                \item no , more like small yachts              \item they look older                 \item no , there are 2 larger ones and 1 smaller 1         \item no the boats are all on the sand            \item no they are small                \item no about medium like the kind that cut hair           \item yes                   \end{enumerate}
 	﻿\includegraphics[width=1\linewidth]{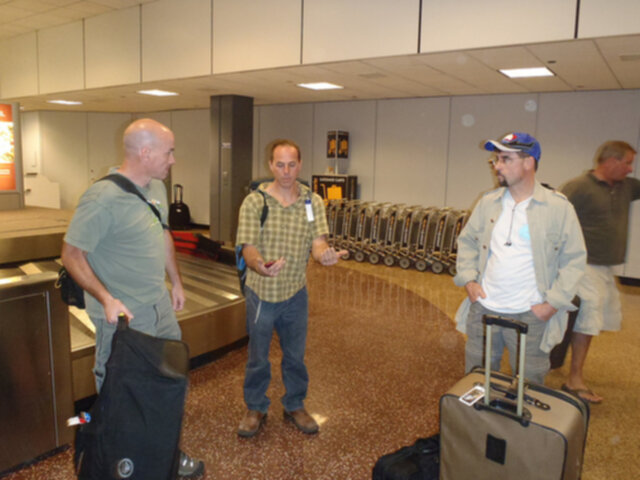}
 	
 	\noindent\textbf{Question:}are there other people there ?

 	\noindent\textbf{MRR candidate set}
 	\begin{enumerate}
 		\item ﻿\textcolor{orange}{no                   }\item ﻿\textcolor{orange}{yes                   }\item ﻿\textcolor{purple}{yes there are                 }\end{enumerate}
 	\noindent\textbf{Top 10 from the remaining NDCG candidates}
 	\begin{enumerate}
 		\item yes a few others                \item yes , there are                \item not that i can see               \item yeah , 3 total                \item not in the photo                \item nope                   \item yes , there are 2 other people in the background          \item a couple in the background , and 1 woman barely seen in the foreground      \item yes a lot                 \item 3 people                  \end{enumerate}
 	﻿\includegraphics[width=1\linewidth]{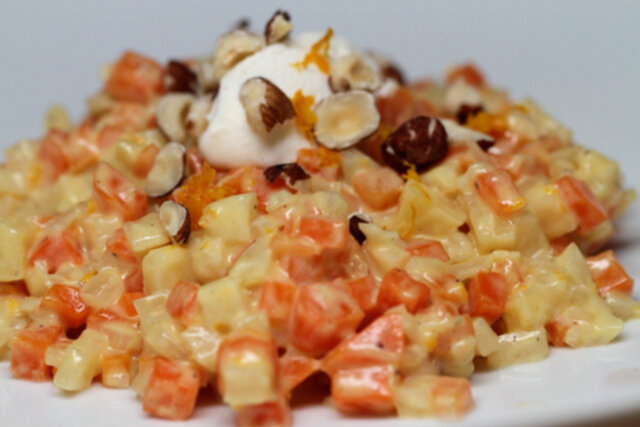}
 	
 	\noindent\textbf{Question:}what is there the most of ?

 	\noindent\textbf{MRR candidate set}
 	\begin{enumerate}
 		\item ﻿\textcolor{orange}{the carrots and what looks like maybe corn are about equal         }\item ﻿\textcolor{purple}{i can not tell                }\item ﻿\textcolor{purple}{i ca n't tell                }\item ﻿\textcolor{purple}{i ca n't see the image              }\item ﻿\textcolor{purple}{not sure                  }\end{enumerate}
 	\noindent\textbf{Top 10 from the remaining NDCG candidates}
 	\begin{enumerate}
 		\item meat                   \item ca n't tell                 \item it looks quite fresh                \item it looks like it might be apple             \item it appears to be people but i do n't have the best view       \item looks brown or black spots on cream             \item light colored                  \item a little bit                 \item hard to tell , tomato sandwich              \item just the ocean                 \end{enumerate}
 	﻿\includegraphics[width=1\linewidth]{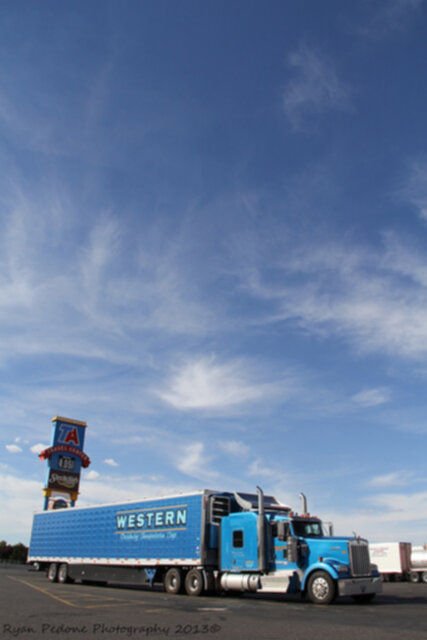}
 	
 	\noindent\textbf{Question:}does the field have a lot of grass ?

 	\noindent\textbf{MRR candidate set}
 	\begin{enumerate}
 		\item ﻿\textcolor{orange}{no                   }\item ﻿\textcolor{orange}{it is in a parking lot              }\item ﻿\textcolor{purple}{no , it does n't               }\item ﻿\textcolor{purple}{nope                   }\item ﻿\textcolor{purple}{not that i can see               }\end{enumerate}
 	\noindent\textbf{Top 10 from the remaining NDCG candidates}
 	\begin{enumerate}
 		\item that part is not in the photo             \item no not in view                \item not really                  \item not especially                  \item 0                   \item ca n't tell                 \item i do n't think so               \item i can not tell                \item i ca n't tell                \item grass                   \end{enumerate}
 	﻿\includegraphics[width=1\linewidth]{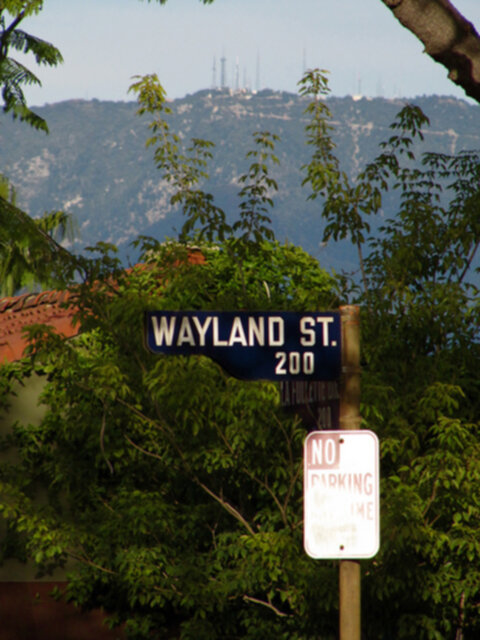}
 	
 	\noindent\textbf{Question:}what color is the sign ?

 	\noindent\textbf{MRR candidate set}
 	\begin{enumerate}
 		\item ﻿\textcolor{orange}{white                   }\item ﻿\textcolor{orange}{there is a blue sign and a white with red writing sign        }\item ﻿\textcolor{purple}{blue                   }\item ﻿\textcolor{purple}{black and white                 }\item ﻿\textcolor{purple}{white and black                 }\end{enumerate}
 	\noindent\textbf{Top 10 from the remaining NDCG candidates}
 	\begin{enumerate}
 		\item white with writing                 \item it is white                 \item white , black and red               \item the sign is black                \item yes , it is                \item red and white                 \item they are white with black around the edges            \item 1 yellow and 1 white               \item yes it is                 \item UNK                   \end{enumerate}
 	﻿\includegraphics[width=1\linewidth]{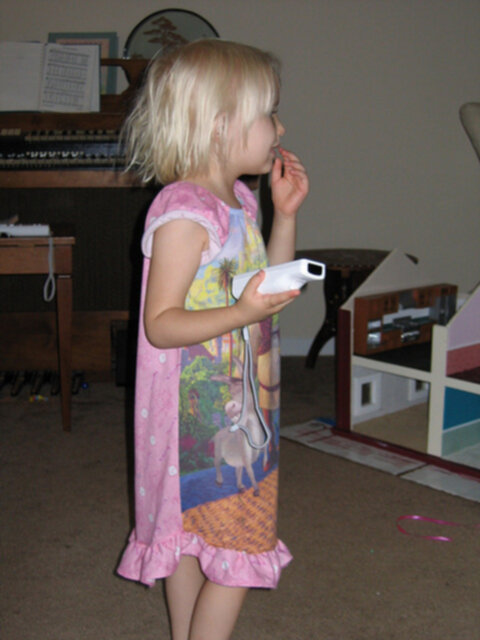}
 	
 	\noindent\textbf{Question:}how old does she look ?

 	\noindent\textbf{MRR candidate set}
 	\begin{enumerate}
 		\item ﻿\textcolor{red}{2                   }\item ﻿\textcolor{purple}{3                   }\item ﻿\textcolor{red}{about 9 , i am not sure             }\item ﻿\textcolor{purple}{maybe 2 at oldest                }\item ﻿\textcolor{purple}{maybe under 1 year old               }\item ﻿\textcolor{purple}{maybe 5                  }\end{enumerate}
 	\noindent\textbf{Top 10 from the remaining NDCG candidates}
 	\begin{enumerate}
 		\item maybe 7                  \item 4                   \item 1                   \item maybe 7-10                  \item looks like a preteen girl               \item 7-9                   \item 5                   \item not sure                  \item 7 or 8                 \item around 8                  \end{enumerate}
 	﻿\includegraphics[width=1\linewidth]{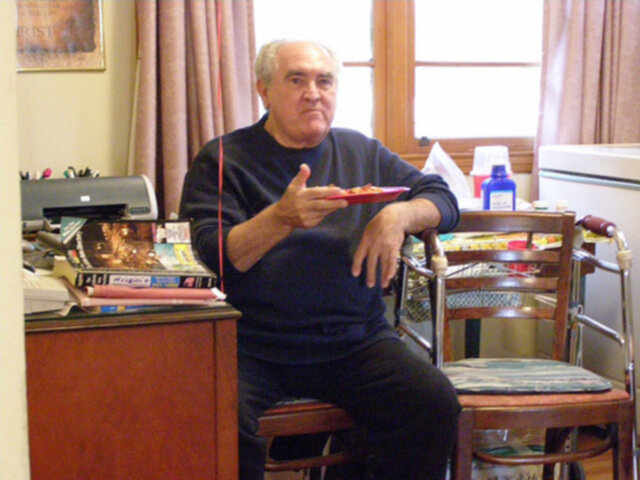}
 	
 	\noindent\textbf{Question:}is there a computer on the desk ?

 	\noindent\textbf{MRR candidate set}
 	\begin{enumerate}
 		\item ﻿\textcolor{orange}{no                   }\item ﻿\textcolor{orange}{yes                   }\item ﻿\textcolor{purple}{i think so                 }\item ﻿\textcolor{purple}{looks like it                 }\end{enumerate}
 	\noindent\textbf{Top 10 from the remaining NDCG candidates}
 	\begin{enumerate}
 		\item yes , it is                \item not that i can see               \item yes but it is not on              \item yes , 1                 \item yeds                   \item nope                   \item not seen                  \item no , there is n't               \item the top of the desk is not visible            \item no there is n't                \end{enumerate}
 	﻿\includegraphics[width=1\linewidth]{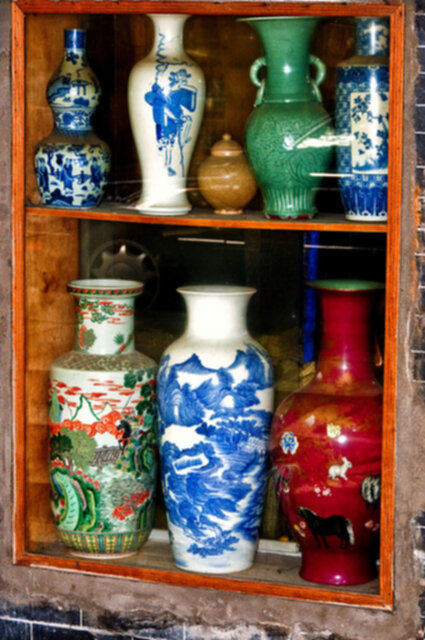}
 	
 	\noindent\textbf{Question:}are there people ?

 	\noindent\textbf{MRR candidate set}
 	\begin{enumerate}
 		\item ﻿\textcolor{orange}{no                   }\item ﻿\textcolor{orange}{no people                  }\item ﻿\textcolor{purple}{nope                   }\item ﻿\textcolor{purple}{no there are no people               }\item ﻿\textcolor{purple}{no there are n't                }\end{enumerate}
 	\noindent\textbf{Top 10 from the remaining NDCG candidates}
 	\begin{enumerate}
 		\item no people are visible                \item not that i can see               \item no people at all                \item i do n't see any               \item 0 that i can see               \item 0                   \item not that can be seen               \item nobody in the photo                \item no , this is a looking up picture            \item i do n't think so               \end{enumerate}
 	﻿\includegraphics[width=1\linewidth]{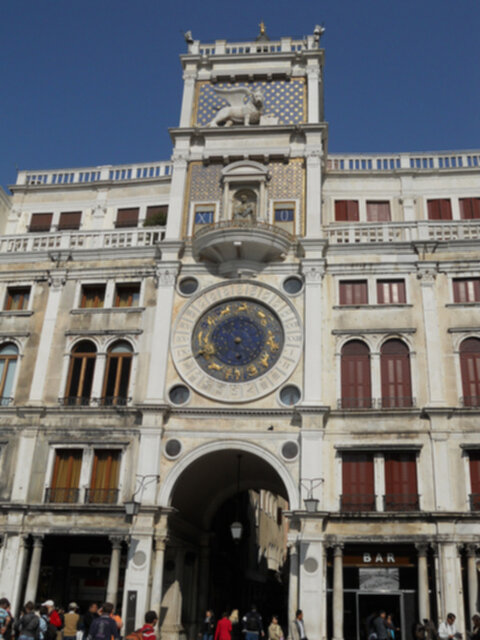}
 	
 	\noindent\textbf{Question:}how many stories is it ?

 	\noindent\textbf{MRR candidate set}
 	\begin{enumerate}
 		\item ﻿\textcolor{purple}{ca n't tell                 }\item ﻿\textcolor{red}{i can not tell                }\item ﻿\textcolor{red}{ca n't tell because the photo is cut but at least 3        }\item ﻿\textcolor{purple}{i ca n't tell                }\item ﻿\textcolor{purple}{maybe 3                  }\item ﻿\textcolor{purple}{not sure                  }\end{enumerate}
 	\noindent\textbf{Top 10 from the remaining NDCG candidates}
 	\begin{enumerate}
 		\item 3                   \item i think 4                 \item at least 4                 \item ca n't see the bottom part              \item 2                   \item at least 5                 \item 4                   \item 5 that i can see               \item i see 4                 \item there are only 2                \end{enumerate}
 	﻿\includegraphics[width=1\linewidth]{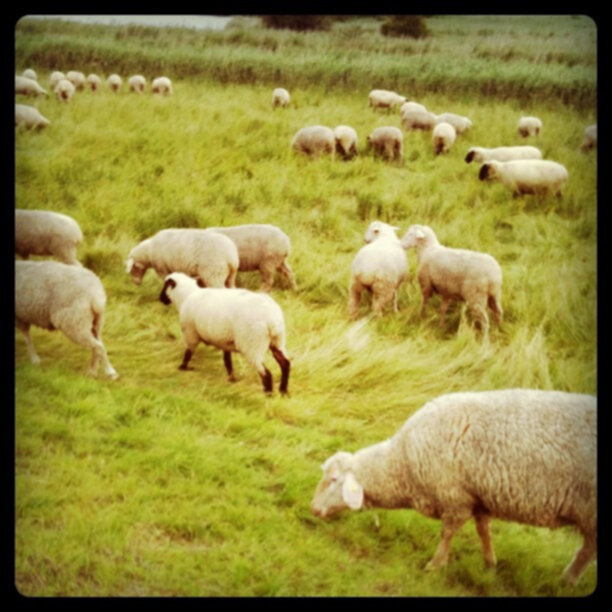}
 	
 	\noindent\textbf{Question:}is the grass green ?

 	\noindent\textbf{MRR candidate set}
 	\begin{enumerate}
 		\item ﻿\textcolor{orange}{yes                   }\item ﻿\textcolor{orange}{yes , some is                }\item ﻿\textcolor{purple}{some of it yes                }\item ﻿\textcolor{purple}{yes it is                 }\item ﻿\textcolor{purple}{yes , it is                }\end{enumerate}
 	\noindent\textbf{Top 10 from the remaining NDCG candidates}
 	\begin{enumerate}
 		\item there is , yes                \item some                   \item yes , i think so               \item some parts of it                \item light greenish yellow                 \item lots of grass but yellow               \item i think so                 \item green                   \item yes , they are                \item yes light brown                 \end{enumerate}
 	﻿\includegraphics[width=1\linewidth]{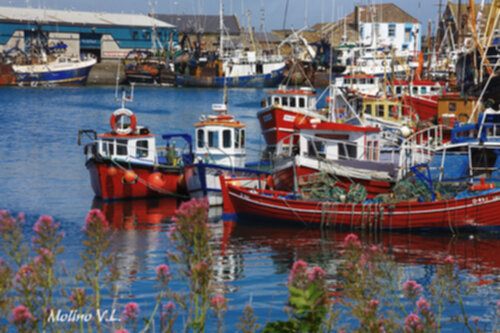}
 	
 	\noindent\textbf{Question:}anything else interesting about the photo ?

 	\noindent\textbf{MRR candidate set}
 	\begin{enumerate}
 		\item ﻿\textcolor{orange}{the colors are very vibrant               }\item ﻿\textcolor{orange}{no                   }\item ﻿\textcolor{orange}{not really                  }\item ﻿\textcolor{purple}{nope                   }\end{enumerate}
 	\noindent\textbf{Top 10 from the remaining NDCG candidates}
 	\begin{enumerate}
 		\item yes                   \item not that i can see               \item no , there are n't               \item i do n't think so               \item looks great                  \item just the sea doo                \item 0                   \item no there are n't                \item not visibly , but i 'd say probably so           \item yes , it is                \end{enumerate}
 	﻿\includegraphics[width=1\linewidth]{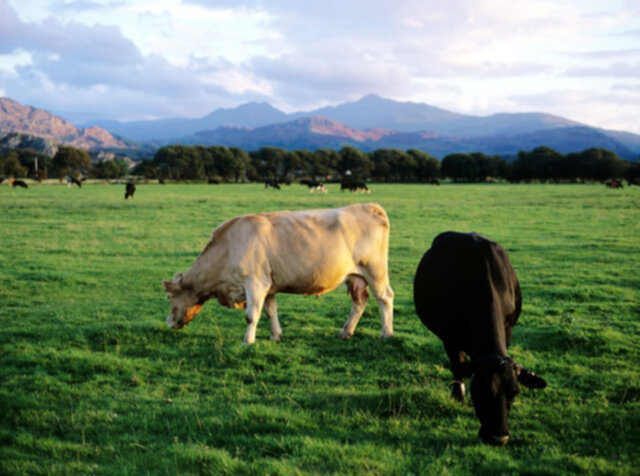}
 	
 	\noindent\textbf{Question:}are there any farmers near them ?

 	\noindent\textbf{MRR candidate set}
 	\begin{enumerate}
 		\item ﻿\textcolor{red}{there are no people visible               }\item ﻿\textcolor{red}{no , there are n't any people in the image          }\item ﻿\textcolor{purple}{no , there is not               }\item ﻿\textcolor{purple}{no there are not                }\item ﻿\textcolor{purple}{no there are n't                }\item ﻿\textcolor{purple}{nope                   }\end{enumerate}
 	\noindent\textbf{Top 10 from the remaining NDCG candidates}
 	\begin{enumerate}
 		\item no                   \item not that i can see               \item there are no people in the image             \item no there isn ’ t               \item no , there are no people              \item i do n't see any               \item no there 's not                \item there are no people                \item no people                  \item 0 i can see                \end{enumerate}
 	﻿\includegraphics[width=1\linewidth]{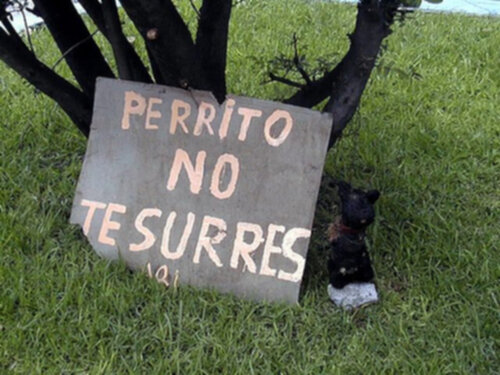}
 	
 	\noindent\textbf{Question:}what does the sign say ?

 	\noindent\textbf{MRR candidate set}
 	\begin{enumerate}
 		\item ﻿\textcolor{red}{welcome to UNK                 }\item ﻿\textcolor{purple}{UNK no UNK                 }\item ﻿\textcolor{red}{it just says UNK dame with an arrow pointing left          }\item ﻿\textcolor{purple}{words                   }\item ﻿\textcolor{purple}{UNK UNK                  }\item ﻿\textcolor{purple}{s gay st                 }\end{enumerate}
 	\noindent\textbf{Top 10 from the remaining NDCG candidates}
 	\begin{enumerate}
 		\item join our csa                 \item beautiful code : leading UNK explain how they think           \item the UNK of heaven                \item the original tour                 \item twin peaks with 200 sign on top             \item number 9                  \item not sure                  \item UNK                   \item no                   \item alaska                   \end{enumerate}
 	﻿\includegraphics[width=1\linewidth]{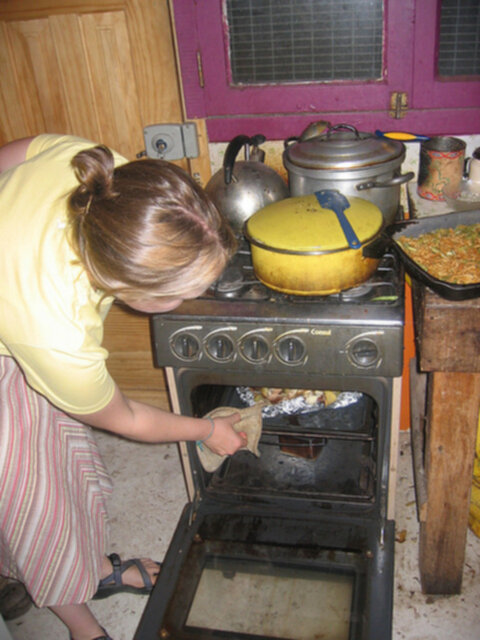}
 	
 	\noindent\textbf{Question:}what is made out of wood ?

 	\noindent\textbf{MRR candidate set}
 	\begin{enumerate}
 		\item ﻿\textcolor{orange}{wood                   }\item ﻿\textcolor{orange}{wood maybe                  }\item ﻿\textcolor{purple}{i can not tell                }\item ﻿\textcolor{red}{cabinets , stove                 }\item ﻿\textcolor{purple}{ca n't tell                 }\item ﻿\textcolor{purple}{hard to tell                 }\item ﻿\textcolor{purple}{i ca n't tell                }\item ﻿\textcolor{purple}{not sure                  }\end{enumerate}
 	\noindent\textbf{Top 10 from the remaining NDCG candidates}
 	\begin{enumerate}
 		\item looks like metal with wood slats              \item looks like a table                \item table                   \item the walls                  \item ca n't really tell but i would assume so           \item yes                   \item yes , it is                \item i think so                 \item i ca n't tell but i am guessing a large bin         \item it looks to be scissors and a beer can box          \end{enumerate}
 	﻿\includegraphics[width=1\linewidth]{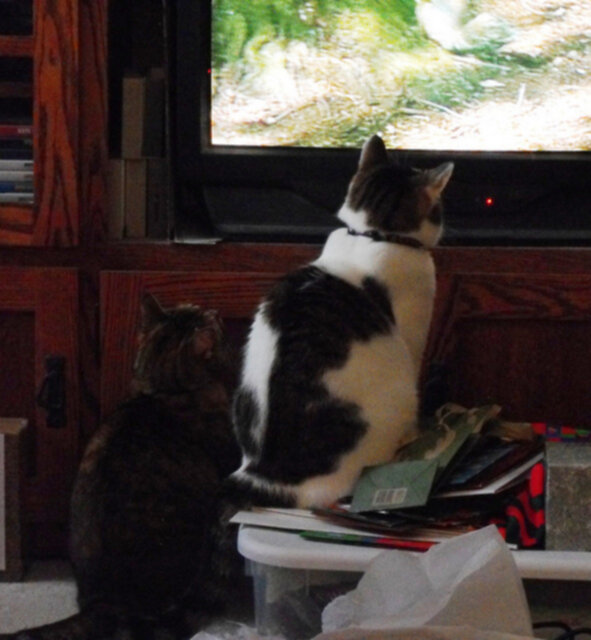}
 	
 	\noindent\textbf{Question:}are there laying on the floor ?

 	\noindent\textbf{MRR candidate set}
 	\begin{enumerate}
 		\item ﻿\textcolor{orange}{no                   }\item ﻿\textcolor{orange}{yes                   }\item ﻿\textcolor{purple}{nope                   }\item ﻿\textcolor{purple}{no there are n't                }\end{enumerate}
 	\noindent\textbf{Top 10 from the remaining NDCG candidates}
 	\begin{enumerate}
 		\item 1 is                  \item no there is not                \item i do n't think so               \item not that i see                \item you can 't see the floor              \item 1 is the other is sitting on top of a container         \item not that i can see               \item ys                   \item i think so                 \item not really                  \end{enumerate}
 	﻿\includegraphics[width=1\linewidth]{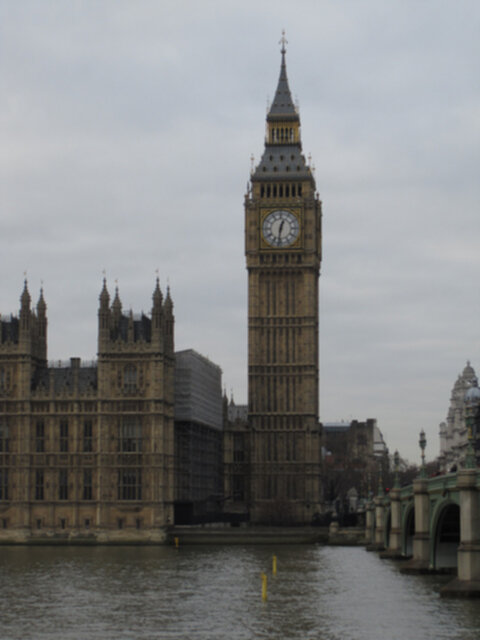}
 	
 	\noindent\textbf{Question:}is there a clock ?

 	\noindent\textbf{MRR candidate set}
 	\begin{enumerate}
 		\item ﻿\textcolor{orange}{yes                   }\item ﻿\textcolor{orange}{no                   }\item ﻿\textcolor{purple}{yes there is                 }\item ﻿\textcolor{purple}{yes , there is                }\item ﻿\textcolor{purple}{yep                   }\end{enumerate}
 	\noindent\textbf{Top 10 from the remaining NDCG candidates}
 	\begin{enumerate}
 		\item UNK                   \item i think so                 \item yes , it is                \item yes it is                 \item yes behind                  \item yes there are 2                \item yes , several of them               \item yes it 's black                \item yes there is a rug               \item yes on a hamper                \end{enumerate}
 	﻿\includegraphics[width=1\linewidth]{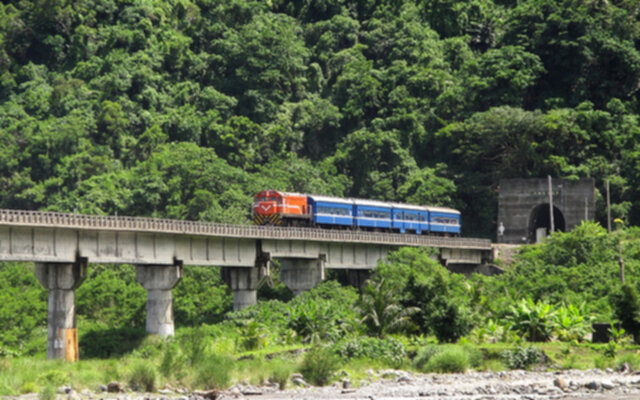}
 	
 	\noindent\textbf{Question:}do the trees have leaves ?

 	\noindent\textbf{MRR candidate set}
 	\begin{enumerate}
 		\item ﻿\textcolor{orange}{yes                   }\item ﻿\textcolor{purple}{yes a lot                 }\item ﻿\textcolor{purple}{yes , very                 }\item ﻿\textcolor{purple}{yes they do                 }\item ﻿\textcolor{purple}{'yes                   }\end{enumerate}
 	\noindent\textbf{Top 10 from the remaining NDCG candidates}
 	\begin{enumerate}
 		\item yes , they are                \item some have leaves                 \item yes a few of them               \item i think so                 \item kinda                   \item yes , they look fine               \item yes , it is                \item yes , a few                \item green                   \item yes it is                 \end{enumerate}
 	﻿\includegraphics[width=1\linewidth]{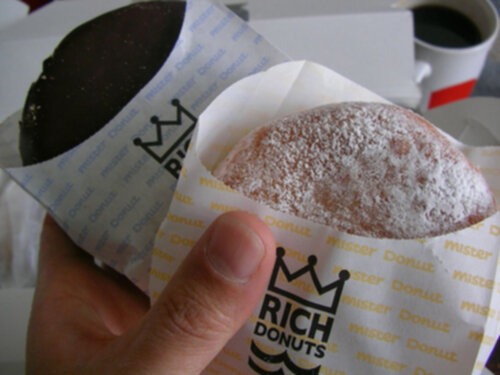}
 	
 	\noindent\textbf{Question:}is the chocolate donut frosted ?

 	\noindent\textbf{MRR candidate set}
 	\begin{enumerate}
 		\item ﻿\textcolor{orange}{yes                   }\item ﻿\textcolor{orange}{no                   }\item ﻿\textcolor{purple}{yes , it is                }\item ﻿\textcolor{purple}{yes it is                 }\end{enumerate}
 	\noindent\textbf{Top 10 from the remaining NDCG candidates}
 	\begin{enumerate}
 		\item it is                  \item nope                   \item looks like it                 \item yes , there 's some white frosting             \item i think it                 \item i think so                 \item yes , they are in both              \item i do n't think so               \item not completely sure , does n't look like it           \item not that i can tell               \end{enumerate}
 	﻿\includegraphics[width=1\linewidth]{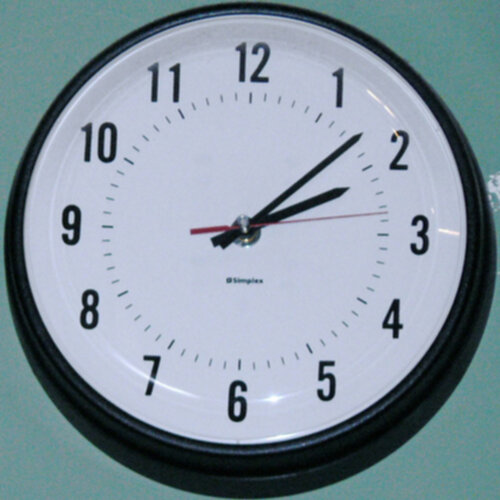}
 	
 	\noindent\textbf{Question:}if you were to guess , what building is the image set in ?

 	\noindent\textbf{MRR candidate set}
 	\begin{enumerate}
 		\item ﻿\textcolor{orange}{i can not tell                }\item ﻿\textcolor{red}{i can not see it from here             }\item ﻿\textcolor{purple}{i ca n't tell                }\item ﻿\textcolor{purple}{ca n't tell                 }\item ﻿\textcolor{purple}{not sure                  }\end{enumerate}
 	\noindent\textbf{Top 10 from the remaining NDCG candidates}
 	\begin{enumerate}
 		\item it is in a home               \item maybe an office or a school              \item can not tell since it is outside             \item it looks like a restaurant               \item this was not taken by a professional             \item in a parking lot somewhere               \item i can not tell by the picture how nice the restaurant is        \item it is hard to tell , i can only see the table cloth       \item no , you can not               \item he is n't in a kitchen              \end{enumerate}
 	﻿\includegraphics[width=1\linewidth]{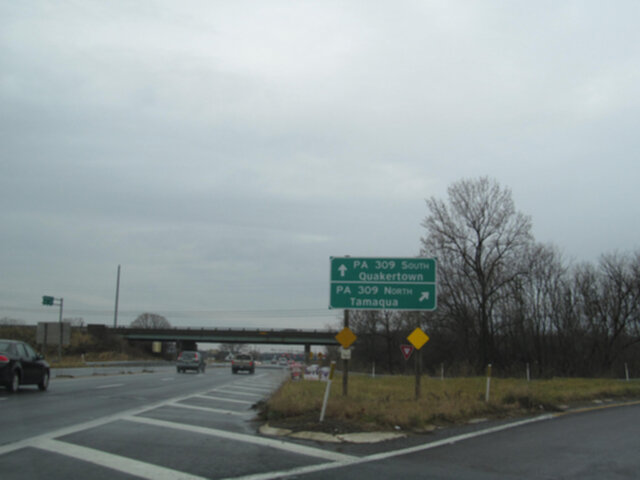}
 	
 	\noindent\textbf{Question:}what is on the sign ?

 	\noindent\textbf{MRR candidate set}
 	\begin{enumerate}
 		\item ﻿\textcolor{orange}{letters                   }\item ﻿\textcolor{orange}{1 says stop and the other says danger            }\item ﻿\textcolor{purple}{pictures                   }\item ﻿\textcolor{purple}{i do n't understand the question              }\item ﻿\textcolor{purple}{a flickr UNK                 }\end{enumerate}
 	\noindent\textbf{Top 10 from the remaining NDCG candidates}
 	\begin{enumerate}
 		\item look like a game                \item land in the background                \item some website not sure                \item UNK                   \item not sure                  \item a track and a little gate building             \item a few penguins out in penguin element             \item yes , i see like 6 billboard signs            \item it 's a dessert                \item large rocks on the side of the road            \end{enumerate}
 	﻿\includegraphics[width=1\linewidth]{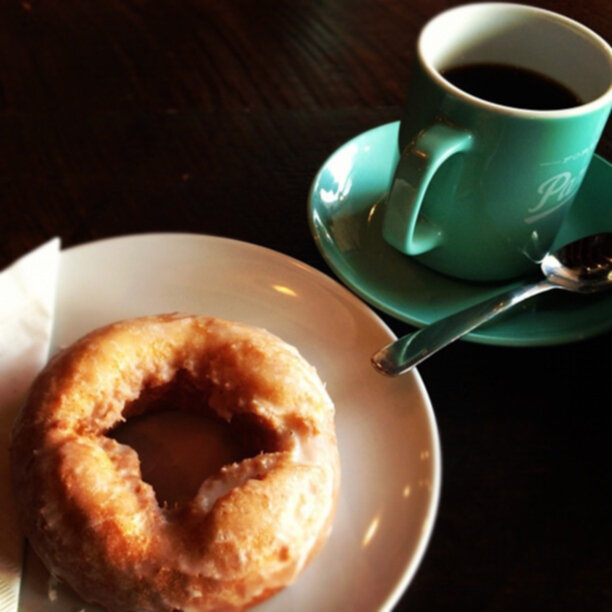}
 	
 	\noindent\textbf{Question:}is there any cream or sugar near the mug ?

 	\noindent\textbf{MRR candidate set}
 	\begin{enumerate}
 		\item ﻿\textcolor{orange}{no , there is a spoon next to the mug on the plate however      }\item ﻿\textcolor{orange}{yes , a small scoop of yellowish ice cream on top of the pastry      }\item ﻿\textcolor{purple}{no there is not                }\item ﻿\textcolor{purple}{no that i can see               }\item ﻿\textcolor{purple}{not that i can see               }\end{enumerate}
 	\noindent\textbf{Top 10 from the remaining NDCG candidates}
 	\begin{enumerate}
 		\item no                   \item nope                   \item not visible in the picture , no             \item i do n't think so               \item yes                   \item yes there is                 \item i can not tell                \item 0                   \item mug                   \item not really                  \end{enumerate}
 	﻿\includegraphics[width=1\linewidth]{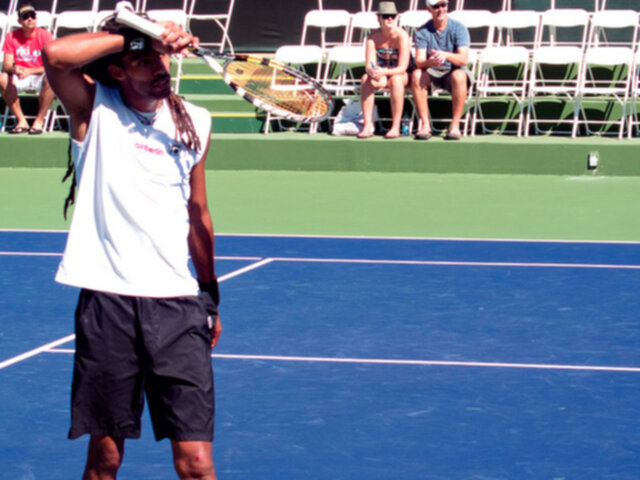}
 	
 	\noindent\textbf{Question:}can you see his opponent ?

 	\noindent\textbf{MRR candidate set}
 	\begin{enumerate}
 		\item ﻿\textcolor{orange}{no                   }\item ﻿\textcolor{orange}{no i can not                }\item ﻿\textcolor{orange}{no , i can not               }\item ﻿\textcolor{purple}{nope                   }\end{enumerate}
 	\noindent\textbf{Top 10 from the remaining NDCG candidates}
 	\begin{enumerate}
 		\item no , not at all               \item not in my view                \item no just 1 person                \item i do n't think so               \item not that i can see               \item i 'm not sure                \item no , only people                \item not really                  \item there is no scoreboard                \item sure why not                 \end{enumerate}
 	﻿\includegraphics[width=1\linewidth]{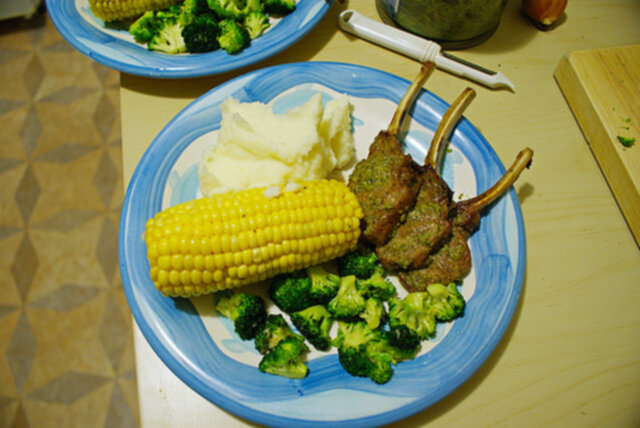}
 	
 	\noindent\textbf{Question:}are any beverages nearby ?

 	\noindent\textbf{MRR candidate set}
 	\begin{enumerate}
 		\item ﻿\textcolor{orange}{no                   }\item ﻿\textcolor{purple}{not that i can see               }\item ﻿\textcolor{purple}{nope                   }\item ﻿\textcolor{purple}{not that i can tell               }\item ﻿\textcolor{purple}{0 that are visible                }\end{enumerate}
 	\noindent\textbf{Top 10 from the remaining NDCG candidates}
 	\begin{enumerate}
 		\item i do n't think so               \item 0                   \item no , thankfully                 \item yes                   \item i do n't know                \item i think so                 \item it is a close up of the plate            \item not really                  \item i can not tell                \item yes 1                  \end{enumerate}
 	﻿\includegraphics[width=1\linewidth]{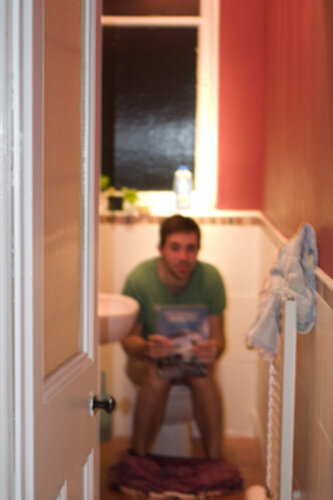}
 	
 	\noindent\textbf{Question:}is this in color ?

 	\noindent\textbf{MRR candidate set}
 	\begin{enumerate}
 		\item ﻿\textcolor{orange}{yes                   }\item ﻿\textcolor{orange}{yes , it is                }\item ﻿\textcolor{orange}{yes it is                 }\item ﻿\textcolor{purple}{yes in color                 }\item ﻿\textcolor{purple}{yes it appears to be               }\end{enumerate}
 	\noindent\textbf{Top 10 from the remaining NDCG candidates}
 	\begin{enumerate}
 		\item it is                  \item yes , i think so               \item yea                   \item i think so                 \item yes* and yes                 \item appears to be                 \item i believe so                 \item most                   \item hell yeah                  \item most likely                  \end{enumerate}
 	﻿\includegraphics[width=1\linewidth]{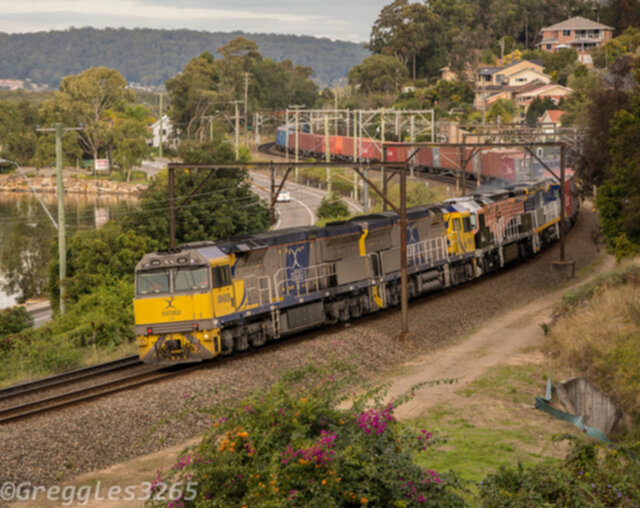}
 	
 	\noindent\textbf{Question:}what is in background ?

 	\noindent\textbf{MRR candidate set}
 	\begin{enumerate}
 		\item ﻿\textcolor{orange}{buildings houses like a village               }\item ﻿\textcolor{orange}{trees and mountains                 }\item ﻿\textcolor{orange}{houses                   }\item ﻿\textcolor{purple}{lots of trees and bushes               }\item ﻿\textcolor{purple}{trees                   }\end{enumerate}
 	\noindent\textbf{Top 10 from the remaining NDCG candidates}
 	\begin{enumerate}
 		\item a house                  \item a UNK and the sky               \item tall grass                  \item yes , there are a few buildings             \item rocks and small bushes                \item the brick building                 \item a tree                  \item yes                   \item not sure                  \item out in the open                \end{enumerate}
 	﻿\includegraphics[width=1\linewidth]{./VisualDialog_test2018/VisualDialog_test2018_000000356633.jpg}
 	
 	\noindent\textbf{Question:}is there anyone on the motorcycle ?

 	\noindent\textbf{MRR candidate set}
 	\begin{enumerate}
 		\item ﻿\textcolor{orange}{no                   }\item ﻿\textcolor{orange}{i can not tell                }\item ﻿\textcolor{orange}{i can not see                }\item ﻿\textcolor{purple}{not that i can see               }\item ﻿\textcolor{purple}{nope                   }\end{enumerate}
 	\noindent\textbf{Top 10 from the remaining NDCG candidates}
 	\begin{enumerate}
 		\item i ca n't tell                \item ca n't tell                 \item i do n't think so               \item no 1                  \item i ca n't see any but probably             \item not sure                  \item it 's too blurry to tell              \item 0                   \item only his feet are visible               \item no people                  \end{enumerate}
 	﻿\includegraphics[width=1\linewidth]{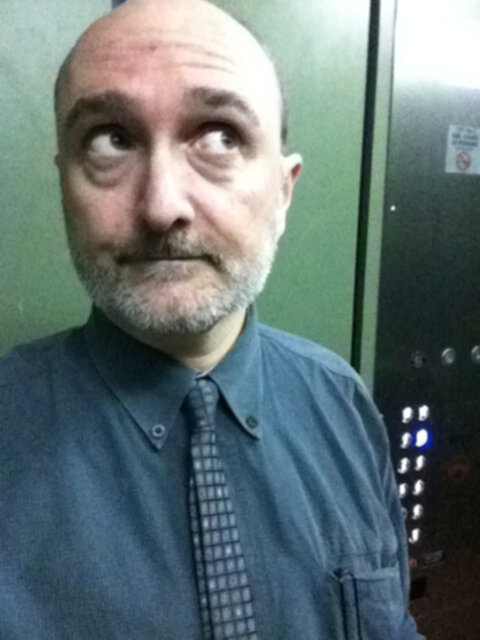}
 	
 	\noindent\textbf{Question:}is the elevator door open ?

 	\noindent\textbf{MRR candidate set}
 	\begin{enumerate}
 		\item ﻿\textcolor{orange}{no                   }\item ﻿\textcolor{orange}{yes                   }\item ﻿\textcolor{red}{i can not tell                }\item ﻿\textcolor{purple}{i ca n't tell                }\item ﻿\textcolor{purple}{ca n't tell                 }\item ﻿\textcolor{purple}{ca n't see                 }\item ﻿\textcolor{purple}{not sure                  }\end{enumerate}
 	\noindent\textbf{Top 10 from the remaining NDCG candidates}
 	\begin{enumerate}
 		\item ca n't really tell                \item no it is n't                \item no , it is not               \item yes it is                 \item not in the picture no               \item yes , it is                \item no not open                 \item no it 's not                \item i think so                 \item it looks like it                \end{enumerate}
 	﻿\includegraphics[width=1\linewidth]{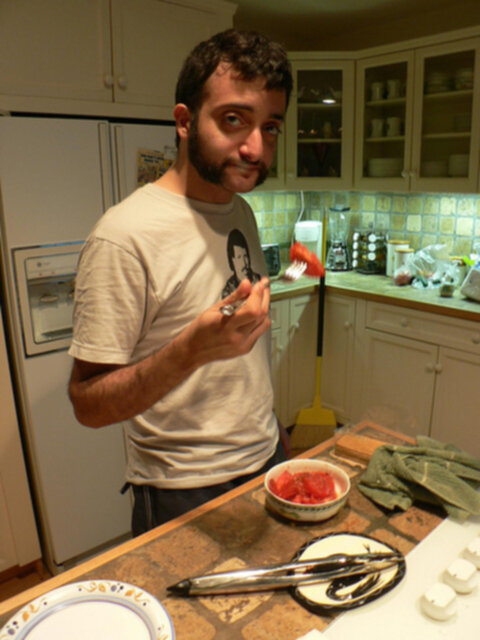}
 	
 	\noindent\textbf{Question:}is he eating over the sink ?

 	\noindent\textbf{MRR candidate set}
 	\begin{enumerate}
 		\item ﻿\textcolor{orange}{no he is eating over the island counter            }\item ﻿\textcolor{purple}{no he is n't                }\item ﻿\textcolor{purple}{no he is not                }\item ﻿\textcolor{purple}{no he isn ’ t               }\item ﻿\textcolor{purple}{no , he is not               }\item ﻿\textcolor{purple}{nope                   }\end{enumerate}
 	\noindent\textbf{Top 10 from the remaining NDCG candidates}
 	\begin{enumerate}
 		\item no                   \item no not eating                 \item i do n't think so               \item yes                   \item yes he is                 \item i can not tell                \item i ca n't tell                \item yes , he is                \item not that i can see               \item i ca n't see that part              \end{enumerate}
 	﻿\includegraphics[width=1\linewidth]{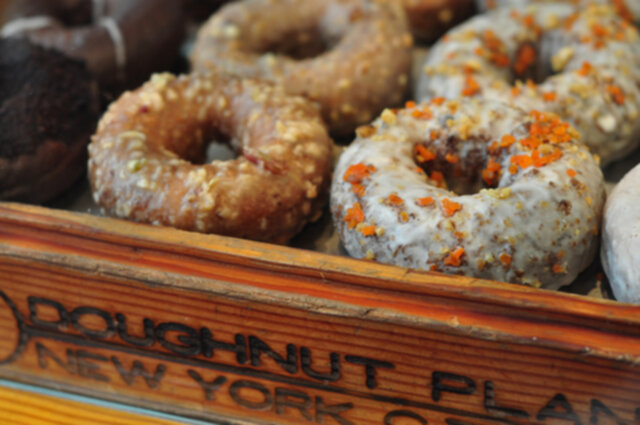}
 	
 	\noindent\textbf{Question:}can you tell if any donuts have been taken from the box ?

 	\noindent\textbf{MRR candidate set}
 	\begin{enumerate}
 		\item ﻿\textcolor{orange}{no                   }\item ﻿\textcolor{red}{i can not tell                }\item ﻿\textcolor{purple}{i do n't think so               }\item ﻿\textcolor{purple}{no i can not                }\item ﻿\textcolor{purple}{nope                   }\item ﻿\textcolor{purple}{not that i can see               }\end{enumerate}
 	\noindent\textbf{Top 10 from the remaining NDCG candidates}
 	\begin{enumerate}
 		\item 0                   \item no i can not tell               \item no , ca n't tell               \item i ca n't tell                \item not really                  \item there are a few but don ’ t know the exact number        \item yes                   \item not sure                  \item ca n't tell                 \item yes 1                  \end{enumerate}
 	﻿\includegraphics[width=1\linewidth]{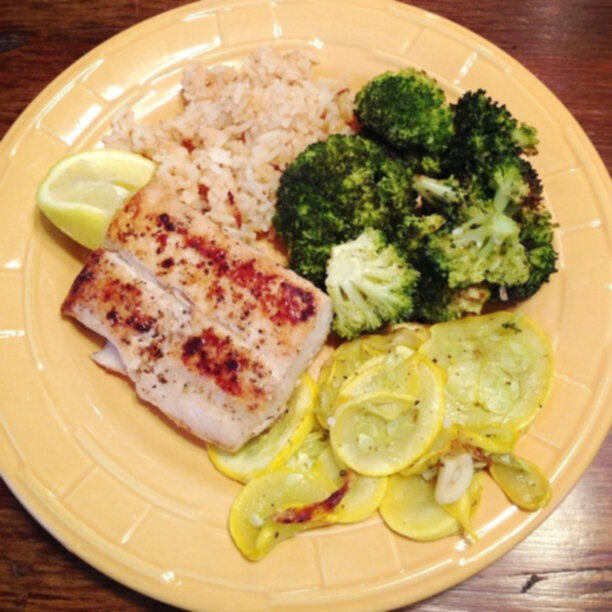}
 	
 	\noindent\textbf{Question:}do you see any hand of a person ?

 	\noindent\textbf{MRR candidate set}
 	\begin{enumerate}
 		\item ﻿\textcolor{orange}{no                   }\item ﻿\textcolor{purple}{no there are no people visible              }\item ﻿\textcolor{purple}{nope                   }\item ﻿\textcolor{purple}{no you can not see any              }\end{enumerate}
 	\noindent\textbf{Top 10 from the remaining NDCG candidates}
 	\begin{enumerate}
 		\item no people are visible                \item no i don ’ t               \item there are no people                \item no people at all                \item no people                  \item not that i can see               \item no , just the food               \item 0 at all                 \item 0                   \item there are 0 of those               \end{enumerate}
 	﻿\includegraphics[width=1\linewidth]{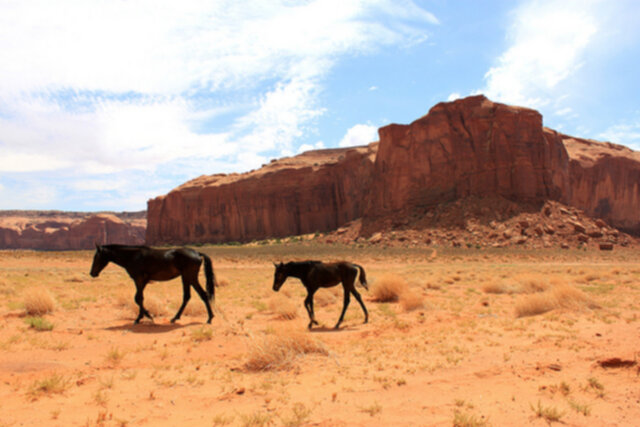}
 	
 	\noindent\textbf{Question:}are there any cacti ?

 	\noindent\textbf{MRR candidate set}
 	\begin{enumerate}
 		\item ﻿\textcolor{orange}{no                   }\item ﻿\textcolor{orange}{0                   }\item ﻿\textcolor{purple}{nope                   }\item ﻿\textcolor{purple}{not that i can see               }\item ﻿\textcolor{purple}{no there are not                }\end{enumerate}
 	\noindent\textbf{Top 10 from the remaining NDCG candidates}
 	\begin{enumerate}
 		\item 0 at all                 \item no i do n't see any              \item i don ’ t see any              \item not any in the picture               \item i do n't think so               \item sadly , no                 \item non                   \item do n't have a clue               \item yes                   \item no , just the grasses               \end{enumerate}
 	﻿\includegraphics[width=1\linewidth]{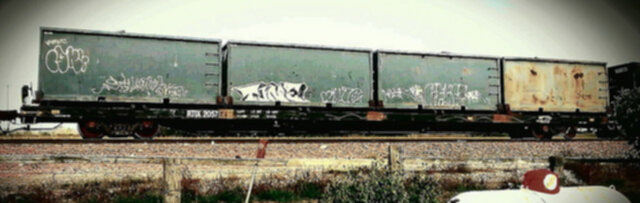}
 	
 	\noindent\textbf{Question:}is there writing on the train ?

 	\noindent\textbf{MRR candidate set}
 	\begin{enumerate}
 		\item ﻿\textcolor{orange}{there is some spray painted graffiti              }\item ﻿\textcolor{purple}{yes                   }\item ﻿\textcolor{purple}{yes , in english and foreign              }\item ﻿\textcolor{purple}{yes , but ca n't make it out            }\item ﻿\textcolor{purple}{yes there is                 }\end{enumerate}
 	\noindent\textbf{Top 10 from the remaining NDCG candidates}
 	\begin{enumerate}
 		\item just numbers                  \item there may be but it is too difficult to see          \item amtrak                   \item yes , it is                \item not that i can see               \item i think so                 \item no                   \item nope                   \item no there is not                \item yes , a little                \end{enumerate}
 	﻿\includegraphics[width=1\linewidth]{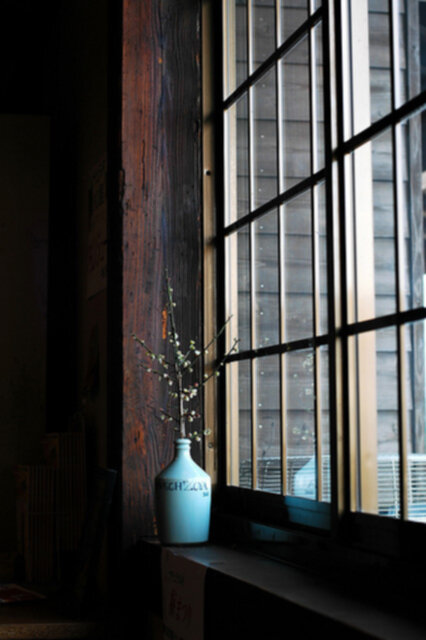}
 	
 	\noindent\textbf{Question:}does it look sunny ?

 	\noindent\textbf{MRR candidate set}
 	\begin{enumerate}
 		\item ﻿\textcolor{orange}{yes                   }\item ﻿\textcolor{red}{yes , it is                }\item ﻿\textcolor{purple}{yes it does                 }\item ﻿\textcolor{purple}{no                   }\item ﻿\textcolor{purple}{no , it is n't               }\item ﻿\textcolor{purple}{not really                  }\end{enumerate}
 	\noindent\textbf{Top 10 from the remaining NDCG candidates}
 	\begin{enumerate}
 		\item i can not tell                \item it looks mostly sunny                \item nope                   \item i ca n't tell                \item i guess so , it is hard to tell           \item yes , i 'd say so              \item not that i can see               \item not sure                  \item ca n't tell                 \item i do n't think so               \end{enumerate}
 	﻿\includegraphics[width=1\linewidth]{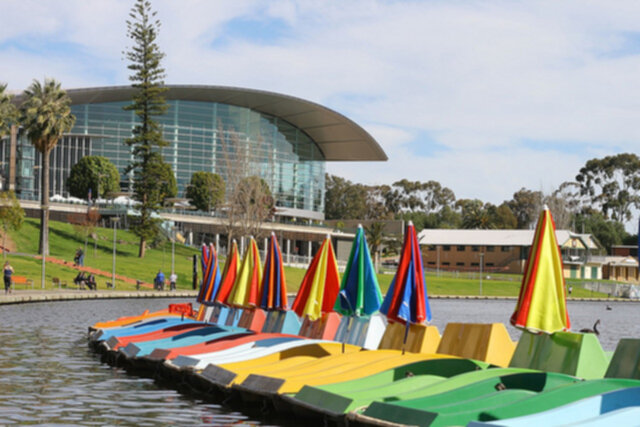}
 	
 	\noindent\textbf{Question:}is the lake quite full ?

 	\noindent\textbf{MRR candidate set}
 	\begin{enumerate}
 		\item ﻿\textcolor{orange}{yes very full                 }\item ﻿\textcolor{orange}{yes                   }\item ﻿\textcolor{orange}{it 's too far away to tell             }\item ﻿\textcolor{purple}{i can not tell                }\item ﻿\textcolor{purple}{i ca n't tell                }\item ﻿\textcolor{purple}{it seems so                 }\end{enumerate}
 	\noindent\textbf{Top 10 from the remaining NDCG candidates}
 	\begin{enumerate}
 		\item i would say yes                \item ca n't tell                 \item yes it is                 \item looks like it                 \item yes , it is                \item i think so                 \item i only see a portion of it , so ca n't tell        \item yes i can see it               \item not sure                  \item yes absolutely                  \end{enumerate}
 	﻿\includegraphics[width=1\linewidth]{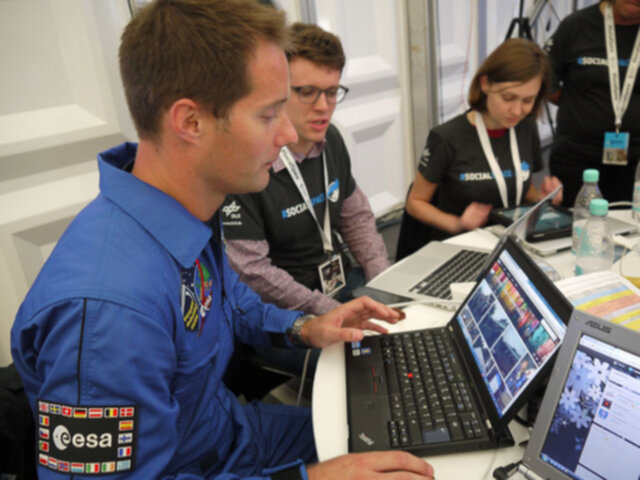}
 	
 	\noindent\textbf{Question:}can you see the floor ?

 	\noindent\textbf{MRR candidate set}
 	\begin{enumerate}
 		\item ﻿\textcolor{orange}{no                   }\item ﻿\textcolor{purple}{no , you can not               }\item ﻿\textcolor{purple}{no , i ca n't               }\item ﻿\textcolor{purple}{nope                   }\item ﻿\textcolor{purple}{no i ca n't                }\end{enumerate}
 	\noindent\textbf{Top 10 from the remaining NDCG candidates}
 	\begin{enumerate}
 		\item no you can not                \item no it is not visible               \item not really                  \item no , i ca n't see the ground            \item yes the floor is visible               \item yes , the floors are visible              \item i do n't think so               \item no this is inside on a table             \item yes                   \item a little bit                 \end{enumerate}
 	﻿\includegraphics[width=1\linewidth]{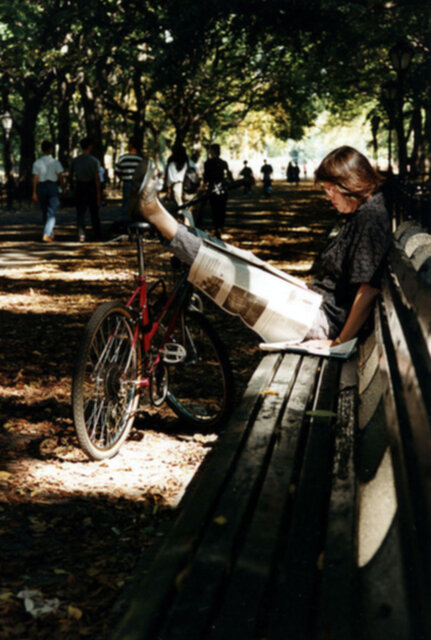}
 	
 	\noindent\textbf{Question:}are there any other people in the picture ?

 	\noindent\textbf{MRR candidate set}
 	\begin{enumerate}
 		\item ﻿\textcolor{orange}{yes there are several people walking in the background           }\item ﻿\textcolor{orange}{yes                   }\item ﻿\textcolor{orange}{there are people in the background              }\item ﻿\textcolor{purple}{there are a few                }\item ﻿\textcolor{purple}{yes lots of them                }\end{enumerate}
 	\noindent\textbf{Top 10 from the remaining NDCG candidates}
 	\begin{enumerate}
 		\item there are people                 \item ye                   \item there are trees in the background , yes            \item yes , a couple ball boys              \item yes , audience                 \item yes , it is                \item i think so                 \item no other people can be seen in the photo           \item there is 1 man in the background             \item there are buildings far off in the distance , it is the only close 1     \end{enumerate}
 	﻿\includegraphics[width=1\linewidth]{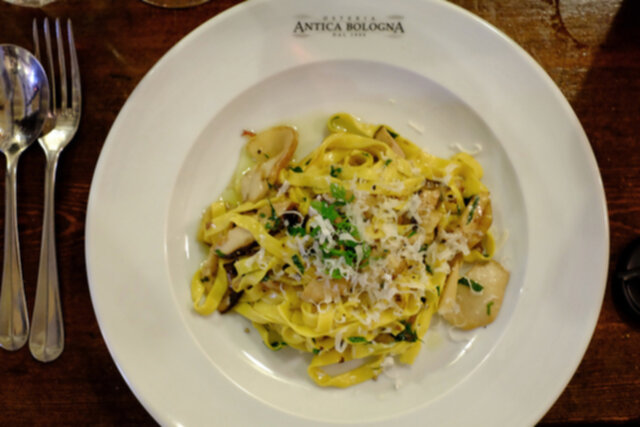}
 	
 	\noindent\textbf{Question:}is the noodles beside the chicken ?

 	\noindent\textbf{MRR candidate set}
 	\begin{enumerate}
 		\item ﻿\textcolor{red}{no it is all mixed together              }\item ﻿\textcolor{orange}{i can not tell                }\item ﻿\textcolor{orange}{yes                   }\item ﻿\textcolor{purple}{no                   }\item ﻿\textcolor{purple}{yes it is                 }\item ﻿\textcolor{purple}{yes , it is                }\item ﻿\textcolor{purple}{yes !                  }\end{enumerate}
 	\noindent\textbf{Top 10 from the remaining NDCG candidates}
 	\begin{enumerate}
 		\item no it is n't                \item across from it                 \item nope                   \item i think so                 \item possibly yes                  \item i do n't believe so               \item not that i can see               \item i do n't think so               \item i think it is , but hard to say for sure         \item not really                  \end{enumerate}
 	﻿\includegraphics[width=1\linewidth]{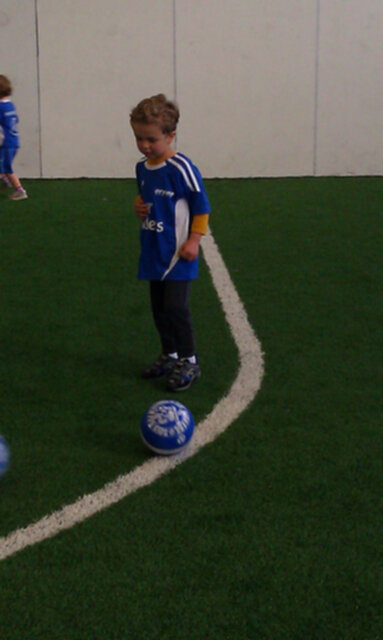}
 	
 	\noindent\textbf{Question:}can you read the writing on the ball ?

 	\noindent\textbf{MRR candidate set}
 	\begin{enumerate}
 		\item ﻿\textcolor{red}{no , i ca n't make it out            }\item ﻿\textcolor{orange}{no                   }\item ﻿\textcolor{orange}{i ca n't , it 's blurry             }\item ﻿\textcolor{purple}{it 's in another language               }\item ﻿\textcolor{purple}{yes i can but it is in a foreign language          }\end{enumerate}
 	\noindent\textbf{Top 10 from the remaining NDCG candidates}
 	\begin{enumerate}
 		\item yes                   \item yes i can                 \item nope                   \item there are a lot of small words it is hard to read        \item yes i do                 \item not really                  \item i can not tell                \item there are several , but i ca n't see the names         \item ca n't see                 \item i ca n't tell                \end{enumerate}
 	﻿\includegraphics[width=1\linewidth]{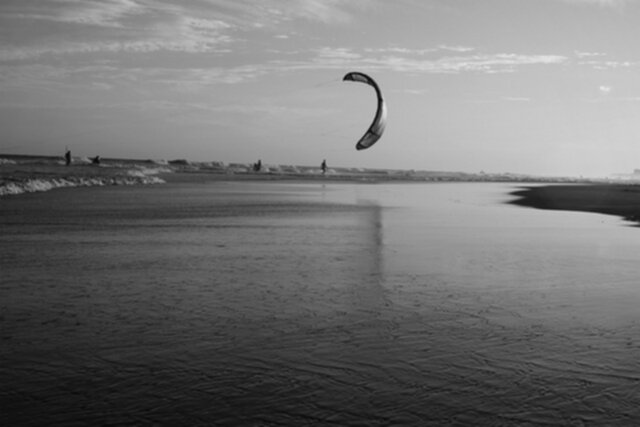}
 	
 	\noindent\textbf{Question:}is the water clear or dirty ?

 	\noindent\textbf{MRR candidate set}
 	\begin{enumerate}
 		\item ﻿\textcolor{red}{i can not tell                }\item ﻿\textcolor{orange}{it appears clean                 }\item ﻿\textcolor{purple}{i ca n't tell                }\item ﻿\textcolor{red}{very clear                  }\item ﻿\textcolor{purple}{can not tell                 }\item ﻿\textcolor{purple}{ca n't tell                 }\end{enumerate}
 	\noindent\textbf{Top 10 from the remaining NDCG candidates}
 	\begin{enumerate}
 		\item it is clean                 \item clean                   \item neither                   \item can barely see the water               \item not sure                  \item it 's a fairly deep bluish-green , does n't look like you could see through it    \item it does not look brown               \item grey                   \item choppy                   \item looks pretty gross                 \end{enumerate}
 	﻿\includegraphics[width=1\linewidth]{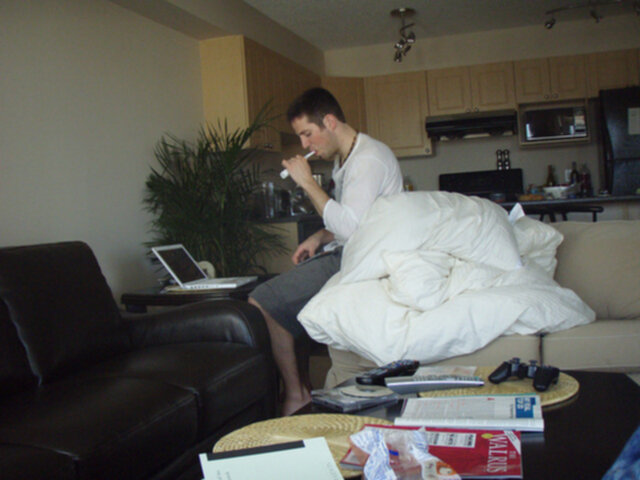}
 	
 	\noindent\textbf{Question:}is it a living room ?

 	\noindent\textbf{MRR candidate set}
 	\begin{enumerate}
 		\item ﻿\textcolor{orange}{yes                   }\item ﻿\textcolor{purple}{i think so                 }\item ﻿\textcolor{purple}{it appears to be                }\item ﻿\textcolor{purple}{yes it is                 }\item ﻿\textcolor{purple}{yes i think so                }\end{enumerate}
 	\noindent\textbf{Top 10 from the remaining NDCG candidates}
 	\begin{enumerate}
 		\item looks like it                 \item yes , it is                \item seems to be                 \item it looks like it could be              \item it could be                 \item maybe                   \item it is                  \item might be , not sure               \item no                   \item not sure                  \end{enumerate}
 	﻿\includegraphics[width=1\linewidth]{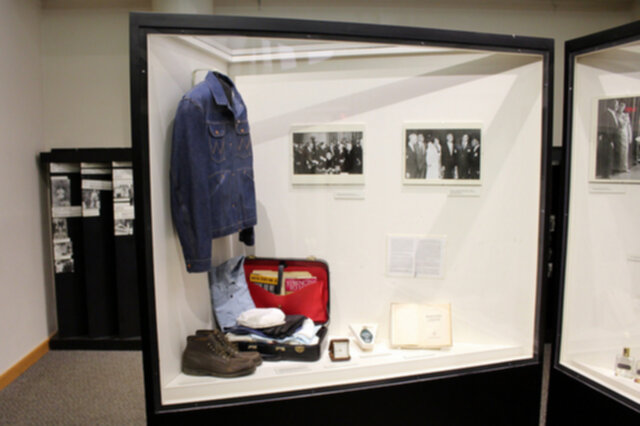}
 	
 	\noindent\textbf{Question:}what color are the boots ?

 	\noindent\textbf{MRR candidate set}
 	\begin{enumerate}
 		\item ﻿\textcolor{orange}{brown                   }\item ﻿\textcolor{orange}{black                   }\item ﻿\textcolor{purple}{i ca n't see from the angle             }\item ﻿\textcolor{purple}{i can not tell                }\item ﻿\textcolor{purple}{i ca n't tell                }\end{enumerate}
 	\noindent\textbf{Top 10 from the remaining NDCG candidates}
 	\begin{enumerate}
 		\item gold                   \item beige and brown                 \item brown and green                 \item ca n't tell                 \item not sure                  \item the look like they are black and white            \item beige                   \item looks like they are red               \item gray                   \item light grey                  \end{enumerate}
 	﻿\includegraphics[width=1\linewidth]{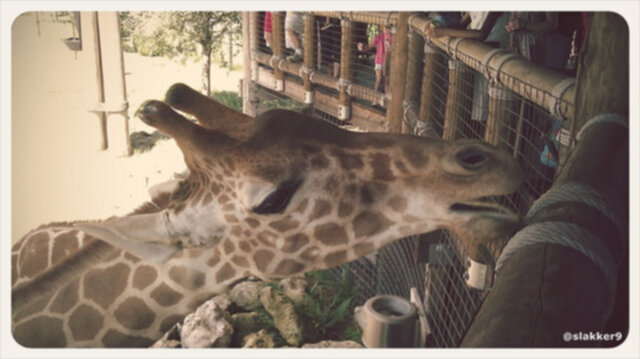}
 	
 	\noindent\textbf{Question:}can you see the clouds ?

 	\noindent\textbf{MRR candidate set}
 	\begin{enumerate}
 		\item ﻿\textcolor{orange}{no                   }\item ﻿\textcolor{purple}{no i can not see the sky             }\item ﻿\textcolor{purple}{no ca n't see sky               }\item ﻿\textcolor{purple}{no , i ca n't see the sky            }\end{enumerate}
 	\noindent\textbf{Top 10 from the remaining NDCG candidates}
 	\begin{enumerate}
 		\item no i ca n't                \item no i can not                \item nope                   \item i ca n't                 \item i can not                 \item i can not tell                \item no the sky is clear               \item i ca n't tell                \item i do n't think so               \item not that i can see               \end{enumerate}
 	﻿\includegraphics[width=1\linewidth]{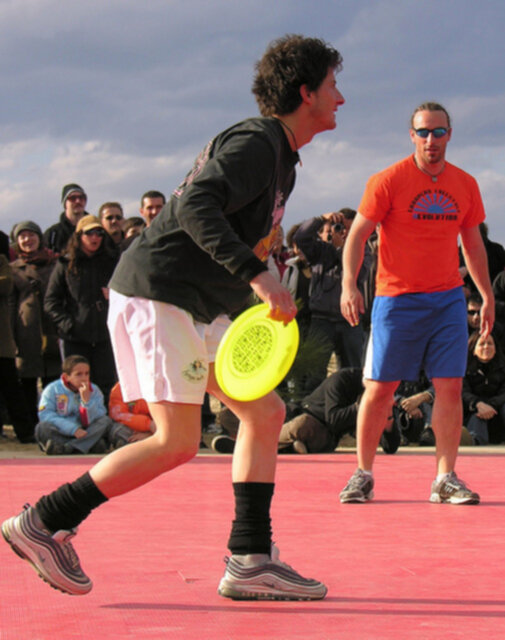}
 	
 	\noindent\textbf{Question:}what color is the frisbee ?

 	\noindent\textbf{MRR candidate set}
 	\begin{enumerate}
 		\item ﻿\textcolor{orange}{yellow                   }\item ﻿\textcolor{purple}{yellow and black                 }\item ﻿\textcolor{purple}{green                   }\item ﻿\textcolor{purple}{yellow and red                 }\item ﻿\textcolor{purple}{green yellow and red                }\end{enumerate}
 	\noindent\textbf{Top 10 from the remaining NDCG candidates}
 	\begin{enumerate}
 		\item yellow as well as his shirt              \item yes , it is                \item olive green                  \item yes                   \item yes it is                 \item red , yellow , white and black             \item a round beige 1                \item just about every color UNK               \item it is white                 \item white                   \end{enumerate}
 	﻿\includegraphics[width=1\linewidth]{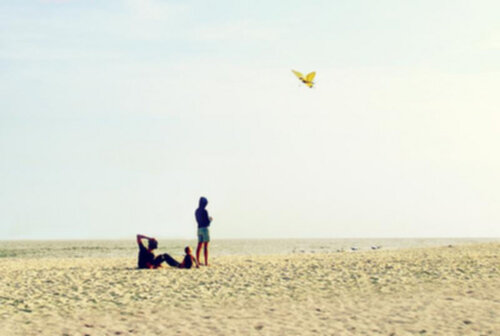}
 	
 	\noindent\textbf{Question:}how old is the kid ?

 	\noindent\textbf{MRR candidate set}
 	\begin{enumerate}
 		\item ﻿\textcolor{red}{not sure maybe 3-4                }\item ﻿\textcolor{red}{around eleven                  }\item ﻿\textcolor{purple}{can not tell                 }\item ﻿\textcolor{purple}{ca n't tell                 }\item ﻿\textcolor{purple}{i can not say                }\item ﻿\textcolor{purple}{i can not tell                }\end{enumerate}
 	\noindent\textbf{Top 10 from the remaining NDCG candidates}
 	\begin{enumerate}
 		\item ca n't tell from the picture              \item i ca n't tell                \item probably around 2                 \item not sure                  \item about 2-3                  \item maybe 1                  \item he looks young maybe 6 or 7             \item i ca n't make out their age             \item maybe 8                  \item the boy looks about 7 or 8             \end{enumerate}
 	﻿\includegraphics[width=1\linewidth]{./VisualDialog_test2018/VisualDialog_test2018_000000235753.jpg}
 	
 	\noindent\textbf{Question:}is he indoors ?

 	\noindent\textbf{MRR candidate set}
 	\begin{enumerate}
 		\item ﻿\textcolor{orange}{yes                   }\item ﻿\textcolor{purple}{i think so                 }\item ﻿\textcolor{purple}{yes , he is                }\item ﻿\textcolor{purple}{yep                   }\end{enumerate}
 	\noindent\textbf{Top 10 from the remaining NDCG candidates}
 	\begin{enumerate}
 		\item yes , i think so               \item he 's inside                 \item seems to be                 \item it is indoors                 \item it looks like it is               \item yes it looks like it was taken indoors            \item yes it is                 \item yes , it is                \item looks like                  \item inside i think                 \end{enumerate}
 	﻿\includegraphics[width=1\linewidth]{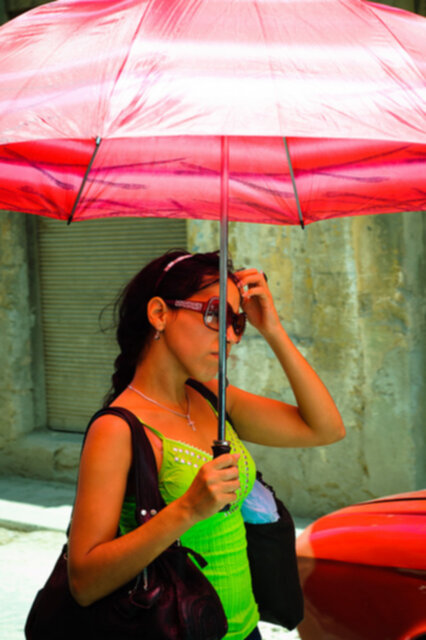}
 	
 	\noindent\textbf{Question:}is the woman old ?

 	\noindent\textbf{MRR candidate set}
 	\begin{enumerate}
 		\item ﻿\textcolor{orange}{no                   }\item ﻿\textcolor{orange}{it is hard to tell , but i would say she is in her twenties     }\item ﻿\textcolor{orange}{probably in her 30s                }\item ﻿\textcolor{purple}{not very young                 }\item ﻿\textcolor{purple}{i do n't know                }\item ﻿\textcolor{purple}{nope                   }\end{enumerate}
 	\noindent\textbf{Top 10 from the remaining NDCG candidates}
 	\begin{enumerate}
 		\item i do n't think so               \item no , she seems like she could be in early college         \item not sure                  \item doubtful                   \item middle aged i 'd say               \item no not really                 \item ca n't tell                 \item not really                  \item not that i can see               \item i think so , ca n't see her face well          \end{enumerate}
 	﻿\includegraphics[width=1\linewidth]{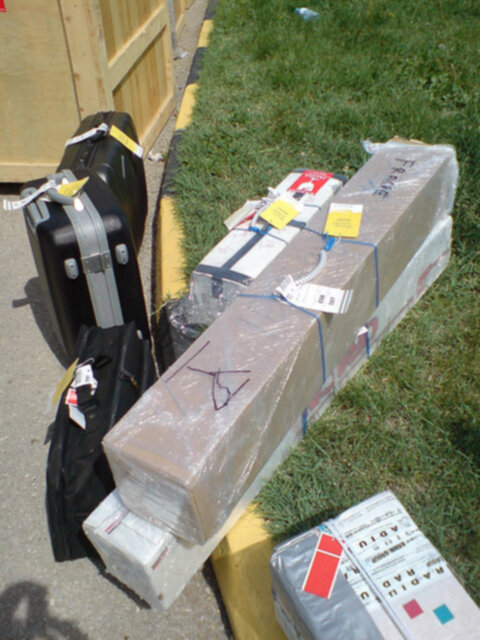}
 	
 	\noindent\textbf{Question:}are there buildings ?

 	\noindent\textbf{MRR candidate set}
 	\begin{enumerate}
 		\item ﻿\textcolor{orange}{no                   }\item ﻿\textcolor{purple}{not that i can see               }\item ﻿\textcolor{purple}{i ca n't see any               }\item ﻿\textcolor{purple}{nope                   }\end{enumerate}
 	\noindent\textbf{Top 10 from the remaining NDCG candidates}
 	\begin{enumerate}
 		\item no there are n't                \item it does n't seem like there are any            \item 0 seen                  \item 0                   \item ca n't see                 \item i do n't think so               \item no , on a sidewalk of a street            \item no it is a close up of them            \item not really                  \item yes                   \end{enumerate}
 	﻿\includegraphics[width=1\linewidth]{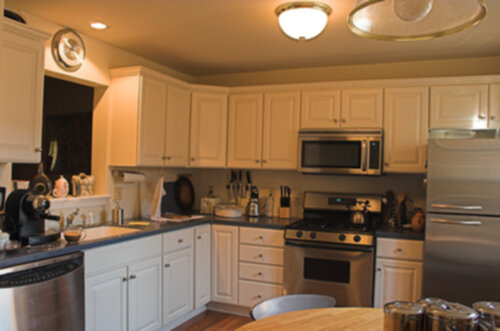}
 	
 	\noindent\textbf{Question:}what color are the countertops ?

 	\noindent\textbf{MRR candidate set}
 	\begin{enumerate}
 		\item ﻿\textcolor{orange}{white                   }\item ﻿\textcolor{purple}{beige                   }\item ﻿\textcolor{purple}{they look beige                 }\item ﻿\textcolor{purple}{they are white                 }\item ﻿\textcolor{purple}{stainless steel                  }\end{enumerate}
 	\noindent\textbf{Top 10 from the remaining NDCG candidates}
 	\begin{enumerate}
 		\item i think off white                \item black                   \item like a beige                 \item dark wood                  \item brown                   \item tan                   \item eggshell                   \item egg white                  \item variety of colors                 \item black i think                 \end{enumerate}
 	﻿\includegraphics[width=1\linewidth]{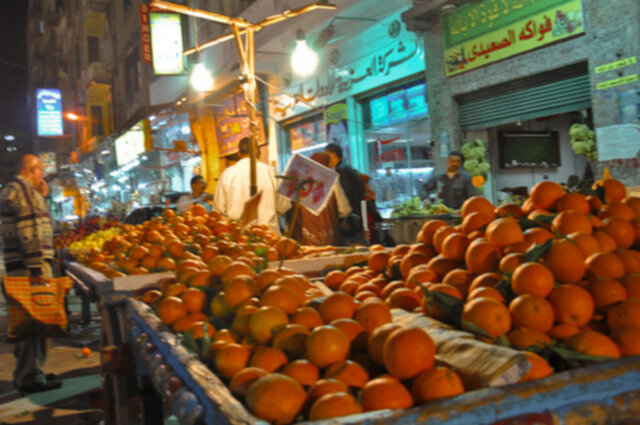}
 	
 	\noindent\textbf{Question:}do you see a scale ?

 	\noindent\textbf{MRR candidate set}
 	\begin{enumerate}
 		\item ﻿\textcolor{orange}{no                   }\item ﻿\textcolor{orange}{yes                   }\item ﻿\textcolor{purple}{nope                   }\end{enumerate}
 	\noindent\textbf{Top 10 from the remaining NDCG candidates}
 	\begin{enumerate}
 		\item no i do n't                \item i do n't                 \item no i do not                \item no i don ’ t               \item no ,                  \item i do n't think so               \item not in the picture                \item not that i can see               \item no it doesn ’ t look like it            \item not really                  \end{enumerate}
 	﻿\includegraphics[width=1\linewidth]{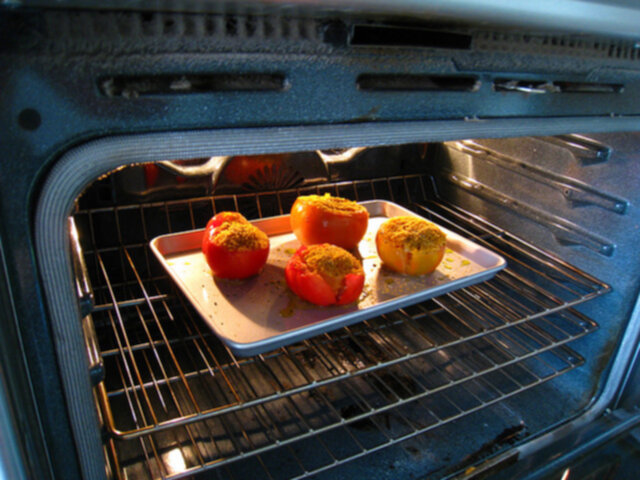}
 	
 	\noindent\textbf{Question:}can you tell what they are stuffed with ?

 	\noindent\textbf{MRR candidate set}
 	\begin{enumerate}
 		\item ﻿\textcolor{red}{no , they are in an self serving orange bottle , it almost looks like a fruit drink  }\item ﻿\textcolor{red}{i can not tell                }\item ﻿\textcolor{purple}{no                   }\item ﻿\textcolor{purple}{not really                  }\item ﻿\textcolor{purple}{nope                   }\end{enumerate}
 	\noindent\textbf{Top 10 from the remaining NDCG candidates}
 	\begin{enumerate}
 		\item i ca n't tell                \item not sure                  \item not that i can see               \item only the top , which looks to be crumbs for some reason beyond me      \item ca n't tell                 \item i do n't see any               \item no the big ones                \item yes                   \item i do n't think so               \item it looks like a seasoned french bread             \end{enumerate}
 	﻿\includegraphics[width=1\linewidth]{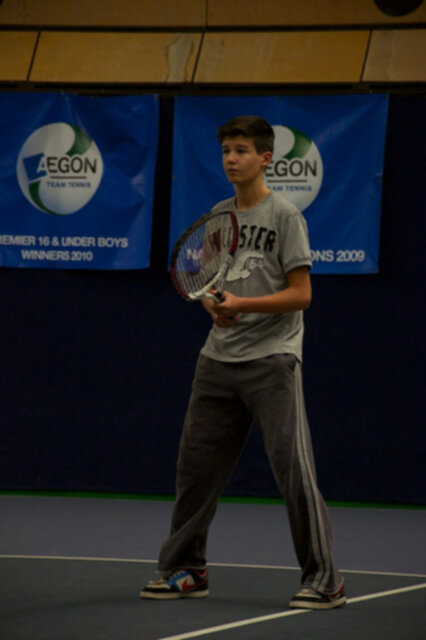}
 	
 	\noindent\textbf{Question:}what color are the sweatpants ?

 	\noindent\textbf{MRR candidate set}
 	\begin{enumerate}
 		\item ﻿\textcolor{red}{grey and white                 }\item ﻿\textcolor{orange}{grey                   }\item ﻿\textcolor{purple}{gray                   }\item ﻿\textcolor{purple}{brown                   }\item ﻿\textcolor{purple}{black                   }\item ﻿\textcolor{red}{they are khaki                 }\end{enumerate}
 	\noindent\textbf{Top 10 from the remaining NDCG candidates}
 	\begin{enumerate}
 		\item gray mostly                  \item not clear                  \item 1 is black and the other is grey            \item 1 is tan and the other is brown            \item they 're gray metal                \item look orange or tan                \item grey , black , white               \item 1 is dark , the other a lighter brown           \item black and gold                 \item not sure                  \end{enumerate}
 	﻿\includegraphics[width=1\linewidth]{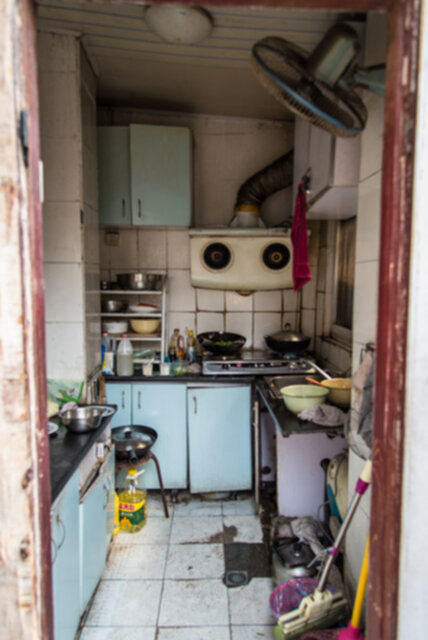}
 	
 	\noindent\textbf{Question:}are the cupboards open or closed ?

 	\noindent\textbf{MRR candidate set}
 	\begin{enumerate}
 		\item ﻿\textcolor{orange}{closed                   }\item ﻿\textcolor{purple}{open                   }\item ﻿\textcolor{purple}{there is a cabinet slightly open              }\item ﻿\textcolor{purple}{halfway open                  }\item ﻿\textcolor{purple}{ca n't tell but looks closed              }\end{enumerate}
 	\noindent\textbf{Top 10 from the remaining NDCG candidates}
 	\begin{enumerate}
 		\item i ca n't tell if there are doors            \item it is open                 \item yes                   \item i can t tell from this angle             \item yes ,                  \item no they are closed                \item no they are all closed               \item no                   \item yes it is open                \item halfway it looks like                \end{enumerate}
 	﻿\includegraphics[width=1\linewidth]{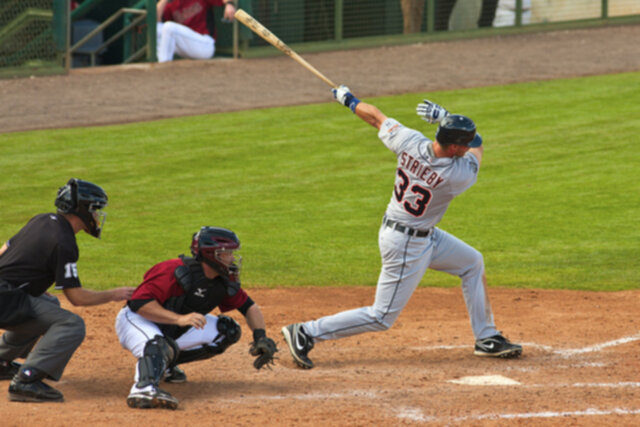}
 	
 	\noindent\textbf{Question:}is there only 1 baseball player ?

 	\noindent\textbf{MRR candidate set}
 	\begin{enumerate}
 		\item ﻿\textcolor{orange}{1 batting and a catcher               }\item ﻿\textcolor{orange}{yes                   }\item ﻿\textcolor{purple}{yes there is                 }\item ﻿\textcolor{purple}{1 that is visible yes               }\end{enumerate}
 	\noindent\textbf{Top 10 from the remaining NDCG candidates}
 	\begin{enumerate}
 		\item only 3 players can be seen from the angle           \item yes , he plays for the mets             \item no i see 2                \item there are 2                 \item it appears to be                \item es                   \item no                   \item yes about 5                 \item yes , it is                \item i think so                 \end{enumerate}
 	﻿\includegraphics[width=1\linewidth]{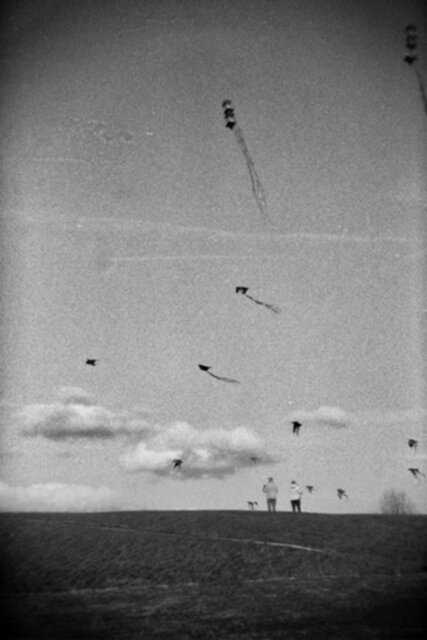}
 	
 	\noindent\textbf{Question:}how old do they seem to be ?

 	\noindent\textbf{MRR candidate set}
 	\begin{enumerate}
 		\item ﻿\textcolor{orange}{they are pretty far away but seem to be an older couple        }\item ﻿\textcolor{purple}{i can not tell                }\item ﻿\textcolor{purple}{i ca n't tell                }\item ﻿\textcolor{purple}{ca n't tell                 }\end{enumerate}
 	\noindent\textbf{Top 10 from the remaining NDCG candidates}
 	\begin{enumerate}
 		\item middle aged                  \item i ca n't tell his back is facing the camera          \item i would say 30s                \item maybe mid thirties                 \item not sure                  \item 20-30                   \item pretty old , in their 60s              \item 20 's or 30 's               \item mid thirties                  \item maybe late 20 's to early 30 's            \end{enumerate}
 	﻿\includegraphics[width=1\linewidth]{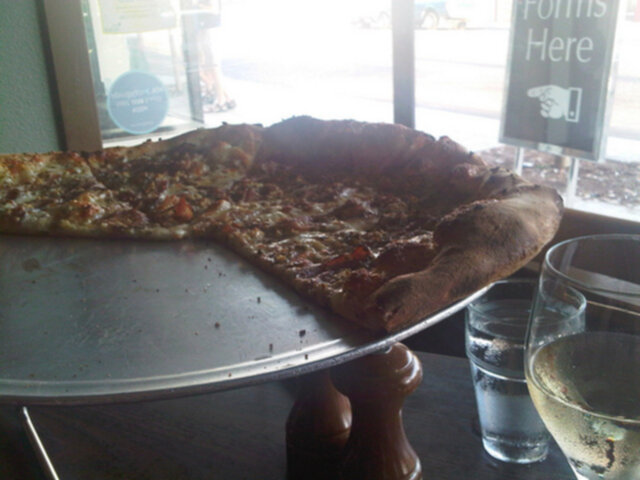}
 	
 	\noindent\textbf{Question:}are there any utensils shown ?

 	\noindent\textbf{MRR candidate set}
 	\begin{enumerate}
 		\item ﻿\textcolor{orange}{no                   }\item ﻿\textcolor{orange}{yes                   }\item ﻿\textcolor{purple}{nope                   }\item ﻿\textcolor{purple}{no there are n't                }\item ﻿\textcolor{purple}{yes , there are                }\end{enumerate}
 	\noindent\textbf{Top 10 from the remaining NDCG candidates}
 	\begin{enumerate}
 		\item not that i can see               \item yes there are some                \item not really                  \item i do n't think so               \item 0                   \item i think so                 \item 0 at all                 \item yes a glass bowl                \item lots                   \item 1                   \end{enumerate}
 	﻿\includegraphics[width=1\linewidth]{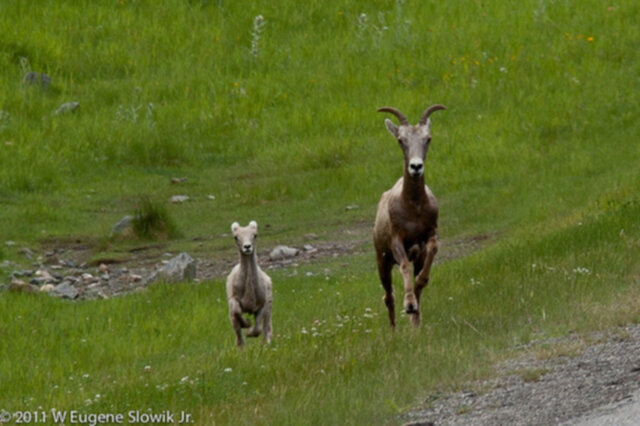}
 	
 	\noindent\textbf{Question:}is the photo in color ?

 	\noindent\textbf{MRR candidate set}
 	\begin{enumerate}
 		\item ﻿\textcolor{orange}{yes                   }\item ﻿\textcolor{orange}{yes it is                 }\item ﻿\textcolor{orange}{yes , it is                }\item ﻿\textcolor{purple}{yes , it 's a color image             }\item ﻿\textcolor{purple}{yes , the image is in color             }\item ﻿\textcolor{purple}{yes it 's in color               }\end{enumerate}
 	\noindent\textbf{Top 10 from the remaining NDCG candidates}
 	\begin{enumerate}
 		\item yes , the picture is in color             \item yes the picture is in color              \item yes in color                 \item it is                  \item it is in color                \item yes ,                  \item yep                   \item definitely                   \item i believe it is , yes              \item yes , it look professional               \end{enumerate}
 	﻿\includegraphics[width=1\linewidth]{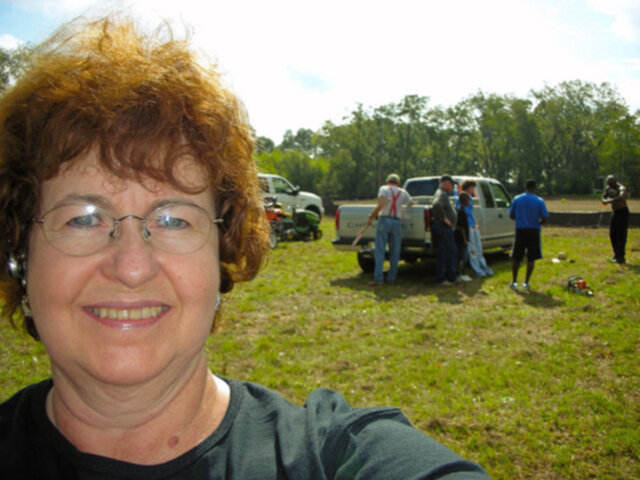}
 	
 	\noindent\textbf{Question:}approximately how many people are there ?

 	\noindent\textbf{MRR candidate set}
 	\begin{enumerate}
 		\item ﻿\textcolor{orange}{6                   }\item ﻿\textcolor{red}{2 in the foreground , about 6 in the back          }\item ﻿\textcolor{purple}{4                   }\item ﻿\textcolor{purple}{5                   }\item ﻿\textcolor{purple}{5 people                  }\item ﻿\textcolor{purple}{i see 4 people                }\end{enumerate}
 	\noindent\textbf{Top 10 from the remaining NDCG candidates}
 	\begin{enumerate}
 		\item i see 7                 \item 3                   \item there are 4 total                \item 8                   \item 3 that i can see               \item about 10                  \item around 9                  \item 4 or 5 , they are very tiny and hard to count        \item about 3                  \item looks like 3                 \end{enumerate}
 	﻿\includegraphics[width=1\linewidth]{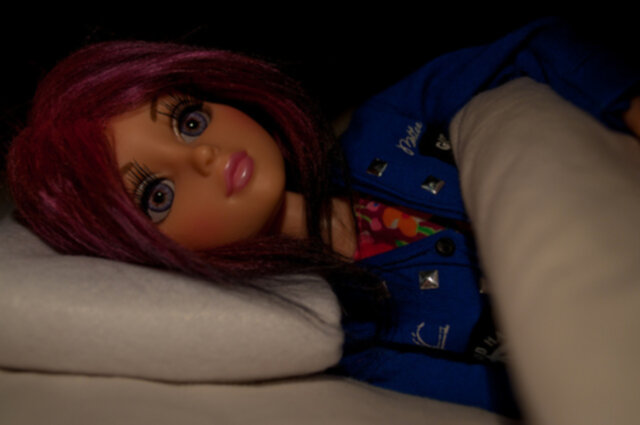}
 	
 	\noindent\textbf{Question:}is there any people ?

 	\noindent\textbf{MRR candidate set}
 	\begin{enumerate}
 		\item ﻿\textcolor{orange}{no                   }\item ﻿\textcolor{orange}{just a photo of 1               }\item ﻿\textcolor{purple}{nope                   }\item ﻿\textcolor{purple}{not that i can see               }\end{enumerate}
 	\noindent\textbf{Top 10 from the remaining NDCG candidates}
 	\begin{enumerate}
 		\item no there is n't                \item no people                  \item not in the image                \item 0 that i can see               \item i do n't see any               \item no 1                  \item no people in the shot               \item no people present                 \item there are no people                \item 0                   \end{enumerate}
 	﻿\includegraphics[width=1\linewidth]{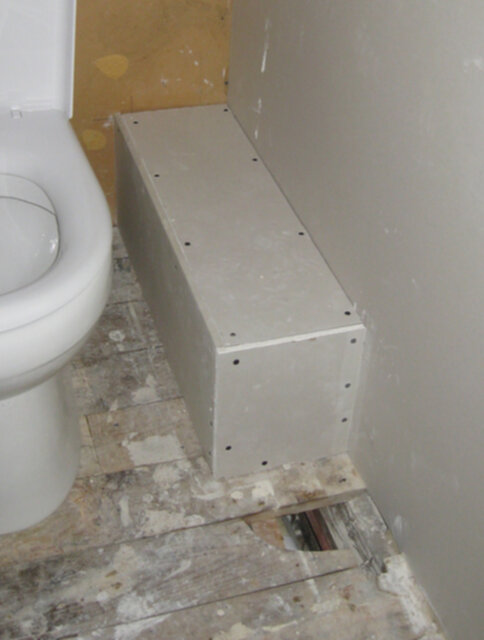}
 	
 	\noindent\textbf{Question:}is it clean ?

 	\noindent\textbf{MRR candidate set}
 	\begin{enumerate}
 		\item ﻿\textcolor{orange}{yes                   }\item ﻿\textcolor{orange}{kind of , it 's under construction             }\item ﻿\textcolor{orange}{no                   }\item ﻿\textcolor{purple}{yes , it is                }\end{enumerate}
 	\noindent\textbf{Top 10 from the remaining NDCG candidates}
 	\begin{enumerate}
 		\item appears to be                 \item yes it is                 \item it looks like it                \item somewhat                   \item hard to tell                 \item kind of                  \item i think so                 \item sort of                  \item not really                  \item nope                   \end{enumerate}
 	﻿\includegraphics[width=1\linewidth]{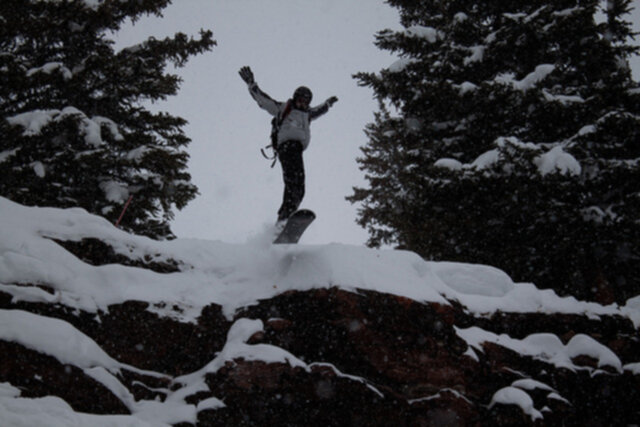}
 	
 	\noindent\textbf{Question:}is it a male ?

 	\noindent\textbf{MRR candidate set}
 	\begin{enumerate}
 		\item ﻿\textcolor{orange}{yes                   }\item ﻿\textcolor{orange}{hard to tell                 }\item ﻿\textcolor{purple}{i can not tell                }\item ﻿\textcolor{red}{is like that                 }\item ﻿\textcolor{purple}{ca n't tell                 }\item ﻿\textcolor{purple}{i ca n't tell                }\end{enumerate}
 	\noindent\textbf{Top 10 from the remaining NDCG candidates}
 	\begin{enumerate}
 		\item not sure                  \item i think so                 \item possibly , i ca n't tell              \item no sure                  \item looks like                  \item it appears to be                \item could be                  \item seems to be                 \item yes , it is                \item yes it is                 \end{enumerate}
 	﻿\includegraphics[width=1\linewidth]{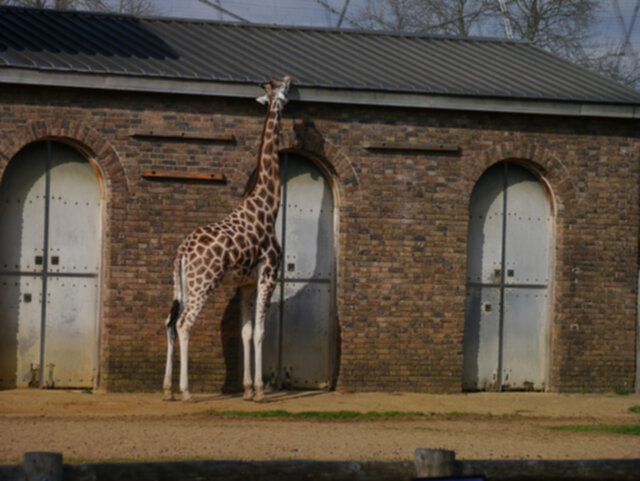}
 	
 	\noindent\textbf{Question:}is the weather UNK ?

 	\noindent\textbf{MRR candidate set}
 	\begin{enumerate}
 		\item ﻿\textcolor{red}{yes , it is                }\item ﻿\textcolor{red}{it looks a bit overcast               }\item ﻿\textcolor{orange}{it looks clear                 }\item ﻿\textcolor{purple}{i can not tell                }\item ﻿\textcolor{purple}{no                   }\item ﻿\textcolor{purple}{i ca n't tell                }\end{enumerate}
 	\noindent\textbf{Top 10 from the remaining NDCG candidates}
 	\begin{enumerate}
 		\item ca n't tell                 \item yes , it is very clear              \item no it is not                \item ca n't tell from photo               \item it is sunny                 \item i do n't think so               \item yes                   \item no it 's not                \item yes , from what i can see             \item yes , the sun is shining              \end{enumerate}
 	﻿\includegraphics[width=1\linewidth]{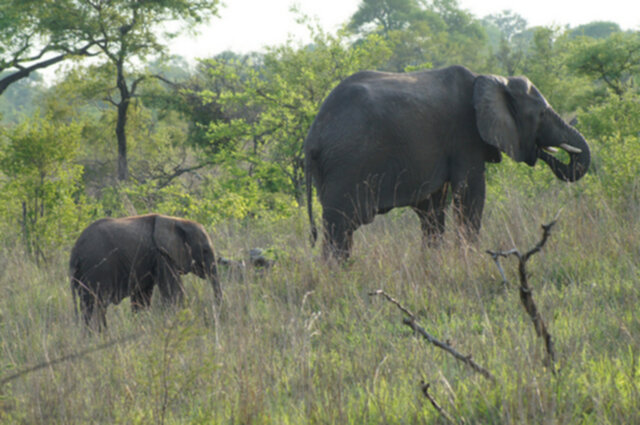}
 	
 	\noindent\textbf{Question:}would you say this is a mother and child ?

 	\noindent\textbf{MRR candidate set}
 	\begin{enumerate}
 		\item ﻿\textcolor{orange}{yes i would                 }\item ﻿\textcolor{orange}{yes                   }\item ﻿\textcolor{purple}{possibly                   }\item ﻿\textcolor{purple}{i think so                 }\item ﻿\textcolor{purple}{yes it is                 }\item ﻿\textcolor{purple}{not sure                  }\end{enumerate}
 	\noindent\textbf{Top 10 from the remaining NDCG candidates}
 	\begin{enumerate}
 		\item maybe                   \item i can not tell                \item yes , it is                \item i ca n't tell                \item ca n't really tell , but i would believe so          \item no                   \item no can not tell                \item i do n't think so               \item ca n't tell                 \item nope                   \end{enumerate}
 	﻿\includegraphics[width=1\linewidth]{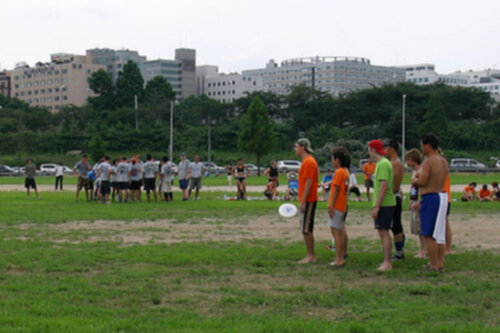}
 	
 	\noindent\textbf{Question:}is it day or night ?

 	\noindent\textbf{MRR candidate set}
 	\begin{enumerate}
 		\item ﻿\textcolor{orange}{day                   }\item ﻿\textcolor{orange}{daytime                   }\item ﻿\textcolor{purple}{it is daytime                 }\item ﻿\textcolor{purple}{day time                  }\item ﻿\textcolor{purple}{it 's daytime                 }\end{enumerate}
 	\noindent\textbf{Top 10 from the remaining NDCG candidates}
 	\begin{enumerate}
 		\item it looks to be day time              \item it is mid day                \item day time i think                \item it looks like it may be late afternoon            \item in the even or morning               \item probably afternoon                  \item yes                   \item yes it is                 \item morning                   \item yes , it is                \end{enumerate}
 	﻿\includegraphics[width=1\linewidth]{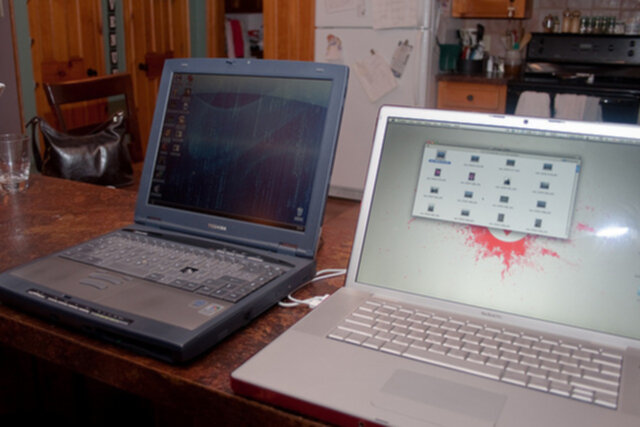}
 	
 	\noindent\textbf{Question:}what brand of laptop are they ?

 	\noindent\textbf{MRR candidate set}
 	\begin{enumerate}
 		\item ﻿\textcolor{orange}{toshiba , i ca n't tell on the other           }\item ﻿\textcolor{orange}{i can not tell                }\item ﻿\textcolor{purple}{i ca n't tell                }\item ﻿\textcolor{purple}{ca n't tell                 }\item ﻿\textcolor{purple}{not sure                  }\end{enumerate}
 	\noindent\textbf{Top 10 from the remaining NDCG candidates}
 	\begin{enumerate}
 		\item no sure                  \item hard to tell                 \item i do n't know                \item can 's really tell                \item you ca n't see it , but if it is truly a laptop you can only see the screen ,\item not necessarily                  \item silver                   \item hard to tell , some kind of pine i think          \item all gray                  \item grey                   \end{enumerate}
 	﻿\includegraphics[width=1\linewidth]{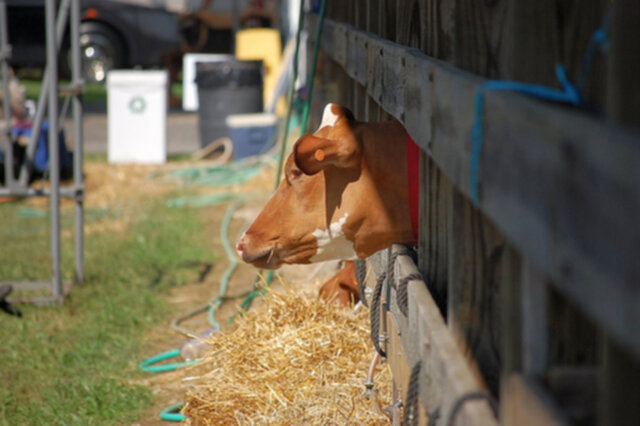}
 	
 	\noindent\textbf{Question:}can you see any other cows ?

 	\noindent\textbf{MRR candidate set}
 	\begin{enumerate}
 		\item ﻿\textcolor{orange}{no                   }\item ﻿\textcolor{purple}{nope                   }\item ﻿\textcolor{purple}{no i can not                }\end{enumerate}
 	\noindent\textbf{Top 10 from the remaining NDCG candidates}
 	\begin{enumerate}
 		\item no , i ca n't               \item just the 1                 \item only 1                  \item i do not                 \item no i do n't                \item not that i can see               \item 0 at all                 \item no3                   \item 0                   \item i do n't think so               \end{enumerate}
 	﻿\includegraphics[width=1\linewidth]{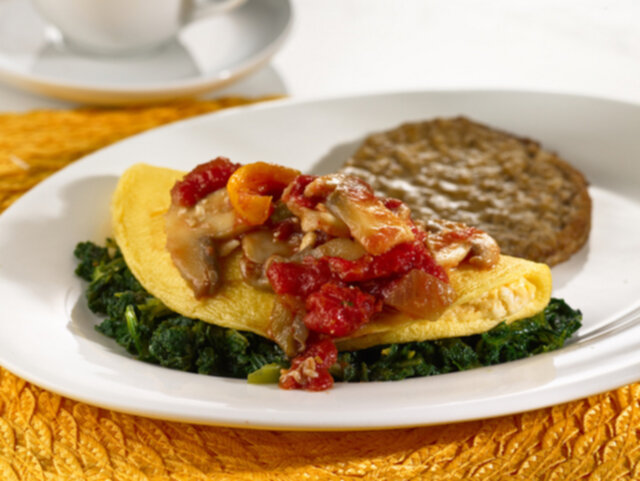}
 	
 	\noindent\textbf{Question:}is there a bun ?

 	\noindent\textbf{MRR candidate set}
 	\begin{enumerate}
 		\item ﻿\textcolor{orange}{yes                   }\item ﻿\textcolor{purple}{yes , there is                }\item ﻿\textcolor{purple}{yes there is                 }\item ﻿\textcolor{purple}{there is                  }\end{enumerate}
 	\noindent\textbf{Top 10 from the remaining NDCG candidates}
 	\begin{enumerate}
 		\item no                   \item yep                   \item nope                   \item yes , i think so               \item no there is n't                \item yes on the side                \item yes , it is                \item i think so                 \item not that i can see               \item i do n't think so               \end{enumerate}

%% file: eacl2021.bbl
\begin{thebibliography}{27}
\expandafter\ifx\csname natexlab\endcsname\relax\def\natexlab#1{#1}\fi

\bibitem[{Antol et~al.(2015)Antol, Agrawal, Lu, Mitchell, Batra, Zitnick, and
  Parikh}]{VQA}
Stanislaw Antol, Aishwarya Agrawal, Jiasen Lu, Margaret Mitchell, Dhruv Batra,
  C.~Lawrence Zitnick, and Devi Parikh. 2015.
\newblock {VQA}: {V}isual {Q}uestion {A}nswering.
\newblock In \emph{ICCV}.

\bibitem[{Das et~al.(2017)Das, Kottur, Gupta, Singh, Yadav, Moura, Parikh, and
  Batra}]{visdial}
Abhishek Das, Satwik Kottur, Khushi Gupta, Avi Singh, Deshraj Yadav,
  Jos{\'e}~MF Moura, Devi Parikh, and Dhruv Batra. 2017.
\newblock Visual dialog.
\newblock In \emph{CVPR}.

\bibitem[{Gan et~al.(2019)Gan, Cheng, Kholy, Li, Liu, and Gao}]{gan2019multi}
Zhe Gan, Yu~Cheng, Ahmed~EI Kholy, Linjie Li, Jingjing Liu, and Jianfeng Gao.
  2019.
\newblock Multi-step reasoning via recurrent dual attention for visual dialog.
\newblock \emph{ACL}.

\bibitem[{Geman et~al.(2015)Geman, Geman, Hallonquist, and
  Younes}]{geman2015visual}
Donald Geman, Stuart Geman, Neil Hallonquist, and Laurent Younes. 2015.
\newblock Visual turing test for computer vision systems.
\newblock \emph{NAS}.

\bibitem[{Guo et~al.(2019)Guo, Xu, and Tao}]{guo2019image}
Dalu Guo, Chang Xu, and Dacheng Tao. 2019.
\newblock Image-question-answer synergistic network for visual dialog.
\newblock In \emph{CVPR}.

\bibitem[{Hu et~al.(2017)Hu, Andreas, Rohrbach, Darrell, and
  Saenko}]{hu2017learning}
Ronghang Hu, Jacob Andreas, Marcus Rohrbach, Trevor Darrell, and Kate Saenko.
  2017.
\newblock Learning to reason: End-to-end module networks for visual question
  answering.
\newblock \emph{ICCV}.

\bibitem[{Hudson and Manning(2019)}]{hudson2019gqa}
Drew~A Hudson and Christopher~D Manning. 2019.
\newblock Gqa: A new dataset for real-world visual reasoning and compositional
  question answering.
\newblock In \emph{CVPR}.

\bibitem[{Jain et~al.(2019)Jain, Lazebnik, and Schwing}]{jain2018two}
Unnat Jain, Svetlana Lazebnik, and Alexander Schwing. 2019.
\newblock Two can play this game: Visual dialog with discriminative question
  generation and answering.
\newblock \emph{CVPR}.

\bibitem[{Jiang et~al.(2020)Jiang, Yu, Qin, Zhuang, Zhang, Hu, and
  Wu}]{jiang2019dualvd}
Xiaoze Jiang, Jing Yu, Zengchang Qin, Yingying Zhuang, Xingxing Zhang, Yue Hu,
  and Qi~Wu. 2020.
\newblock Dualvd: An adaptive dual encoding model for deep visual understanding
  in visual dialogue.
\newblock \emph{AAAI}.

\bibitem[{Kang et~al.(2019)Kang, Lim, and Zhang}]{kang2019dual}
Gi-Cheon Kang, Jaeseo Lim, and Byoung-Tak Zhang. 2019.
\newblock Dual attention networks for visual reference resolution in visual
  dialog.
\newblock \emph{ACL}.

\bibitem[{Kottur et~al.(2018)Kottur, Moura, Parikh, Batra, and
  Rohrbach}]{kottur2018visual}
Satwik Kottur, Jos{\'e}~MF Moura, Devi Parikh, Dhruv Batra, and Marcus
  Rohrbach. 2018.
\newblock Visual coreference resolution in visual dialog using neural module
  networks.
\newblock In \emph{ECCV}.

\bibitem[{Liu et~al.(2016)Liu, Lowe, Serban, Noseworthy, Charlin, and
  Pineau}]{liu2016not}
Chia-Wei Liu, Ryan Lowe, Iulian~V Serban, Michael Noseworthy, Laurent Charlin,
  and Joelle Pineau. 2016.
\newblock How not to evaluate your dialogue system: An empirical study of
  unsupervised evaluation metrics for dialogue response generation.
\newblock \emph{arXiv preprint arXiv:1603.08023}.

\bibitem[{Malinowski and Fritz(2014)}]{malinowski2014multi}
Mateusz Malinowski and Mario Fritz. 2014.
\newblock A multi-world approach to question answering about real-world scenes
  based on uncertain input.
\newblock In \emph{NIPS}.

\bibitem[{Miller(1995)}]{miller1995wordnet}
George~A Miller. 1995.
\newblock Wordnet: a lexical database for english.
\newblock \emph{Communications of the ACM}.

\bibitem[{Murahari et~al.(2020)Murahari, Batra, Parikh, and
  Das}]{murahari2019large}
Vishvak Murahari, Dhruv Batra, Devi Parikh, and Abhishek Das. 2020.
\newblock Large-scale pretraining for visual dialog: A simple state-of-the-art
  baseline.
\newblock \emph{ECCV}.

\bibitem[{Nguyen et~al.(2020)Nguyen, Suganuma, and
  Okatani}]{nguyen2019efficient}
Van-Quang Nguyen, Masanori Suganuma, and Takayuki Okatani. 2020.
\newblock Efficient attention mechanism for handling all the interactions
  between many inputs with application to visual dialog.
\newblock \emph{ECCV}.

\bibitem[{Niu et~al.(2019)Niu, Zhang, Zhang, Zhang, Lu, and
  Wen}]{niu2019recursive}
Yulei Niu, Hanwang Zhang, Manli Zhang, Jianhong Zhang, Zhiwu Lu, and Ji-Rong
  Wen. 2019.
\newblock Recursive visual attention in visual dialog.
\newblock In \emph{CVPR}.

\bibitem[{Qi et~al.(2020)Qi, Niu, Huang, and Zhang}]{qi2019two}
Jiaxin Qi, Yulei Niu, Jianqiang Huang, and Hanwang Zhang. 2020.
\newblock Two causal principles for improving visual dialog.
\newblock \emph{CVPR}.

\bibitem[{Schwartz et~al.(2019{\natexlab{a}})Schwartz, Schwing, and
  Hazan}]{schwartz2019simple}
Idan Schwartz, Alexander~G Schwing, and Tamir Hazan. 2019{\natexlab{a}}.
\newblock A simple baseline for audio-visual scene-aware dialog.
\newblock In \emph{CVPR}.

\bibitem[{Schwartz et~al.(2019{\natexlab{b}})Schwartz, Yu, Hazan, and
  Schwing}]{schwartz2019factor}
Idan Schwartz, Seunghak Yu, Tamir Hazan, and Alexander~G Schwing.
  2019{\natexlab{b}}.
\newblock Factor graph attention.
\newblock In \emph{CVPR}.

\bibitem[{Seo et~al.(2017)Seo, Lehrmann, Han, and Sigal}]{seo2017visual}
Paul~Hongsuck Seo, Andreas Lehrmann, Bohyung Han, and Leonid Sigal. 2017.
\newblock Visual reference resolution using attention memory for visual dialog.
\newblock In \emph{NIPS}.

\bibitem[{Serban et~al.(2017)Serban, Sordoni, Lowe, Charlin, Pineau, Courville,
  and Bengio}]{serban2017hierarchical}
Iulian~Vlad Serban, Alessandro Sordoni, Ryan Lowe, Laurent Charlin, Joelle
  Pineau, Aaron~C Courville, and Yoshua Bengio. 2017.
\newblock A hierarchical latent variable encoder-decoder model for generating
  dialogues.
\newblock In \emph{AAAI}.

\bibitem[{Sharma et~al.(2018)Sharma, Ding, Goodman, and
  Soricut}]{sharma2018conceptual}
Piyush Sharma, Nan Ding, Sebastian Goodman, and Radu Soricut. 2018.
\newblock Conceptual captions: A cleaned, hypernymed, image alt-text dataset
  for automatic image captioning.
\newblock In \emph{ACL}.

\bibitem[{de~Vries et~al.(2017)de~Vries, Strub, Chandar, Pietquin, Larochelle,
  and Courville}]{de2016guesswhat}
Harm de~Vries, Florian Strub, Sarath Chandar, Olivier Pietquin, Hugo
  Larochelle, and Aaron~C Courville. 2017.
\newblock Guesswhat?! visual object discovery through multi-modal dialogue.
\newblock In \emph{CVPR}.

\bibitem[{Wang et~al.(2020)Wang, Joty, Lyu, King, Xiong, and Hoi}]{wang2020vd}
Yue Wang, Shafiq Joty, Michael~R Lyu, Irwin King, Caiming Xiong, and Steven~CH
  Hoi. 2020.
\newblock Vd-bert: A unified vision and dialog transformer with bert.
\newblock \emph{EMNLP}.

\bibitem[{Yang et~al.(2019)Yang, Zha, and Zhang}]{yang2019making}
Tianhao Yang, Zheng-Jun Zha, and Hanwang Zhang. 2019.
\newblock Making history matter: History-advantage sequence training for visual
  dialog.
\newblock In \emph{ICCV}.

\bibitem[{Zheng et~al.(2019)Zheng, Wang, Qi, and Zhu}]{zheng2019reasoning}
Zilong Zheng, Wenguan Wang, Siyuan Qi, and Song-Chun Zhu. 2019.
\newblock Reasoning visual dialogs with structural and partial observations.
\newblock In \emph{CVPR}.

\end{thebibliography}
